\documentclass[review,10pt]{JMtemplate}

\usepackage{lastpage}
\usepackage[utf8]{inputenc}
\usepackage[T1]{fontenc}

\usepackage{amsmath}
\usepackage{amsfonts}
\usepackage{booktabs}
\usepackage{multirow}
\usepackage{graphicx}
\usepackage{nicefrac}
\usepackage{microtype}
\usepackage{xcolor}
\usepackage{placeins}
\usepackage{lipsum}
\usepackage{longtable}
\usepackage{booktabs}      
\usepackage{threeparttable}
\usepackage{multicol}
\usepackage{array}
\usepackage{subcaption}
\usepackage{changepage}
\usepackage{pdflscape}
\usepackage{makecell}
\usepackage{adjustbox}
\usepackage{changepage}
\usepackage{caption}
\usepackage[figuresright]{rotating}
\graphicspath{{media/}}
\usepackage{marvosym}
\usepackage{algorithm}
\usepackage{algorithmic}

\usepackage{rotating} 

\usepackage[table]{xcolor}

\usepackage{tcolorbox}

\newtcolorbox{promptbox}[1][]{
  colback=gray!10!white,
  colframe=black!75!black,
  title=\textbf{Prompt Template},
  fonttitle=\bfseries,
  boxrule=1pt,
  arc=2mm,
  #1
}

\usepackage[
    colorlinks=true,    
    linkcolor=blue,     
    urlcolor=blue,      
    citecolor=green,    
    filecolor=magenta   
]{hyperref}
\usepackage{textcomp}

\begin{document}
\begin{frontmatter}
\title{GUT: Quantifying and Optimizing the Reasoning Uncertainty of LLMs via Graph Complexity}

\author{\textbf{Shuang Liang}\textsuperscript{\rm 1,2} \quad
\textbf{Xin-Yu Hu}\textsuperscript{\rm 1,2} \quad
\textbf{Xiang-Jun Ou}\textsuperscript{\rm 2} \quad
\textbf{Shao-Qun Zhang}\textsuperscript{\rm 1,2,\Letter} \\[0.3em]
\small \textsuperscript{1} National Key Laboratory for Novel Software Technology, Nanjing University, China.\\
\small \textsuperscript{2} School of Intelligent Science and Technology, Nanjing University, China.\\
\small \texttt{ zhangsq@lamda.nju.edu.cn }
}

\begin{abstract}
Recent years have witnessed great advances in the reasoning ability of Large Language Models (LLMs). However, the reasoning processes of LLMs often exhibit uncertainty, where LLMs often produce a proliferation of divergent branches at each reasoning step even when fed the same prompting inputs, and certain branches exhibit evidently incredible, even nonsensical, reasoning chains and results. In this paper, we propose the Graph-complexity-based UncerTainty (GUT) method for investigating the reasoning uncertainty of LLMs. The key idea of GUT is to characterize the potential branches of each reasoning chain with a directed acyclic graph, thereby ensuring that all potential branches are comprehensively covered within the graph space. Building upon this recognition, we further build two modules of GUT, that is, a Quantification (GUT-Q) module and an Optimization (GUT-O) module, for quantifying and reducing the reasoning uncertainty of LLMs, respectively. GUT-Q measures LLM reasoning uncertainty by approximating the reasoning space complexity with graph complexity. GUT-O implements uncertainty optimization by treating negative uncertainty as the reward function in reinforcement learning. Experimental results conducted on four LLMs and five datasets validate the effectiveness of GUT.

\end{abstract}
\end{frontmatter}

\section{Introduction}  \label{sec:intro}
Uncertainty arises when the intrinsic stochasticity of the system results in the variability in its outputs or decisions~\citep{raiffa1968decision}. In particular, when developers employ Large Language Models (LLMs) for reasoning, they frequently encounter the reasoning uncertainty originating from temperature-based stochastic sampling over token probability distributions, which in turn induces variability in the generated reasoning chains~\citep{liu2025uncertainty}. Thus, it is necessary and significant to quantify and optimize the reasoning uncertainty of LLMs, which has been widely deployed in fields of medical care~\citep{atf2025challenge}, autonomous driving~\citep{wang2023empowering}, and quantitative trading~\citep{nie2024survey}.

\begin{figure*} [t]
    \centering
    \includegraphics[width=\linewidth]{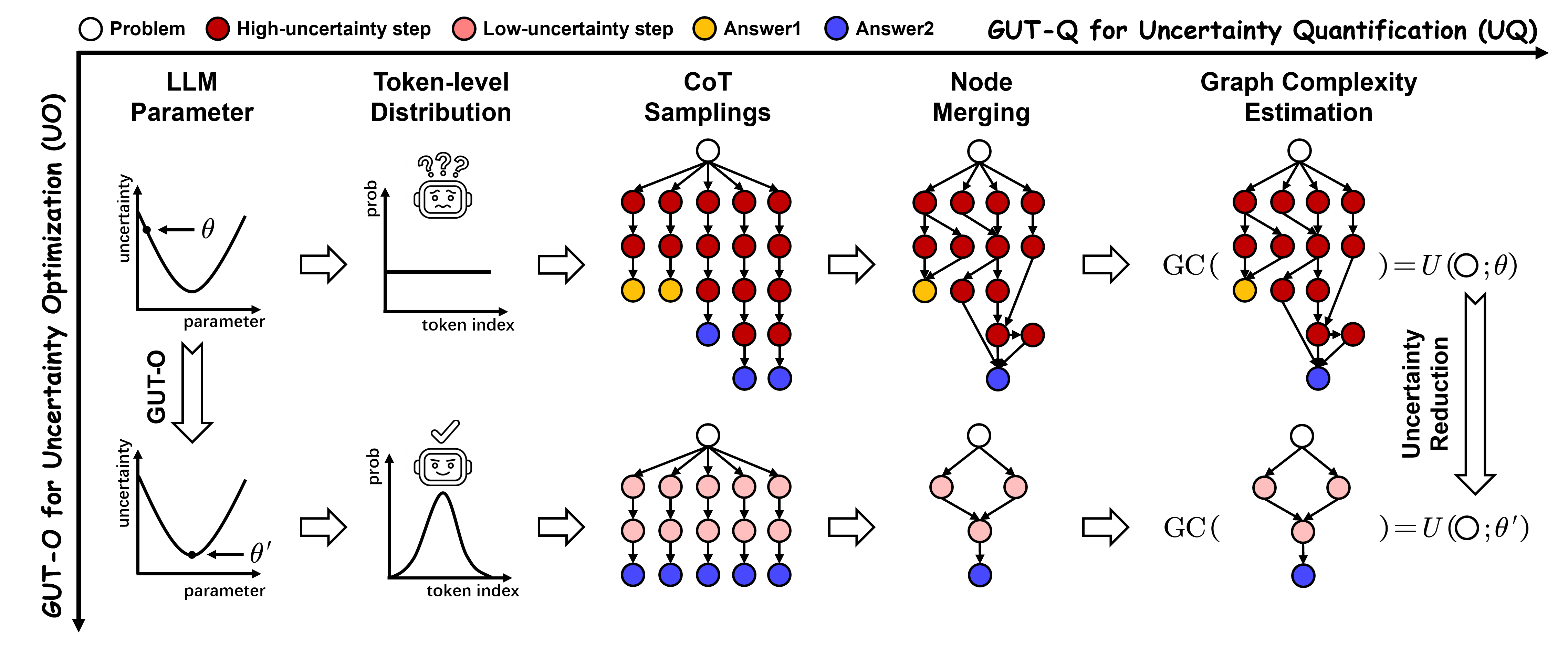}
    \caption{Illustration of our proposed GUT. A real-world case study of GUT is provided in Appendix~\ref{app:case_study}.}
    \label{fig:overview}
\end{figure*}

There have been lots of efforts on quantifying the reasoning uncertainty of LLMs. An intuitive way is to prompt the LLM to judge its own reasoning uncertainty~\citep{kadavath2022ptrue,xiong2024can}. However, the effectiveness of this manner is fundamentally limited by the inherent stochasticity of the prompt-based generation process, induced by the token-level distribution and stochastic sampling, thus recursively amplifying the uncertainty in LLM reasoning. Subsequent studies modeled the reasoning chain as a sequence, based on which researchers quantified the reasoning uncertainty of LLMs by exploiting either the statistical information of a single sampled sequence~\citep{lin2024csl,malinin2021mcse} or the diversity of multiple sampled sequences~~\citep{farquhar2024detecting,lin2024generating}. Nevertheless, modeling reasoning chains as simple sequences fails to accurately capture the inherent uncertainty in LLM reasoning, as it overlooks the potential branches emerging at each individual reasoning step. Recent studies~\citep{da2025understanding,mo2024tree,zhang2025all} constructed graphs to explicitly model potential branches of reasoning steps and quantified the reasoning uncertainty by the topological information of graphs. The graph-based methods achieve a considerably precise modeling of reasoning branches, but rely heavily on self-prompting. Therefore, achieving a precise and effective characterization of the reasoning branches induced by both token-level distributions and stochastic sampling is necessary and challenging for quantifying the LLM reasoning uncertainty.

We implement graph modeling by performing multiple sampling and merging equivalent reasoning steps with a Natural Language Inference (NLI) model, thereby precisely characterizing potential branches in each reasoning step and overcoming effectiveness bottlenecks of self-prompting. Since the collection of all potential branches, that is, the reasoning space~\citep{chen2025towards}, covers all possible reasoning chains induced by stochastic sampling, we can exploit the graph complexity to approximate reasoning space complexity. We construct the graph complexity by integrating statistics over token-level distributions and topological information. The constructed graph complexity is a more comprehensive estimator for quantifying LLM reasoning uncertainty than those of existing UQ methods, since one can trace the reasoning uncertainty back to the stochasticity in both token-level distributions and sampling. By taking the graph-complexity-based uncertainty with a simple proxy approximation as a reward function, we can build the optimization problem within Reinforcement Learning (RL) for reducing the LLM reasoning uncertainty. Figure~\ref{fig:overview} illustrates our key ideas of quantifying and optimizing the reasoning uncertainty of LLMs.

Based on the above recognition, we propose the Graph-complexity-based UncerTainty (GUT) for quantifying and reducing the reasoning uncertainty, comprising a Quantification (GUT-Q) module and an Optimization (GUT-O) module. Our contributions are summarized as follows. Firstly, we propose the GUT-Q for UQ in LLM reasoning that operates by estimating the reasoning space complexity with graph complexity. Empirical evidence across 4 LLMs and 5 datasets shows that the GUT-Q outperforms 45 UQ contenders in the downstream task of selective generation, measured by AUROC, AUPRC, and PRR. Secondly, we propose the GUT-O for Uncertainty Optimization (UO) in LLM reasoning that operates by setting the negative uncertainty proxy as the reward in the RL framework. Empirical evidence across 4 LLM scales and 5 datasets shows that GUT-O effectively reduces the reasoning uncertainty and improves the accuracy.

The rest of this paper is organized as follows. Section~\ref{sec:related_work} reviews related work. Section~\ref{sec:UQ_method} and Section~\ref{sec:UO_method} formally introduce the GUT-Q and GUT-O modules, respectively. Section~\ref{sec:experiments} conducts experiments to validate the effectiveness of our proposed GUT. Section~\ref{sec:conclusion} concludes this work.

\section{Related Work}  \label{sec:related_work}

\paragraph{Reasoning Uncertainty Quantification} Intuitively, one can prompt the LLM to judge its own reasoning uncertainty with sophisticated templates such as multiple choice-based prompts~\citep{kadavath2022ptrue} and reasoning decomposition-based prompts~\citep{xiong2024can}. The prompt-based generation process exhibits inherent stochasticity induced by token-level distribution and stochastic sampling, which recursively amplifies the uncertainty in LLM reasoning. Thus, self-prompting manners are fundamentally limited in their effectiveness. By modeling the reasoning chain as a sequence, subsequent studies quantified the reasoning uncertainty of LLMs through the exploitation of either the statistical information of a single sampled sequence like entropy~\citep{malinin2021mcse} and likelihood~\citep{lin2024csl}, or the diversity among multiple sampled sequences such as eccentricity~\citep{lin2024generating} and the entropy of semantic clusters~\citep{farquhar2024detecting}. These methods depend heavily on the statistical and information-theoretic measures upon the sequence-level modeling. Nevertheless, modeling reasoning chains merely as sequences fails to accurately capture the inherent uncertainty in LLM reasoning, since it neglects the potential branches emerging at each reasoning step. A natural approach to characterize potential branches at each reasoning step is to use a graph to explicitly model the reasoning space, defined as the set of all possible reasoning chains of a problem~\citep{chen2025towards}. Recent studies constructed graphs and quantified reasoning uncertainty by the topological information of graphs, such as graph distance~\citep{da2025understanding} and the number of paths~\citep{zhang2025all}. Although the graph-based methods achieve a considerably precise modeling of reasoning branches, they rely heavily on self-prompting. Therefore, it remains necessary and challenging to achieve a precise and effective characterization of the reasoning branches induced by token distributions and stochastic sampling for quantifying the reasoning uncertainty of LLMs.

\paragraph{Reasoning Uncertainty Optimization} There has been a surge of interest in incorporating uncertainty measures to enhance LLM reasoning. \citet{bi2025forest} utilized LLM logits to identify and prevent the expansion of high-uncertainty nodes in the ToT model~\cite{yao2023tree}. Some studies detected and terminated the generation of highly uncertain reasoning chains by leveraging the statistics of token-level distributions, such as entropy~\citep{fu2025deep}, the probability gap between the top-1 and top-2 tokens~\citep{zhu2025uncertainty}, and the logit variation between consecutive token positions~\citep{yin2024reasoning}. Hence, these studies primarily focused on exploring the reasoning space by exploiting uncertainty as a signal, rather than reducing the reasoning uncertainty. Therefore, the UO in LLM reasoning remains largely unexplored.

\section{Uncertainty Quantification}  \label{sec:UQ_method}
In this section, we formally propose the GUT-Q module for UQ in LLM reasoning. Before that, we introduce some useful notations. Let $[N] = \{1, 2, ..., N\}$ be an integer set for $N \in \mathbb{N}^+$, and $|\cdot|$ denotes the number of elements in a collection, e.g., $|[N]|=N$. Let $\lfloor z \rfloor$ denote the greatest integer less than or equal to $z\in \mathbb{R}$, e.g., $\lfloor 1.1 \rfloor = 1$.

We start by formalizing the process of LLM reasoning. Given a problem $x$ sampled from the distribution $\mathcal{X}$ and an instruction $I$, the LLM parameterized by $\theta$ maps the context $(x,I)$ to a token-level distribution $\pi_{\theta}$ over the vocabulary $\mathcal{V}$. A single reasoning process typically takes the form of a Chain-of-Thought (CoT) $c^i$, containing a sequence of $n_i \in \mathbb{N}^+$ reasoning steps $\{ s^i_k \}_{k \in[n_i]}$ for $i \in \mathbb{N}^+$. At each step, the LLM performs temperature sampling over the token-level distribution $\pi_{\theta}(\cdot | h_{<j} )$ autoregressively to obtain a sequence of tokens and form a new context, where $h_{<j}$ denotes the context $(t_{j-1},...,t_1,x,I)$ for $2 \le j \le |c|$ and $(x,I)$ for $j=1$. Hence, the stochasticity of temperature sampling over the token-level distribution $\pi_{\theta}(\cdot | h_{<j} )$ induces $n_k \in \mathbb{N}^+$ potential branches $\{ s^i_k \}_{i\in[n_k]}$ at the $k$-th step for $k\in [n_i]$. The CoT $c^i$ is generated by recursively repeating this sampling process. Finally, one can obtain the reasoning chain $c^i = \{ s^i_k \}_{k \in[n_i]}$, where the last step $s^i_{n_i}$ contains the final answer. This final answer admits potential candidates induced by the potential branches at each step. 

As shown in Figure~\ref{fig:overview}, after merging semantically equivalent nodes, for the LLM with parameter $\theta$ and flat token-level distributions, there are four potential branches at the first reasoning step and two final answers; while for the LLM with parameter $\theta^{\prime}$ and peak token-level distributions, there are only two potential branches at the first step and one final answer. It is observed that more potential branches at each step would induce a more complex topology of the underlying reasoning space, which can be characterized as higher complexity. Based on this observation, it is natural to quantify the reasoning uncertainty by the complexity of the reasoning space, which can be approximated by the graph complexity. 

The proposed GUT-Q constructs a graph to reveal the potential branches of each step, then quantifies step-level uncertainty to capture token-level stochasticity, and finally exploits graph complexity to approximate reasoning space complexity, which in turn enables the estimation of reasoning uncertainty. GUT-Q comprises graph construction, step-level uncertainty quantification, and graph complexity estimation, detailed in Subsections~\ref{subsec:graph_construction},~\ref{subsec:steplevel_uq}, and~\ref{subsec:graph_complexity_estimation}, respectively.

\subsection{Graph Construction}  \label{subsec:graph_construction}
This subsection formalizes how GUT-Q constructs a Directed Acyclic Graph (DAG) to characterize the reasoning space. The key idea is to first initialize a DAG with multiple CoT chains and then merge semantically equivalent nodes to uncover the intrinsic potential branches of each step within the reasoning space. We begin by sampling $K \in \mathbb{N}^+$ reasoning chains $\{c^i\}_{i \in [K]}$ and extracting reasoning steps $\{s^i_j\}_{j \in [n_i]}$ for $n_i \in \mathbb{N}^+$. Next, we initialize a DAG rooted at the problem $x$. This DAG comprises nodes representing reasoning steps $s_{j}^{i}$. These nodes are connected by directed edges that point to the next step. Specifically, directed edges connect the root $x$ to the initial steps $s_{1}^{i}$ and link subsequent steps $s_{j-1}^{i}$ to $s_{j}^{i}$. Each path terminates at a leaf node representing the final answer $s^i_{n_i}$, which is typically a real-valued scalar or vector for mathematical datasets like MATH-500.

Intuitively, there are three ways to merge semantically equivalent nodes. First, one can prompt an LLM to judge the equivalence of two nodes. This method suffers from high LLM inference costs and the inherent stochasticity of prompt-based generation. Second, equivalence can be established when the cosine similarity of the two nodes' corresponding embeddings exceeds a pre-specified threshold. This method may require a task-specific choice of embedding model and careful tuning of this threshold. Third, two nodes are considered equivalent if they are bidirectionally entailed by an NLI model, i.e., a node entails the other and vice versa. This approach overcomes the limitations of the previous two methods by running a small-scale NLI model that yields deterministic entailment judgments. Thus, as an example in this work, we adopt the third implementation and validate its effectiveness in Subsection~\ref{subsec:UQ_experiment} and Appendix~\ref{app:UQ_additional_exp}. When two nodes are merged, the directed edges are redirected to the newly formed node. Notably, we traverse the node pairs in the AOV-based order~\citep{horowitz1976fundamentals} to ensure that the graph remains a DAG after merging, a property that a naive traversal over all node pairs cannot preserve. Figure~\ref{fig:overview} illustrates the constructed DAG $G=(V, E)$. Prompt template, pseudocode of DAG construction procedure, and configurations of hyperparameters are provided in Appendices~\ref{app:subsec:prompt_template},~\ref{app:subsec:algorithm}, and~\ref{app:subsec:UQ_hyperparameter}, respectively.

\subsection{Step-level Uncertainty Quantification}  \label{subsec:steplevel_uq}
This subsection introduces step-level uncertainty quantification for capturing stochasticity originating from token-level distributions. Let $v \in V$ be an arbitrary node that represents a reasoning step $s$ consisting of tokens $\{t_j\}_{j\in[ |s| ]}$. The key idea is to calculate token-level uncertainty and then aggregate them to derive step-level uncertainty. We employ four kinds of token-level uncertainty measures $U(t_j)$ as follows.
\begin{itemize}
    \item Neg Max Prob $U( t_j ) = - \max _{v\in \mathcal{V}}\pi_{\theta}( v | h_{<j}) $ \ , 
    \item Avg Log Prob $U(t_j) = \sum_{v\in \mathcal{V}}^{}{\log \pi_{\theta}(v | h_{<j})} / | \mathcal{V} | $ \ ,
    \item Entropy $U(t_j)=-\sum_{v\in \mathcal{V}}^{}{\pi_{\theta}( v | h_{<j} ) \log}\pi_{\theta}( v | h_{<j}) $ \ ,
    \item Neg Token Prob $U( t_j ) = - \pi_{\theta}( t_j | h_{<j}) $ \ ,
\end{itemize}
where Neg, Prob, and Avg denote negative, probability, and average, respectively. A larger $U(t_j)$ indicates higher uncertainty of token $t_j$. Next, we perform group-level aggregation as $U_i =\sum\nolimits_{j=i}^{i+w-1}{U( t_j )} / w$, yielding a set $\{U_i\}_{i\in[m-w+1]}$ for $w \in [m]$. This approach aims to capture local uncertainty in reasoning steps, which a naive average of $U(t_j)$ over the entire step~\citep{fomicheva2020Ppl,manakul2023selfcheckgpt} fails to capture~\citep{fu2025deep}. Prior work indicated that certain tokens, such as the initial~\citep{zhu2025uncertainty} and final ones~\citep{wang2024chain}, are vital to the uncertainty quantification of a reasoning step. Thus, we derive the step-level uncertainty $U(s)$ by averaging the token-level uncertainties at specific positions. Specifically, we consider the following three ways to select these positions.
\begin{itemize}
    \item Top-$d\%$. Selecting the largest $d\%$ in $\{U_i\}_{i\in[m-w+1]}$.
    \item Head-$d\%$. Selecting the initial $d\%$ in $\{U_i\}_{i\in[m-w+1]}$, namely $\{ U_i \} _{i\in [ \lfloor (m-w+1) d\% \rfloor ]}$.
    \item Tail-$d\%$. Selecting the final $d\%$ in $\{U_i\}_{i\in[m-w+1]}$, namely $\{ U_i \} _{i\in \left\{ m-1-\lfloor (m-w+1)d\% \rfloor ,...,m \right\}}$.
\end{itemize}
The step-level uncertainty $U(s)$ can be viewed as an attribute of its corresponding node, termed node uncertainty $U(v)$.

\subsection{Graph Complexity Estimation}  \label{subsec:graph_complexity_estimation}
In this subsection, we construct graph complexity by exploiting node uncertainty and graph topology to quantify reasoning uncertainty, as illustrated in Figure~\ref{fig:overview}. Specifically, we introduce three graph complexity estimation methods, including Width (GUT-Q-W), Height (GUT-Q-H), and Uncertainty Propagation (GUT-Q-UP). 

\begin{algorithm}[t]
\caption{GUT-Q-W}
\label{alg:width_prob}
\textbf{Input:} DAG $G=(V,E)$, node uncertainty $\{ U(v) \}_{v\in V}$ \\
\textbf{Output:} Uncertainty-weighted width $W_u$ \\
\textbf{Procedures:}
\begin{algorithmic}[1]
\STATE $L \gets \text{TopologicalSort}(G)$ \COMMENT{Get topological order}
\STATE $R[v] \gets 0$ for all $v \in V$ \COMMENT{Initialize topological level}
\FOR{$u \in L$}
    \STATE $R[u] \gets \max(\{R[p] + 1 \mid (p, u) \in E\} \cup \{0\}) $
\ENDFOR
\STATE $C \gets \text{CountFrequencies}(R)$
\STATE $r_w \gets \min ( \{ r \mid C[r] = \max (C) \} )$
\STATE $V_w \gets \{v \mid R[v] = r_w \}$ \COMMENT{Get width-path nodes}
\STATE $W_u \gets \sum_{v \in V_w} U(v)$
\end{algorithmic}
\end{algorithm}

\paragraph{GUT-Q-W} It is intuitive to first find the nodes on the width path and then sum the node uncertainty over these nodes. The width of a DAG represents the maximum number of branches at any single step in the reasoning space. A large width indicates the existence of at least one step with numerous potential branches. Thus, the sum of node uncertainty over the width path, that is, uncertainty-weighted width $W_u$, can reflect the divergence of potential reasoning branches. The key idea for finding the width is to first perform topological sorting~\citep{cormen2022introduction} to obtain the topological order of the nodes and then find the maximum number of nodes at the same topological level. Algorithm~\ref{alg:width_prob} summarizes the calculation process of $W_u$ for estimating graph complexity.

\paragraph{GUT-Q-H} Dual to GUT-Q-W, one can find the nodes on the height path and then sum the node uncertainty over them. The height of a DAG represents the length of the longest reasoning chain from the root to a leaf node, measured by the number of edges along this path. A large height indicates the existence of at least one reasoning chain with numerous steps. Hence, the sum of node uncertainty over the height path, namely uncertainty-weighted height, captures both the depth of the reasoning space and the token-level stochasticity along this path. The key idea of finding the height is to first perform topological sorting to obtain the topological order of the nodes, identify the maximum topological level, and then select a sequence of nodes across these levels that are connected by edges. Algorithm~\ref{alg:height_prob} in Appendix~\ref{app:subsec:algorithm} lists the procedure of GUT-Q-H for graph complexity estimation.

\begin{algorithm}[t]
\caption{GUT-Q-UP}
\label{alg:DAG_UP}
\textbf{Input:} DAG $G=(V,E)$, node uncertainty $\{U(v)\}_{v \in V}$, weight $\omega \in \mathbb{R}^+$, activation function $\phi$ \\
\textbf{Output:} Auxiliary node uncertainty $U(s_a)$ \\
\textbf{Procedures:}
\begin{algorithmic}[1]
\STATE Construct augmented graph $G'=(V', E')$ by connecting all leaf nodes of $G$ to a new auxiliary node $s_a$
\STATE Construct an FNN corresponding to $G'$ with $\phi$
\STATE Set weights $w_{uv} \gets \omega$ for all $(u, v) \in E'$
\STATE Set biases $b_v \gets U(v)$ for all $v \in V$, and $b_{s_a} \gets 0$
\STATE $U(s_a) \gets$ output of $s_a$ from forward propagation
\end{algorithmic}
\end{algorithm}

\paragraph{GUT-Q-UP} GUT-Q-W and GUT-Q-H may not fully exploit node uncertainty $U(v)$ or the detailed connectivity among nodes. Thus, we propose GUT-Q-UP to make fuller use of them. Since a DAG is naturally a Feedforward Neural Network (FNN)~\citep{scarselli2008graph} and uncertainty accumulates along reasoning steps~\citep{gan2025rethinking,zhang2024how}, it is intuitive to propagate node uncertainty $U(v)$ from the first reasoning steps $\{s^i_1\}_{i \in [K]}$ to an auxiliary node $s_a$, where all final-answer nodes $\{ s_{n_i }^{i} \}_{i \in [K]}$ point to it. Therefore, $U(s_a)$, the output of this FNN after forward propagation, incorporates node uncertainty propagated throughout the reasoning space. Algorithm~\ref{alg:DAG_UP} outlines the GUT-Q-UP algorithm to estimate graph complexity, where the node uncertainty $U(v)$ is used to initialize the biases of the FNN associated with $G$. Algorithm~\ref{alg:DAG_UP_weights} in Appendix~\ref{app:subsec:algorithm} lists a variant that initializes this FNN's weights with $U(v)$.

GUT-Q-W and GUT-Q-H excel in explainability over GUT-Q-UP, since they characterize the intrinsic topology of the reasoning space; GUT-Q-W captures the maximum number of potential branches at any single step, and GUT-Q-H integrates the length of the longest reasoning chain. In contrast, GUT-Q-UP typically outperforms GUT-Q-W and GUT-Q-H in the downstream task of selective generation~\citep{ren2023out}, as shown by the empirical evidence in Subsection~\ref{subsec:UQ_experiment}. These findings reveal the usage scenarios of these three graph complexity estimation methods, where one may employ the GUT-Q-W and GUT-Q-H for the explainability of the intrinsic topology of the reasoning space, and employ the GUT-Q-UP for the downstream task of selective generation.

\section{Uncertainty Optimization}  \label{sec:UO_method}
This section formally proposes the GUT-O module for optimizing the LLM reasoning uncertainty $U(x)$ measured by GUT-Q. Intuitively, one can treat $U(x)$ as an optimization target and solve this optimization by exploiting standard gradient-based algorithms. Unfortunately, this way is infeasible, as $U(x)$, with the discrete nature caused by node merging, is non-differentiable with respect to LLM parameters. We adopt indirect optimization by exploiting the law of large numbers~\citep{bill1995probability} as $U( x ) =\mathbb{E} _c[ U( x, c ) ] \approx \sum_{i=1}^N{U( x, c^i )} / N$ for $N \in \mathbb{N}^+$. Thus, optimizing $U(x, c)$ helps optimize $U(x)$, enabling us to employ a differentiable $U(x, c)$ as an optimization target for indirect optimization.

In this work, we take Mean Token Log Probability (MTLP)~\citep{manakul2023selfcheckgpt} as an example of $U(x,c)$, which is a differentiable function over the token-level distribution and is thus directly differentiable with respect to the LLM parameters. To verify the validity of the MTLP optimization target, we calculate the Pearson Correlation Coefficient (PCC) between $U(x)$ and $\sum_{i=1}^N U(x, c^i)/N$, where a positive PCC indicates that optimizing $U(x,c)$ can effectively optimize $U(x)$. Table~\ref{tab:PCC_4B_maintext} reports these PCCs on the investigated mathematical reasoning datasets, averaged across four Qwen3 LLMs. It is observed that PCCs are consistently positive across all datasets, which validates the effectiveness of the MTLP optimization target. Inspired by~\citet{prabhudesai2025maximizing} and~\citet{zhao2025learning}, we employ the gradient-based GRPO algorithm~\citep{shao2024deepseekmath} to optimize the MTLP objective $U(x,c)$ by setting its negative as the reward. Details regarding PCC results of individual LLMs and the GUT-O formulation are provided in Appendices~\ref{app:subsec:proxy_validation} and~\ref{app:subsec:UO_implementation}, respectively. 

\begin{table}[ht]
  \centering
  \begin{tabular}{lccc}
    \toprule
    \multicolumn{1}{c}{\multirow{1}{*}{\textbf{UQ}}} & \textbf{GSM8K} & \textbf{MATH-500} & \textbf{AMC2022-2024} \\
    \midrule
    GUT-Q-W     &  0.1909 $\pm$ 0.0207 &  0.2364 $\pm$ 0.0315 &  0.3603 $\pm$ 0.0592 \\
    GUT-Q-H    &  0.1365 $\pm$ 0.0229 &  0.1945 $\pm$ 0.0318 &  0.3197 $\pm$ 0.0602 \\
    GUT-Q-UP      &  0.2110 $\pm$ 0.0213 &  0.2069 $\pm$ 0.0334 &  0.2586 $\pm$ 0.0600 \\
    \bottomrule
  \end{tabular}
  \caption{PCCs between MTLP and complexity-based uncertainty metrics across investigated mathematical reasoning datasets, averaged across 4 Qwen3 LLMs.}
  \label{tab:PCC_4B_maintext}
\end{table}

\section{Experiments}  \label{sec:experiments}
In this section, we conduct experiments to validate the proposed GUT framework. The experiments are performed to answer (Q1) whether and to what extent the proposed UQ can contribute to discriminating between correct and incorrect generations over classical UQ methods, and (Q2) whether and to what extent the proposed GUT-O reduces the graph-complexity-based uncertainty.

Experiments were conducted on NVIDIA RTX PRO 6000 96G GPUs ($\times$8). Evaluated LLMs involve the Qwen3~\citep{an2025qwen} family across scales of 8B, 4B, 1.7B, and 0.6B. The configuration of sampling parameters is detailed in Appendix~\ref{app:subsec:UQ_hyperparameter}. We follow~\citet{farquhar2024detecting} and use the Deberta-large~\citep{he2021deberta} that is fine-tuned on the MNLI~\citep{williams2018broad} dataset as the NLI model. The evaluated datasets span three task types, including mathematical reasoning, first-order logic reasoning, and long-form Question Answering (QA). For mathematical reasoning, we select three datasets at increasing levels of difficulty, namely GSM8K~\citep{cobbe2021training}, MATH-500~\citep{lightman2024lets}, and AMC2022-2024. For first-order logic reasoning, we use FOLIO~\citep{han2024folio}. For long-form QA, we adopt MMLU-Pro~\citep{wang2024mmlu}. Details on the evaluated datasets are provided in Appendix~\ref{app:subsec:dataset}.

\subsection{Verifications on GUT-Q} \label{subsec:UQ_experiment}
This subsection validates the effectiveness of the proposed GUT-Q module. Before presenting the results, we introduce the UQ performance evaluations and contenders. Following prior work~\citep{lin2024generating,farquhar2024detecting}, we use the performance on the downstream task of selective generation~\citep{ren2023out} to evaluate a UQ method, i.e., measure a UQ method's ability to discriminate between correct and incorrect generations. We use three metrics to quantify this ability, including the Area Under the Receiver Operating Characteristic curve (AUROC), the Area Under the Precision-Recall Curve (AUPRC), and the Prediction Rejection Ratio (PRR)~\citep{vashurin2025benchmarking}, which is a normalized version of the Area Under the Accuracy-Rejection Curve (AUARC)~\citep{lin2024generating,nadeem2009accuracy}. Higher values indicate better UQ performance for all three metrics. Following~\citet{vashurin2025benchmarking}, we bootstrap datasets 1000 times and report the mean and standard deviation of the metrics as mean$\pm$std.

We compared 45 white-box and black-box UQ contenders across five categories, including the diversity-based methods like semantic entropy~\citep{farquhar2024detecting} and graph Laplacian~\citep{lin2024generating}, the information-based methods such as CSL~\citep{lin2024csl}, the self-reflexive methods like P(True)~\citep{kadavath2022ptrue}, the reasoning-enhanced methods like CoT-UQ~\citep{zhang2025cot}, and the graph-based methods like TopoUQ~\citep{da2025understanding}. Details are provided in Appendix~\ref{app:subsec:UQ_contenders}.

Table~\ref{tab:UQ_4B_main_text} shows the comparisons of UQ performance for Qwen3-4B across partial datasets, where the upper group lists the performance of contenders and the lower group details that of ours. Bold and underline denote the best and second-best results within each group, respectively. Results across five datasets are presented in Table~\ref{tab:UQ_4B_app} in Appendix~\ref{app:subsec:UQ_performance_evaluations}. It is observed that the GUT-Q implemented with the UP algorithm consistently outperforms all contenders across three evaluation metrics and datasets spanning three difficulty levels. There are similar observations for all 4 investigated LLMs and 5 datasets, as detailed in Appendix~\ref{app:subsec:UQ_performance_evaluations}. These observations show that GUT-Q outperforms the best UQ contenders by an average of $11.79$\% in PRR, $13.33$\% in AUROC, and $9.66$\% in AUPRC, across four evaluated LLMs and five datasets, which answers Q1. Moreover, GUT-Q exhibits comparable runtime complexity to existing start-of-the-art methods such as SE and Eig-C, where Appendix~\ref{app:subsec:runtime} provides a detailed analysis.

\begin{table*}[ht]
  \centering
  \footnotesize
  \setlength{\tabcolsep}{1.5pt}
  \resizebox{0.95\linewidth}{!}{
    \begin{tabular}{lrrrrrrrrr}
    \toprule
    \multicolumn{1}{c}{\multirow{2}{*}[-0.8ex]{\textbf{UQ}}} & \multicolumn{3}{c}{\textbf{GSM8K}} & \multicolumn{3}{c}{\textbf{MATH-500}} & \multicolumn{3}{c}{\textbf{AMC2022-2024}} \\
    \cmidrule(lr){2-4} \cmidrule(lr){5-7} \cmidrule(lr){8-10}
               & \multicolumn{1}{c}{\textbf{PRR}} & \multicolumn{1}{c}{\textbf{AUROC}} & \multicolumn{1}{c}{\textbf{AUPRC}} & \multicolumn{1}{c}{\textbf{PRR}} & \multicolumn{1}{c}{\textbf{AUROC}} & \multicolumn{1}{c}{\textbf{AUPRC}} & \multicolumn{1}{c}{\textbf{PRR}} & \multicolumn{1}{c}{\textbf{AUROC}} & \multicolumn{1}{c}{\textbf{AUPRC}} \\
    \midrule
    MSP        & -33.66$\pm$1.22                  & 58.89$\pm$0.97                     & 86.84$\pm$0.80                     & -8.09$\pm$2.46                   & 58.76$\pm$1.67                     & 60.85$\pm$2.04                     & -37.65$\pm$4.97                  & 67.35$\pm$3.70                     & 52.35$\pm$4.27                     \\
    Ppl        & -33.66$\pm$1.22                  & 58.89$\pm$0.98                     & 86.84$\pm$0.79                     & -8.09$\pm$2.46                   & 58.76$\pm$1.65                     & 60.85$\pm$2.01                     & -37.65$\pm$4.97                  & 67.35$\pm$3.77                     & 52.35$\pm$4.41                     \\
    MTE        & 47.45$\pm$0.86                   & 28.89$\pm$1.73                     & 74.56$\pm$1.26                     & 31.08$\pm$2.36                   & 36.22$\pm$1.99                     & 44.79$\pm$2.13                     & 37.77$\pm$4.71                   & 26.06$\pm$3.79                     & 28.18$\pm$3.57                     \\
    MTLP       & -27.77$\pm$1.13                  & 57.30$\pm$1.52                     & 84.56$\pm$0.91                     & 2.10$\pm$2.40                    & 55.29$\pm$1.95                     & 62.99$\pm$2.53                     & 27.33$\pm$5.00                   & 28.86$\pm$3.86                     & 26.17$\pm$3.20                     \\
    MPMI       & -28.86$\pm$1.13                  & 59.13$\pm$1.58                     & 87.55$\pm$0.86                     & 4.67$\pm$2.37                    & 57.56$\pm$2.01                     & 63.94$\pm$2.56                     & -12.39$\pm$4.79                  & 48.12$\pm$4.37                     & 43.94$\pm$5.05                     \\
    CPMI       & -11.23$\pm$0.99                  & 43.75$\pm$1.61                     & 80.60$\pm$1.17                     & 15.15$\pm$2.32                   & 47.15$\pm$1.99                     & 57.15$\pm$2.37                     & -56.69$\pm$4.14                  & 74.24$\pm$4.09                     & 61.25$\pm$5.55                     \\
    RD         & 14.53$\pm$1.08                   & 41.01$\pm$0.97                     & 83.88$\pm$0.89                     & 25.17$\pm$2.50                   & 41.30$\pm$1.55                     & 55.70$\pm$1.97                     & 41.40$\pm$5.15                   & 33.13$\pm$3.65                     & 39.16$\pm$4.02                     \\
    FRD        & 14.52$\pm$1.08                   & 41.01$\pm$0.99                     & 83.88$\pm$0.83                     & 25.78$\pm$2.50                   & 41.35$\pm$1.61                     & 55.79$\pm$1.94                     & 41.38$\pm$5.13                   & 33.16$\pm$3.65                     & 39.18$\pm$4.33                     \\
    AS         & -17.30$\pm$1.19                  & 53.34$\pm$1.72                     & 83.69$\pm$0.97                     & 18.19$\pm$2.45                   & 48.81$\pm$2.07                     & 57.56$\pm$2.42                     & 25.50$\pm$4.85                   & 29.96$\pm$3.98                     & 30.58$\pm$3.69                     \\
    CSL        & -39.67$\pm$1.17                  & 60.61$\pm$1.66                     & 87.09$\pm$0.88                     & 4.33$\pm$2.40                    & 55.39$\pm$2.01                     & 60.42$\pm$2.51                     & 13.68$\pm$4.96                   & 30.56$\pm$3.99                     & 28.06$\pm$3.56                     \\
    MCSE       & 6.55$\pm$1.11                    & 50.00$\pm$0.00                     & 84.31$\pm$0.79                     & -4.91$\pm$2.52                   & 51.33$\pm$0.41                     & 55.47$\pm$1.75                     & -26.20$\pm$4.61                  & 52.74$\pm$1.12                     & 41.03$\pm$3.58                     \\
    MCNSE      & 6.55$\pm$1.11                    & 50.00$\pm$0.00                     & 84.31$\pm$0.81                     & -4.91$\pm$2.52                   & 51.33$\pm$0.40                     & 55.47$\pm$1.70                     & -26.20$\pm$4.61                  & 52.74$\pm$1.06                     & 41.03$\pm$3.70                     \\
    RAUQ       & -29.39$\pm$1.18                  & 55.62$\pm$1.77                     & 84.21$\pm$0.99                     & 3.25$\pm$2.47                    & 55.51$\pm$2.03                     & 62.72$\pm$2.56                     & 32.58$\pm$4.96                   & 24.12$\pm$3.90                     & 24.97$\pm$3.28                     \\
    RAUQ-E     & -40.55$\pm$1.22                  & 60.84$\pm$1.60                     & 85.45$\pm$0.88                     & 5.56$\pm$2.45                    & 55.45$\pm$2.09                     & 62.28$\pm$2.60                     & 39.79$\pm$4.90                   & 22.54$\pm$3.64                     & 24.80$\pm$3.28                     \\
    SE         & -6.65$\pm$1.09                   & 52.71$\pm$1.18                     & 85.13$\pm$0.79                     & 0.14$\pm$2.42                    & 56.45$\pm$1.93                     & 61.98$\pm$2.32                     & 0.54$\pm$4.92                    & 55.48$\pm$4.29                     & 50.38$\pm$5.25                     \\
    SAR        & 59.47$\pm$0.57                   & 26.42$\pm$1.39                     & 75.20$\pm$1.28                     & 33.61$\pm$2.43                   & 35.38$\pm$1.92                     & 44.61$\pm$1.95                     & 51.98$\pm$4.70                   & 26.63$\pm$3.55                     & 28.42$\pm$3.61                     \\
    TSAR       & 15.14$\pm$1.08                   & 40.38$\pm$1.04                     & 82.05$\pm$0.86                     & 24.77$\pm$2.50                   & 37.47$\pm$1.65                     & 49.21$\pm$1.71                     & 19.41$\pm$5.17                   & 30.59$\pm$3.38                     & 35.57$\pm$3.68                     \\
    SSAR       & -27.96$\pm$1.17                  & \textbf{65.41$\pm$1.45}            & \textbf{90.38$\pm$0.73}            & 10.34$\pm$2.40                   & 51.54$\pm$1.97                     & 61.69$\pm$2.39                     & 6.27$\pm$5.13                    & 52.81$\pm$4.60                     & 55.40$\pm$5.35                     \\
    SD         & 33.10$\pm$0.78                   & 37.76$\pm$1.49                     & 79.16$\pm$1.15                     & 35.85$\pm$2.39                   & 38.87$\pm$1.99                     & 51.82$\pm$2.37                     & -56.66$\pm$3.58                  & 71.11$\pm$3.86                     & 61.06$\pm$5.18                     \\
    ES         & 40.13$\pm$0.78                   & 36.16$\pm$1.55                     & 79.43$\pm$1.20                     & 35.57$\pm$2.45                   & 29.81$\pm$1.82                     & 40.59$\pm$1.83                     & \underline{65.02$\pm$4.28}       & 21.69$\pm$3.66                     & 31.01$\pm$3.66                     \\
    CoCoA-MSP  & -30.10$\pm$1.20                  & 56.65$\pm$1.70                     & 83.73$\pm$0.92                     & 17.21$\pm$2.47                   & 52.97$\pm$2.04                     & 62.26$\pm$2.48                     & 46.54$\pm$4.48                   & 21.17$\pm$3.68                     & 25.03$\pm$3.23                     \\
    CoCoA-Ppl  & -27.79$\pm$1.15                  & 56.25$\pm$1.71                     & 83.04$\pm$0.97                     & 9.62$\pm$2.48                    & 54.15$\pm$2.01                     & 61.35$\pm$2.55                     & 41.29$\pm$4.89                   & 22.42$\pm$3.84                     & 25.15$\pm$3.28                     \\
    CoCoA-MTE  & -14.85$\pm$1.09                  & 51.17$\pm$1.75                     & 81.48$\pm$1.05                     & 12.23$\pm$2.47                   & 53.71$\pm$2.08                     & 61.89$\pm$2.42                     & 43.05$\pm$4.88                   & 21.68$\pm$3.91                     & 24.69$\pm$3.39                     \\
    P(True)    & 13.30$\pm$1.11                   & 38.54$\pm$1.87                     & 76.44$\pm$1.20                     & 26.47$\pm$2.49                   & 41.07$\pm$2.02                     & 49.49$\pm$2.21                     & 21.78$\pm$4.86                   & 27.79$\pm$4.44                     & 25.47$\pm$4.96                     \\
    CoT-UQ-ME  & 14.47$\pm$5.58                   & 55.51$\pm$1.85                     & 87.51$\pm$1.03                     & 33.48$\pm$4.82                   & \textbf{63.45$\pm$1.94}            & \textbf{69.40$\pm$2.43}            & 25.78$\pm$9.85                   & 61.27$\pm$4.54                     & 51.27$\pm$5.95                     \\
    CoT-UQ-MI  & 17.07$\pm$5.68                   & 56.76$\pm$1.77                     & 87.90$\pm$1.03                     & 28.22$\pm$5.11                   & 61.24$\pm$1.98                     & 67.40$\pm$2.49                     & 20.98$\pm$9.99                   & 59.23$\pm$4.51                     & 48.81$\pm$6.23                     \\
    CoT-UQ-SAR & 14.11$\pm$5.65                   & 55.60$\pm$1.77                     & 87.49$\pm$1.05                     & 33.21$\pm$4.80                   & \underline{63.32$\pm$1.90}         & \underline{69.28$\pm$2.40}         & 26.38$\pm$9.91                   & 61.74$\pm$4.62                     & 51.53$\pm$5.97                     \\
    NS         & 5.15$\pm$1.10                    & 48.00$\pm$0.89                     & 83.66$\pm$0.84                     & 4.12$\pm$2.48                    & 50.00$\pm$0.00                     & 54.80$\pm$1.80                     & -19.38$\pm$5.05                  & 50.00$\pm$0.00                     & 39.67$\pm$3.57                     \\
    Eig-E      & 46.57$\pm$0.68                   & 33.01$\pm$1.61                     & 77.54$\pm$1.31                     & -30.53$\pm$2.42                  & 59.51$\pm$1.92                     & 59.90$\pm$2.51                     & -60.23$\pm$4.79                  & \textbf{76.71$\pm$4.02}            & \underline{62.31$\pm$5.61}         \\
    Eig-C      & \underline{59.48$\pm$0.64}       & 26.08$\pm$1.43                     & 74.93$\pm$1.27                     & 2.81$\pm$2.51                    & 50.61$\pm$2.00                     & 56.19$\pm$2.44                     & -54.14$\pm$4.79                  & 73.40$\pm$3.98                     & 61.59$\pm$5.23                     \\
    Eig-J      & 42.17$\pm$0.72                   & 33.30$\pm$1.62                     & 76.53$\pm$1.26                     & \underline{50.42$\pm$2.17}       & 28.84$\pm$1.73                     & 42.30$\pm$1.81                     & 58.01$\pm$4.40                   & 21.35$\pm$3.85                     & 28.86$\pm$3.97                     \\
    Deg-E      & 47.13$\pm$0.66                   & 33.16$\pm$1.46                     & 77.73$\pm$1.27                     & 3.98$\pm$2.42                    & 52.22$\pm$2.05                     & 56.64$\pm$2.54                     & -59.13$\pm$4.09                  & \underline{76.03$\pm$3.74}         & \textbf{64.78$\pm$5.59}            \\
    Deg-C      & \textbf{60.76$\pm$0.63}          & 25.42$\pm$1.22                     & 74.77$\pm$1.24                     & 47.95$\pm$2.03                   & 34.47$\pm$1.98                     & 47.35$\pm$2.33                     & -11.14$\pm$4.93                  & 57.42$\pm$4.48                     & 54.36$\pm$5.05                     \\
    Deg-J      & 43.13$\pm$0.72                   & 32.69$\pm$1.59                     & 76.33$\pm$1.27                     & \textbf{50.95$\pm$2.16}          & 28.38$\pm$1.75                     & 41.99$\pm$1.83                     & 58.55$\pm$4.34                   & 21.15$\pm$4.05                     & 28.84$\pm$4.01                     \\
    Ecc-E      & 44.92$\pm$0.73                   & 33.20$\pm$1.66                     & 76.98$\pm$1.29                     & -11.87$\pm$2.53                  & 53.81$\pm$1.96                     & 56.34$\pm$2.47                     & -55.52$\pm$4.07                  & 73.72$\pm$3.85                     & 59.81$\pm$5.51                     \\
    Ecc-C      & 56.84$\pm$0.63                   & 27.78$\pm$1.44                     & 75.36$\pm$1.24                     & 32.90$\pm$2.10                   & 43.17$\pm$2.04                     & 53.03$\pm$2.25                     & -28.36$\pm$4.20                  & 63.76$\pm$4.19                     & 53.88$\pm$5.28                     \\
    Ecc-J      & 44.89$\pm$0.76                   & 30.82$\pm$1.77                     & 75.43$\pm$1.21                     & 49.68$\pm$2.25                   & 27.01$\pm$1.74                     & 41.30$\pm$1.77                     & \textbf{67.56$\pm$4.39}          & 17.51$\pm$3.70                     & 26.43$\pm$3.30                     \\
    LS-R1      & 42.64$\pm$0.70                   & 33.21$\pm$1.60                     & 76.48$\pm$1.22                     & 2.72$\pm$2.54                    & 48.62$\pm$2.06                     & 53.11$\pm$2.23                     & -0.49$\pm$4.89                   & 46.63$\pm$4.10                     & 36.83$\pm$3.78                     \\
    LS-R2      & 49.43$\pm$0.67                   & 29.25$\pm$1.62                     & 75.12$\pm$1.22                     & 26.35$\pm$2.42                   & 38.13$\pm$2.02                     & 46.66$\pm$2.02                     & 48.92$\pm$4.48                   & 27.51$\pm$3.68                     & 28.95$\pm$3.37                     \\
    LS-RL      & 49.02$\pm$0.66                   & 29.30$\pm$1.58                     & 74.88$\pm$1.30                     & 35.65$\pm$2.34                   & 33.00$\pm$1.87                     & 43.60$\pm$1.79                     & 60.71$\pm$4.17                   & 20.80$\pm$3.46                     & 27.15$\pm$3.15                     \\
    LS-B       & 45.10$\pm$0.71                   & 31.05$\pm$1.64                     & 75.86$\pm$1.25                     & 29.07$\pm$2.40                   & 37.20$\pm$1.92                     & 45.89$\pm$2.15                     & 47.23$\pm$4.57                   & 27.88$\pm$3.71                     & 28.97$\pm$3.76                     \\
    KLE        & -9.26$\pm$1.11                   & 52.32$\pm$1.75                     & 84.98$\pm$0.99                     & 6.65$\pm$2.47                    & 48.55$\pm$1.94                     & 53.49$\pm$2.31                     & -8.96$\pm$4.93                   & 52.03$\pm$4.21                     & 42.45$\pm$4.45                     \\
    LUQ        & 53.52$\pm$0.68                   & 28.97$\pm$1.40                     & 75.76$\pm$1.23                     & 41.59$\pm$2.06                   & 38.59$\pm$1.87                     & 49.99$\pm$2.44                     & -25.97$\pm$4.64                  & 62.50$\pm$4.32                     & 56.86$\pm$5.22                     \\
    CoT-UQ-SP  & -3.37$\pm$6.33                   & 51.38$\pm$1.90                     & 85.11$\pm$1.12                     & -10.00$\pm$5.34                  & 48.60$\pm$2.06                     & 54.18$\pm$2.26                     & -5.71$\pm$9.67                   & 47.69$\pm$4.60                     & 36.71$\pm$4.63                     \\
    Topo-UQ    & 41.75$\pm$3.74                   & \underline{64.77$\pm$1.59}         & \underline{89.94$\pm$0.79}         & 11.12$\pm$4.69                   & 53.01$\pm$2.20                     & 65.75$\pm$2.58                     & 24.04$\pm$9.16                   & 56.96$\pm$5.37                     & 51.65$\pm$5.64                     \\
    \midrule
    GUT-Q-W    & \underline{72.84$\pm$2.61}       & \underline{81.68$\pm$1.29}         & \underline{95.46$\pm$0.49}         & \underline{54.59$\pm$4.13}       & \underline{76.64$\pm$1.71}         & \underline{75.92$\pm$2.29}         & \underline{69.23$\pm$5.69}       & \underline{82.38$\pm$3.06}         & \underline{76.85$\pm$4.09}         \\
    GUT-Q-H    & 55.87$\pm$3.04                   & 71.38$\pm$1.45                     & 92.73$\pm$0.63                     & 46.49$\pm$4.35                   & 72.65$\pm$1.86                     & 72.29$\pm$2.48                     & 59.56$\pm$6.94                   & 78.41$\pm$3.42                     & 69.97$\pm$5.10                     \\
    GUT-Q-UP   & \textbf{81.68$\pm$1.83}          & \textbf{86.61$\pm$1.02}            & \textbf{96.93$\pm$0.36}            & \textbf{61.93$\pm$3.59}          & \textbf{80.13$\pm$1.57}            & \textbf{79.45$\pm$2.07}            & \textbf{80.82$\pm$4.24}          & \textbf{89.48$\pm$2.19}            & \textbf{84.60$\pm$3.43}            \\
    \bottomrule
  \end{tabular}
  }
  \caption{Comparisons of UQ performance for Qwen3-4B across partial contenders and datasets, where bold and underline denote the group best and second-best results, respectively.}
  \label{tab:UQ_4B_main_text}
\end{table*}
\FloatBarrier

\begin{figure*}[ht]
    \centering
    \begin{subfigure}[b]{0.49\linewidth}
        \centering
        \includegraphics[width=\linewidth]{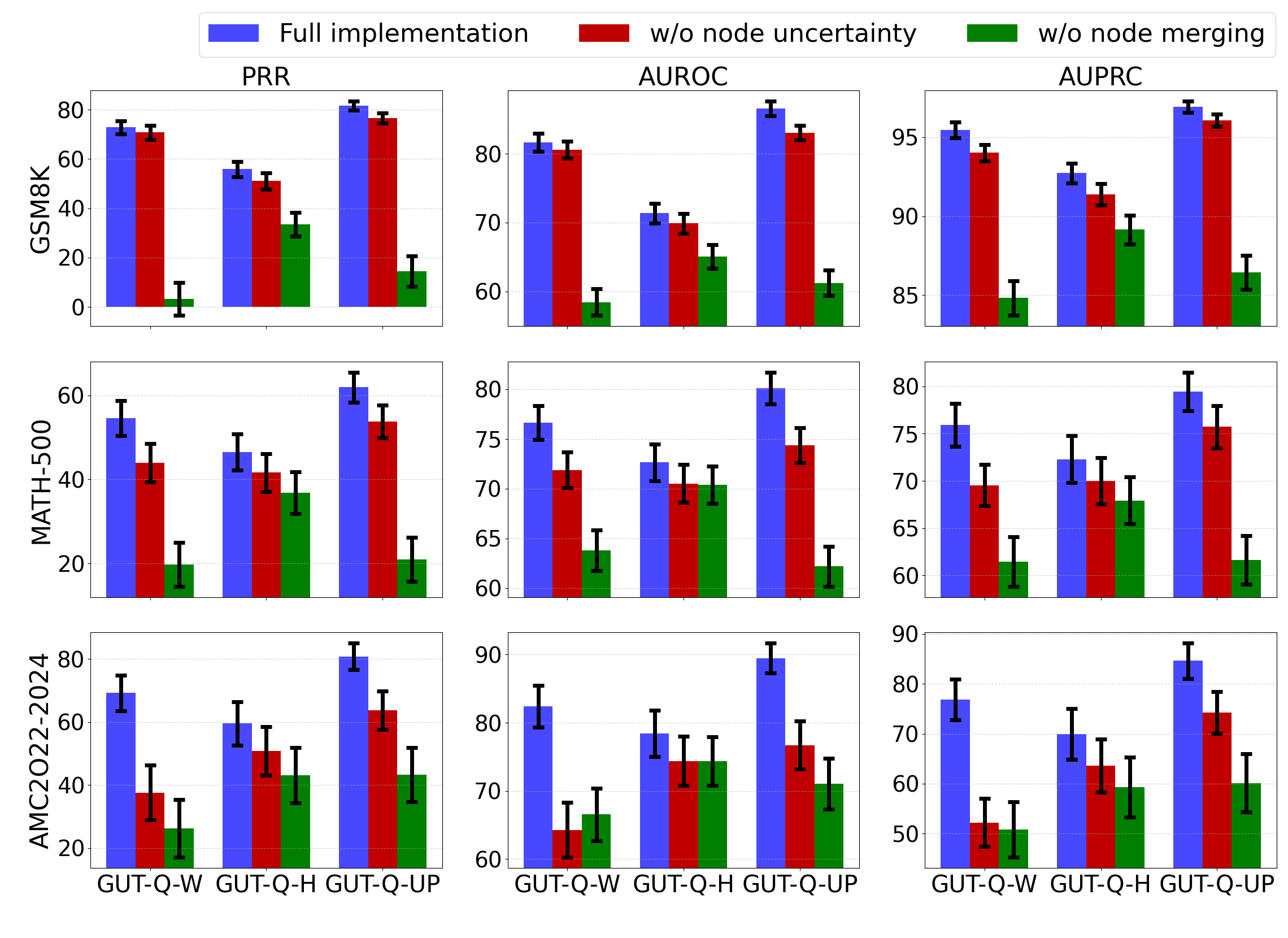}
	    \caption{Ablation comparisons of UQ performance.}
        \label{fig:ablation_4B_maintext}
    \end{subfigure}
    \hfill
    \begin{subfigure}[b]{0.49\linewidth}
        \centering
        \includegraphics[width=\linewidth]{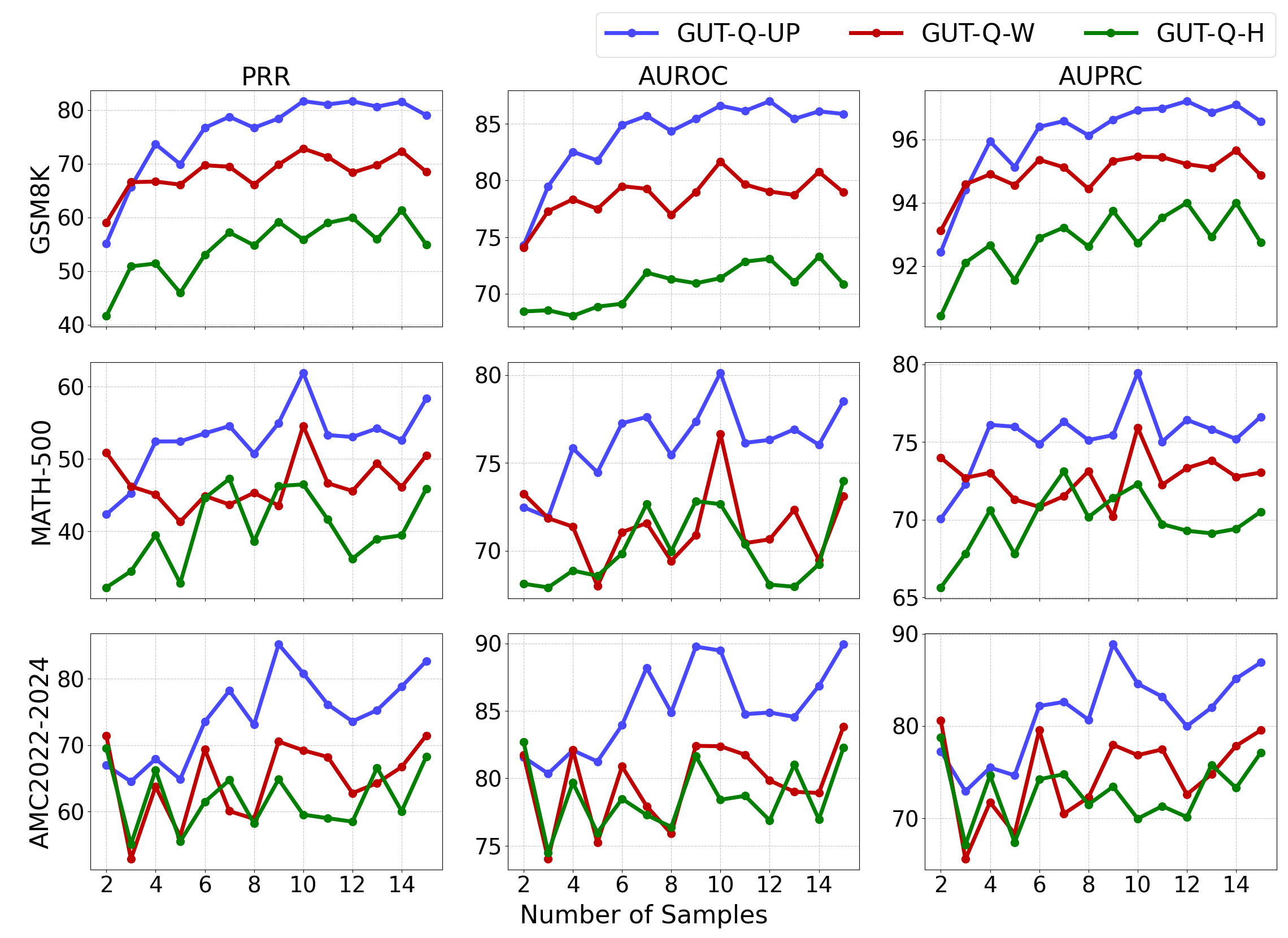}
        \caption{Impact of sample number $K$ on UQ performance.}
        \label{fig:num_of_samples_4B_maintext}
    \end{subfigure}
    \begin{subfigure}[b]{0.49\linewidth}
        \centering
        \includegraphics[width=\linewidth]{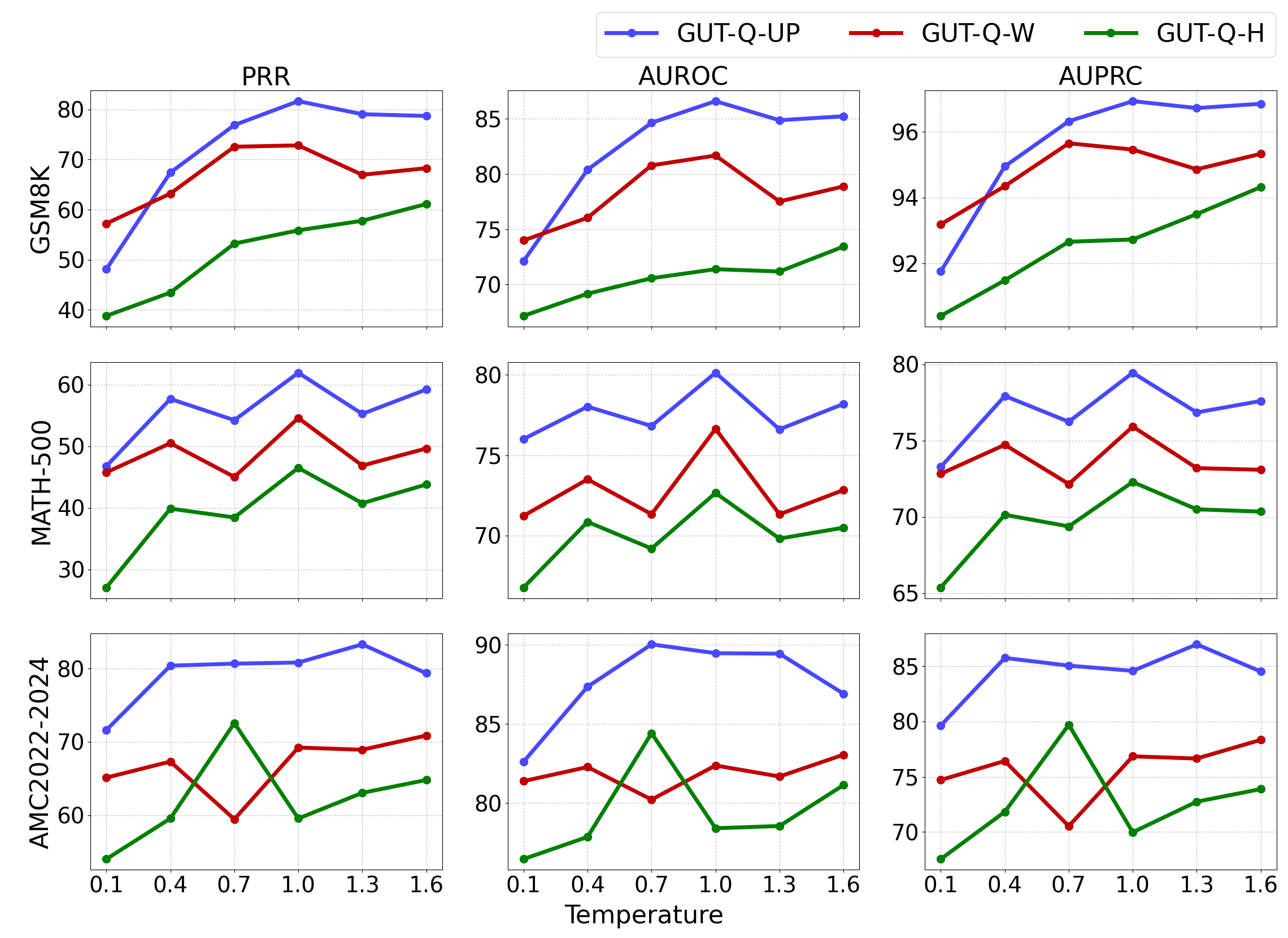}
        \caption{Impact of temperature $T$ on UQ performance.}
        \label{fig:temperature_4B_maintext}
    \end{subfigure}
    \hfill
    \begin{subfigure}[b]{0.49\linewidth}
        \centering
        \includegraphics[width=\linewidth]{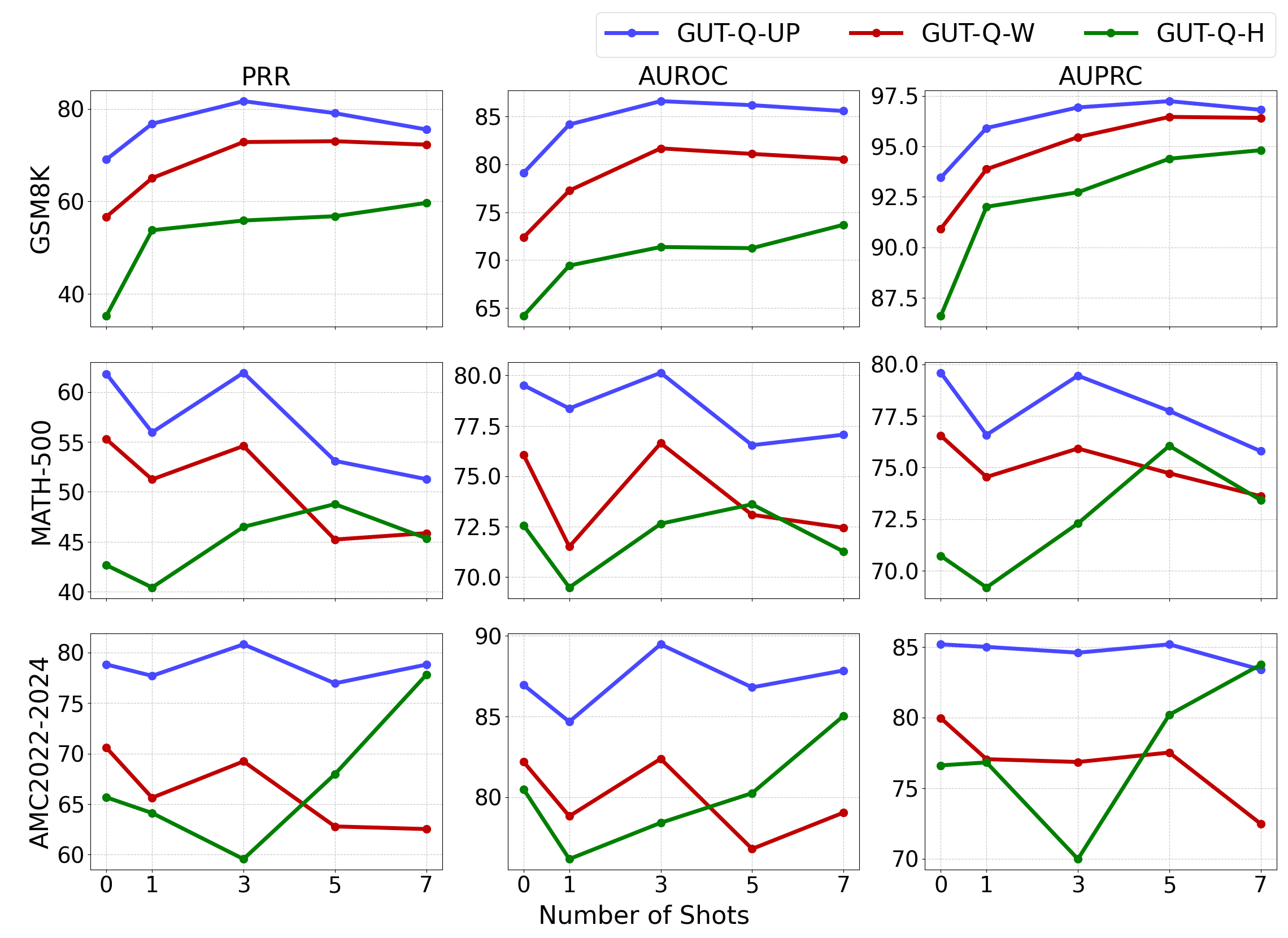}
        \caption{Impact of shot number $F$ on UQ performance.}
        \label{fig:num_of_shots_4B_maintext}
    \end{subfigure}
    \caption{Ablation and sensitivity analyses of the GUT-Q for Qwen3-4B.}
    \label{fig:ablation_sensitivity_4B_maintext}
\end{figure*}

\paragraph{Ablation Analyses} This paragraph analyzes how node uncertainty calculation and node merging affect the performance of GUT-Q. Figure~\ref{fig:ablation_4B_maintext} shows the ablation comparisons of UQ performance for Qwen3-4B across mathematical reasoning datasets. Results for all 4 concerned LLMs and 5 datasets are presented in Appendix~\ref{app:subsec:ablation}. It is observed that, compared to the blue bars, the red bars are closer while the green bars are far shorter for the UP methods, demonstrating that node merging is more important than node uncertainty in the UP algorithm. We also observe that the red and green bars are notably shorter than the blue bars for both the GUT-Q-W and GUT-Q-H, indicating that both node uncertainty and node merging are essential for GUT-Q-W and GUT-Q-H.

\paragraph{Sensitivity Analyses} This paragraph analyzes how UQ performance varies with the number of samples $K$, temperature $T$, and number of shots $F$. Figure~\ref{fig:num_of_samples_4B_maintext} illustrates the impact of the number of samples $K$ on UQ performance. We recommend $K=9$ to balance efficiency and performance since a larger $K$ leads to higher computational costs. Figures~\ref{fig:temperature_4B_maintext} and~\ref{fig:num_of_shots_4B_maintext} illustrate the impact of the temperature $T$ and the number of shots $F$ on UQ performance, respectively. We recommend $T=1.0$ and $F=5$ based on UQ performance. Appendix~\ref{app:subsec:sensitivity} provides complete sensitivity analyses for all four evaluated LLMs across five concerned datasets, as well as for the node merging criterion in Algorithm~\ref{alg:AOV}.

\subsection{Verifications on GUT-O} \label{subsec:UO_experiment}
This subsection validates the effectiveness of the proposed GUT-O module by comparing the mean uncertainty~\cite{ye2024benchmarking} of an LLM on a dataset, defined as $U(D;\theta)=\sum\nolimits_{x\in D}^{}{U( x ; \theta ) /|D|}$, before and after using GUT-O. A greater reduction in GUT-Q-derived mean uncertainty indicates better UO performance. We use GRPO as a contender.

\begin{table*}[ht]
  \footnotesize
  \centering
  \begin{tabular}{ccclll}
    \toprule
    \multirow{2}{*}{\textbf{Models}} & \multirow{2}{*}{\textbf{Accuracy}} & \multirow{2}{*}{\textbf{MTLP}} & \multicolumn{3}{c}{\textbf{Mean Uncertainty}} \\
    \cmidrule(lr){4-6}
                            &                           &                       & \multicolumn{1}{c}{\textbf{GUT-Q-W}} & \multicolumn{1}{c}{\textbf{GUT-Q-H}} & \multicolumn{1}{c}{\textbf{GUT-Q-UP}} \\
    \midrule
    Qwen3-0.6B              & $18.00 \pm 1.31$ & $-7.55 \pm 0.04$  & $0.77 \pm 0.01$          & $1.86 \pm 0.03$          & $1.14 \pm 0.01$ \\
    \quad + GRPO            & $23.40 \pm 1.51$ & $-7.67 \pm 0.04$  & $0.74 \pm 0.01$ (-3.9\%) & $1.88 \pm 0.03$ (+1.1\%) & $1.17 \pm 0.02$ (+2.6\%) \\
    \quad + GUT-O           & $21.40 \pm 1.43$ & $-8.94 \pm 0.04$  & $0.64 \pm 0.01$ (-16.9\%)& $1.51 \pm 0.03$ (-18.8\%)& $0.88 \pm 0.01$ (-22.8\%) \\
    \cmidrule{1-6}
    Qwen3-1.7B              & $42.20 \pm 1.79$ & $-13.45 \pm 0.06$ & $0.43 \pm 0.01$          & $2.03 \pm 0.05$          & $0.47 \pm 0.01$ \\
    \quad + GRPO            & $46.40 \pm 1.81$ & $-13.56 \pm 0.07$ & $0.42 \pm 0.01$ (-2.3\%) & $1.89 \pm 0.05$ (-6.9\%) & $0.46 \pm 0.01$ (-2.1\%) \\
    \quad + GUT-O           & $43.00 \pm 1.67$ & $-14.03 \pm 0.07$ & $0.39 \pm 0.01$ (-9.3\%) & $1.82 \pm 0.05$ (-10.3\%)& $0.43 \pm 0.01$ (-8.5\%) \\
    \cmidrule{1-6}
    Qwen3-4B                & $53.80 \pm 1.71$ & $-14.14 \pm 0.08$ & $0.37 \pm 0.01$          & $1.76 \pm 0.05$          & $0.40 \pm 0.01$ \\
    \quad + GRPO            & $61.00 \pm 1.68$ & $-14.11 \pm 0.09$ & $0.38 \pm 0.01$ (+2.7\%) & $1.78 \pm 0.04$ (+1.1\%) & $0.40 \pm 0.01$ (+0.0\%) \\
    \quad + GUT-O           & $57.00 \pm 1.71$ & $-14.25 \pm 0.07$ & $0.34 \pm 0.00$ (-8.1\%) & $1.70 \pm 0.04$ (-3.4\%) & $0.32 \pm 0.01$ (-20.0\%) \\
    \cmidrule{1-6}
    Qwen3-8B                & $64.20 \pm 1.68$ & $-15.14 \pm 0.09$ & $0.33 \pm 0.01$          & $1.46 \pm 0.04$          & $0.24 \pm 0.01$ \\
    \quad + GRPO            & $66.60 \pm 1.64$ & $-14.97 \pm 0.09$ & $0.33 \pm 0.01$ (+0.0\%) & $1.43 \pm 0.04$ (-2.1\%) & $0.25 \pm 0.01$ (+4.2\%) \\
    \quad + GUT-O           & $64.80 \pm 1.72$ & $-15.78 \pm 0.08$ & $0.31 \pm 0.01$ (-6.1\%) & $1.38 \pm 0.04$ (-5.5\%) & $0.23 \pm 0.01$ (-4.2\%) \\
    \bottomrule
  \end{tabular}
  \caption{Comparisons of UO performance for 4 Qwen3 LLMs on the MATH-500 dataset, where `+' in column ``Models'' indicates the applied fine-tuning method, and values in parentheses denote the percentage change in mean uncertainty.}
  \label{tab:UO_performance_math}
\end{table*}

Table~\ref{tab:UO_performance_math} shows the comparisons of UO performance for 4 Qwen3 LLMs on the MATH-500 dataset, where `+' indicates the applied fine-tuning method, and values in parentheses indicate the percentage change in mean uncertainty. There are two key observations. First, after employing GUT-O, all three GUT-Q-derived mean uncertainties decrease while the accuracy improves. This observation validates the effectiveness of GUT-O. Second, it is observed that mean uncertainties are typically comparable or even increase after applying GRPO, which suggests that the UO performance of GUT-O primarily benefits from the MTLP optimization target rather than task-specific fine-tuning. Similar observations hold for all 4 evaluated LLMs and 5 datasets, as shown in Table~\ref{tab:UO_performance_app} in Appendix~\ref{app:subsec:uncertainty_reduction}. These findings demonstrate that GUT-O separately reduces reasoning uncertainty and improves accuracy by an average of 13.95\% and 1.93\% across 4 LLMs and 5 datasets, which answers Q2.

\section{Conclusions}  \label{sec:conclusion}
In this paper, we investigated the reasoning uncertainty of LLMs by using DAGs to characterize the potential branches of each reasoning step. Built upon this recognition, we proposed the GUT comprising GUT-Q and GUT-O for quantifying and reducing reasoning uncertainty, respectively. The GUT-Q quantifies reasoning uncertainty by the reasoning space complexity, which is approximated by the complexity of the DAG. Defining the negative uncertainty proxy as the reward in the GRPO method, the GUT-O further reduces reasoning uncertainty. Empirical results showed that GUT-Q-UP surpasses the best UQ contenders by an average of 11.79\% in PRR, 13.33\% in AUROC, and 9.66\% in AUPRC. GUT-O reduces reasoning uncertainty and improves accuracy by an average of 13.95\% and 1.93\%, respectively.

\section*{Acknowledgments}
The research was supported by the Natural Science Foundation of China (62406138).

\appendix
\onecolumn

\section*{Appendix}
This appendix provides the supplementary materials for our work, constructed according to the corresponding sections therein. Before that, we review the origin of reasoning uncertainty in LLMs, which has been illustrated in Figure~\ref{fig:overview} and introduced in Section~\ref{sec:UQ_method}. Given a problem, the stochasticity of the LLM system originates from the token-level distribution. The autoregressively stochastic temperature sampling over this distribution yields variability in the output reasoning chains, which ultimately gives rise to reasoning uncertainty. This work focuses on quantifying and optimizing reasoning uncertainty.

\section{Additional Implementation Details of GUT}  \label{app:additional_details_MAPS}

\subsection{Prompt Templates}  \label{app:subsec:prompt_template}
This subsection details the prompt templates used for the CoT sampling in the GUT-Q module and UQ contenders. Due to space constraints, we only list the templates of 0-shot and 1-shot here.

\begin{promptbox}[title=0-shot Prompt]
Answer the following problem. Break down your reasoning process into small steps. Each step should represent a single, minimal reasoning action, and each step must logically follow the previous one. Use the following format for each step: \\
Step i: [Your reasoning process in one cohesive response] \\
After completing all the steps, conclude with:\\
Final Answer: \textbackslash boxed{[Your final answer here without the unit or any additional text]} \#\#\#\# \\
Stop generation immediately after outputting the Final Answer. Ensure that your response strictly follows the format to maintain clarity and consistency.

\vspace{0.5em} 

\textbf{Problem:} [INPUT\_PROBLEM] \\
\textbf{Reasoning:}
\end{promptbox}

\vspace{0.3cm}

\begin{promptbox}[title=1-shot Prompt]
Answer the following problem. Break down your reasoning process into small steps. Each step should represent a single, minimal reasoning action, and each step must logically follow the previous one. Use the following format for each step: \\
Step i: [Your reasoning process in one cohesive response] \\
After completing all the steps, conclude with: \\
Final Answer: \textbackslash boxed\{[Your final answer here without the unit or any additional text]\} \#\#\#\# \\
Stop generation immediately after outputting the Final Answer.

The following is an example:

\vspace{0.5em} 

\textbf{Problem:} Let $M$ be the midpoint of $\overline{AB}$ in regular tetrahedron $ABCD$. $\frac{p}{q}=\cos(\angle CMD)$ is irreducible fraction, what is the value of $p+q$? \\
\textbf{Reasoning:} \\
Step 1: Without loss of generality, let the edge-length of $ABCD$ be $2.$ It follows that $MC=MD=\sqrt3.$ \\
Step 2: Let $O$ be the center of $\triangle ABD,$ so $\overline{CO}\perp\overline{MOD}.$ Note that $MO=\frac13 MD=\frac{\sqrt{3}}{3}.$ \\
Step 3: In right $\triangle CMO,$ we have $\cos(\angle CMD)=\frac{MO}{MC}=\frac13.$ \\
Step 4: So the answer is $1+3=4.$ \\
Final Answer:  \textbackslash boxed\{4\} \#\#\#\# \\

\vspace{0.5em} 

\textbf{Problem:} [INPUT\_PROBLEM] \\
\textbf{Reasoning:}
\end{promptbox}

For the GSM8K and MATH-500 datasets, we use 7 instances from the training sets of GSM8K and MATH~\citep{hendrycks2021measuring}, respectively, to construct the few-shot prompts. For the AMC2022-2024 dataset, we use 7 instances from the AMC2022-2023\footnote{https://huggingface.co/datasets/AI-MO/aimo-validation-amc} split to construct the few-shot prompts and remove these instances from the constructed AMC2022-2024 dataset. We append $\#\#\#\#$ after ``Final Answer'' to serve as a stopping string for efficient sampling. By default, we follow~\citep{zhang2025all} and use a 3-shot prompt to extract the reasoning steps along with the final answer. We also conduct sensitivity analyses on the number of shots in Subsection~\ref{subsec:UQ_experiment} and Appendix~\ref{app:subsec:sensitivity}.


\subsection{Details of GUT-Q Algorithm}  \label{app:subsec:algorithm}
This subsection first lists several algorithms regarding graph complexity estimation proposed in Subsection~\ref{subsec:graph_complexity_estimation}, and then provides the pseudocode of the graph construction procedure in GUT-Q. 

\begin{algorithm}[ht]
\caption{GUT-Q-H}
\label{alg:height_prob}
\textbf{Input:} DAG $G=(V,E)$, node uncertainty $\{ U(v) \}_{v\in V}$ \\
\textbf{Output:} Uncertainty-weighted height $H_u$ \\
\textbf{Procedures:}
\begin{algorithmic}[1]
\STATE $L \gets \text{TopologicalSort}(G)$ 
\STATE $R[v] \gets 0$ for all $v \in V$ \COMMENT{Initialize topological level}
\FOR{$u \in L$}
    \STATE $R[u] \gets \max(\{R[p] + 1 \mid (p, u) \in E\} \cup \{0\}) $
\ENDFOR
\STATE $V_h \gets \text{GetHeightNodes}(G, R)$ \COMMENT{Get height-path nodes}
\STATE $H_u \gets \sum_{v \in V_h} U(v)$ 
\end{algorithmic}
\end{algorithm}

Algorithm~\ref{alg:height_prob} lists the GUT-Q-H graph complexity method introduced in Subsection~\ref{subsec:graph_complexity_estimation}, where its output $H_u$ serves as a measure of graph complexity and is further used to quantify reasoning uncertainty.

\begin{algorithm}[ht]
\caption{GUT-Q-UP-Weights}
\label{alg:DAG_UP_weights}
\textbf{Input:} DAG $G=(V,E)$, node uncertainty $\{U(v)\}_{v \in V}$, bias $\rho \in \mathbb{R}^+$, activation function $\phi$ \\
\textbf{Output:} Auxiliary node uncertainty $U(s_a)$\\
\textbf{Procedures:}
\begin{algorithmic}[1]
\STATE Construct augmented graph $G'=(V', E')$ by connecting all leaf nodes of $G$ to a new auxiliary node $s_a$
\STATE Construct FNN corresponding to $G'$ with $\phi$
\STATE Set weights $w_{uv} \gets U(v)$ for all $(u, v) \in E'$
\STATE Set biases $b_v \gets \rho$ for all $v \in V$, and $b_{s_a} \gets 0$
\STATE $U(s_a) \gets$ output of $s_a$ from forward propagation
\end{algorithmic}
\end{algorithm}

Algorithm~\ref{alg:DAG_UP_weights} is a variant of Algorithm~\ref{alg:DAG_UP}. The key idea is to initialize the weights of the FNN with the node uncertainty and the biases with a hyperparameter $\rho \in \mathbb{R}^+$. This variant is dual to Algorithm~\ref{alg:DAG_UP}, which initializes the biases with the node uncertainty and the weights with the hyperparameter $\omega \in \mathbb{R}^+$. We note that the UQ performance of Algorithms~\ref{alg:DAG_UP} and~\ref{alg:DAG_UP_weights} is comparable across all concerned LLM scales, datasets, and UQ evaluation metrics, as shown in Table~\ref{tab:UP_comparison_app}.

\begin{table*}[ht]  
  \centering
  \footnotesize
  \setlength{\tabcolsep}{2.4pt} 
  \resizebox{\linewidth}{!}{%
    \begin{tabular}{llrrrrrrrrr}
      \toprule
      \multirow{2}{*}[-0.8ex]{\textbf{Scale}} & \multirow{2}{*}[-0.8ex]{\textbf{UQ}} & \multicolumn{3}{c}{\textbf{GSM8K}} & \multicolumn{3}{c}{\textbf{MATH-500}} & \multicolumn{3}{c}{\textbf{AMC2022-2024}} \\
      \cmidrule(lr){3-5} \cmidrule(lr){6-8} \cmidrule(lr){9-11}
        &      & \multicolumn{1}{c}{\textbf{PRR}} & \multicolumn{1}{c}{\textbf{AUROC}} & \multicolumn{1}{c}{\textbf{AUPRC}} & \multicolumn{1}{c}{\textbf{PRR}} & \multicolumn{1}{c}{\textbf{AUROC}} & \multicolumn{1}{c}{\textbf{AUPRC}} & \multicolumn{1}{c}{\textbf{PRR}} & \multicolumn{1}{c}{\textbf{AUROC}} & \multicolumn{1}{c}{\textbf{AUPRC}} \\
      \midrule
      \multirow{2}{*}{\textbf{8B}}
        & GUT-Q-UP & \underline{72.01$\pm$4.34} & \underline{83.72$\pm$1.55} & \underline{97.01$\pm$0.50} & \textbf{53.28$\pm$4.34} & \underline{75.93$\pm$1.90} & \textbf{81.59$\pm$1.87} & \underline{72.85$\pm$4.89} & \underline{81.83$\pm$3.02} & \textbf{84.53$\pm$2.88} \\
        & GUT-Q-UP-W & \textbf{73.91$\pm$4.19} & \textbf{84.42$\pm$1.60} & \textbf{97.39$\pm$0.46} & \underline{52.77$\pm$4.57} & \textbf{76.64$\pm$1.76} & \underline{79.34$\pm$2.11} & \textbf{74.68$\pm$6.32} & \textbf{87.35$\pm$2.54} & \underline{82.27$\pm$4.55} \\
      \midrule
      \multirow{2}{*}{\textbf{4B}}
        & GUT-Q-UP   & \underline{81.68$\pm$1.83}       & \underline{86.61$\pm$1.02}         & \underline{96.93$\pm$0.36}         & \underline{61.93$\pm$3.59}       & \underline{80.13$\pm$1.57}         & \underline{79.45$\pm$2.07}         & \textbf{80.82$\pm$4.24}          & \textbf{89.48$\pm$2.19}            & \textbf{84.60$\pm$3.39}            \\
        & GUT-Q-UP-W & \textbf{81.76$\pm$1.79}          & \textbf{86.63$\pm$1.00}            & \textbf{96.94$\pm$0.35}            & \textbf{62.13$\pm$3.60}          & \textbf{80.23$\pm$1.58}            & \textbf{79.56$\pm$2.15}            & \underline{79.07$\pm$4.50}       & \underline{88.72$\pm$2.19}         & \underline{83.24$\pm$3.72}         \\
      \midrule
      \multirow{2}{*}{\textbf{1.7B}}
        & GUT-Q-UP   & \textbf{77.50$\pm$1.83}          & \underline{85.48$\pm$0.86}         & \textbf{92.83$\pm$0.67}            & \underline{63.39$\pm$3.36}       & \underline{80.68$\pm$1.50}         & \underline{74.08$\pm$2.45}         & \textbf{68.00$\pm$6.93}          & \textbf{82.90$\pm$3.72}            & \textbf{68.00$\pm$6.30}            \\
        & GUT-Q-UP-W & \underline{77.29$\pm$1.92}       & \textbf{85.49$\pm$0.86}            & \underline{92.76$\pm$0.70}         & \textbf{63.95$\pm$3.36}          & \textbf{81.13$\pm$1.51}            & \textbf{74.39$\pm$2.44}            & \underline{66.56$\pm$7.22}       & \underline{81.68$\pm$3.90}         & \underline{66.99$\pm$6.43}         \\
      \midrule
      \multirow{2}{*}{\textbf{0.6B}}
        & GUT-Q-UP   & \underline{52.02$\pm$2.51}       & \underline{75.31$\pm$1.20}         & \underline{57.57$\pm$2.06}         & \textbf{52.04$\pm$4.58}          & \textbf{76.50$\pm$2.31}            & \textbf{48.71$\pm$4.35}            & \underline{14.75$\pm$8.38}       & \textbf{67.52$\pm$nan}             & \underline{15.48$\pm$4.79}         \\
        & GUT-Q-UP-W & \textbf{52.17$\pm$2.51}          & \textbf{75.45$\pm$1.20}            & \textbf{57.64$\pm$2.06}            & \underline{50.37$\pm$4.44}       & \underline{75.05$\pm$2.30}         & \underline{48.14$\pm$4.33}         & \textbf{14.90$\pm$8.74}          & \underline{67.11$\pm$nan}          & \textbf{15.59$\pm$5.05}            \\
      \bottomrule
    \end{tabular}}
      \caption{Comparisons of UQ performance between the GUT-Q implemented with Algorithms~\ref{alg:DAG_UP} (GUT-Q-UP) or~\ref{alg:DAG_UP_weights} (GUT-Q-UP-W).}
  \label{tab:UP_comparison_app}
\end{table*}

Algorithm~\ref{alg:AOV} summarizes the DAG construction procedure in Subsection~\ref{subsec:graph_construction}.

\begin{algorithm}[ht]
\caption{DAG Construction}
\label{alg:AOV}
\textbf{Input:} problem $x$, sampling times $K$, temperature $T$ \\
\textbf{Output:} DAG $G=(V, E)$ \\
\textbf{Procedures:}
\begin{algorithmic} [1]
\STATE Initialize $G=(V, E)$ with root node $x$
\FOR {$i \in [K]$}
    \STATE Sample a reasoning chain $s^i$ at temperature $T$
    \STATE Extract reasoning steps $\{s_j^i\}_{j \in [n_i]}$ from $s^i$
    \STATE Calculate step-level uncertainty $\{U(s_j^i)\}_{j \in [n_i]}$
    \STATE Initialize $UC(s_j^i) \gets \{U(s_j^i)\}$ for all $j \in [n_i]$
    \STATE $V \gets V \cup \{s_j^i\}_{j \in [n_i]}$ \ , \ $E \gets E \cup \{(x, s_1^i)\}$
    \STATE $E \gets E \cup \{(s_j^i, s_{j+1}^i)\}_{j \in [n_i-1]}$
\ENDFOR
\STATE Initialize processed node set $V_p \gets \emptyset$  \COMMENT{Node merging}
\FOR {$i \in [K]$}
    \STATE Initialize node set $V_m \gets V_p$ for merging \COMMENT{AOV-based traversal}
    \STATE $V_p \gets V_p \cup \{s_j^i\}_{j \in [n_i]}$
    \FOR {$j \in [n_i]$}
        \FOR {each node $v \in V_m$}
            \IF {NLI predicts bi-entailment for $v$ and $s_j^i$}
                \STATE $UC(v) \gets UC(v) \cup UC(s_j^i)$
                \STATE Get the parent $p$ of $s_j^i$ where $(p, s_j^i) \in E$
                \STATE $E \gets E \cup \{(p, v), (v, c) \mid (s_{ij} , c) \in E \}$
                \STATE $V_p \gets V_p \setminus \{s_j^i\}$ \ , \ $V_m \gets V_m \setminus \{v\}$
                \STATE $V_m \gets V_m \setminus \{u\}$ for all ancestor nodes $u$ of $v$
                \STATE \textbf{break}
            \ENDIF
        \ENDFOR
    \ENDFOR
\ENDFOR
\STATE $U(v) \gets \text{Average}(UC(v))$ for all $v \in V_{p}$
\STATE Update $G$ to retain only nodes in $V_p \cup \{ x \}$
\end{algorithmic}
\end{algorithm}

\subsection{Details of GUT-O Formulation}  \label{app:subsec:UO_implementation}
This subsection formally introduces GUT-O. We start with the calculation of the MTLP, which is a special case of the step-level uncertainty calculation in Subsection~\ref{subsec:steplevel_uq}. Specifically, we compute the ``Avg Log Prob.'' for each token in $c$, and set $w=1$ and $d=100$ to perform the Top-$d\%$ aggregation over $c$ to yield $U(x, c)$. 

Before listing the optimization procedures, we introduce some useful notations. Let $\text{clip}(a,\epsilon)=\max (1-\epsilon , \min(a, 1+\epsilon))$ for $\epsilon \in (0, 1)$ and $a \in \mathbb{R}$. We denote the current, old, and reference policy models as $\pi_{\theta}$, $\pi_{\theta_{\text{old}}}$, and $\pi_{\text{ref}}$, respectively. With the reward defined as the negative uncertainty proxy $r^i = -U(x, c^i)$ and $M \in \mathbb{N}^+$, the GUT-O optimizes $\pi_{\theta}$ by maximizing
\[
    \mathcal{J}(\theta) = \mathbb{E}_{x \sim \mathcal{X}, \{c^i\}_{i\in[M]} \sim \pi_{\theta_{\text{old}}}( \cdot |x)} \frac{1}{M} \sum_{i=1}^M \frac{1}{|c^i|} \sum_{j=1}^{|c^i|} \left\{ \min \left[ l^i_j(\theta) , \  \text{clip}\left( l^i_j(\theta), \epsilon \right)  \right] A^i - \beta \text{KL} \left[\pi_\theta || \pi_\text{ref}\right] \right\} \ ,
\]
where
\[
    l^i_j(\theta)=\pi _{\theta}(t^i_j|h^i_j) ~/~ \pi _{\theta _\text{old}}(t^i_j|h^i_j) , \
        A^i= \left[ r^i-\text{mean}\left( \left\{ r^i \right\} _{i\in \left[ M \right]} \right) \right] ~/~ \text{std}\left( \left\{ r^i \right\} _{i\in \left[ M \right]} \right)  \ ,
\]
and
\[
    \text{KL}\left[ \pi _{\theta}||\pi _\text{ref} \right] = \pi _{\text{ref}}(t^i_j\mid h^i_{<j}) ~/~ \pi _{\theta}(t^i_j\mid h^i_{<j}) -\log \pi _{\text{ref}}(t^i_j\mid h^i_{<j}) ~/~ \pi _{\theta}(t^i_j\mid h^i_{<j}) - 1 \ .
\]
Within the optimization target $\mathcal{J}(\theta)$, the first term represents the clipped surrogate advantage. By leveraging the normalized advantage $A^i$, it encourages the model to increase the probability of generating reasoning paths that yield higher rewards, i.e., lower MTLP. Meanwhile, the clipping mechanism ensures stable training by restricting overly large policy updates. The second term introduces a Kullback-Leibler (KL) divergence penalty, scaled by the coefficient $\beta \in \mathbb{R}$. This term acts as a regularizer to prevent the active policy $\pi_\theta$ from deviating too drastically from the reference model $\pi_{\text{ref}}$, thereby mitigating reward hacking and preserving the fundamental language capabilities of the LLM.

\clearpage

\section{Additional Experimental Details on GUT-Q}  \label{app:UQ_additional_exp}
This appendix provides additional experimental results and implementation details on the proposed GUT-Q module.

\subsection{Details on Datasets}  \label{app:subsec:dataset}
In this work, we conducted experiments on mathematical reasoning benchmarks across three difficulty levels, ordered from easy to hard as follows. GSM8K contains 8500 grade school math word problems. We use the test set of GSM8K, which contains 1319 problems. MATH-500~\citep{lightman2024lets} consists of 500 high school math problems. AMC2022-2024 comprises 128 high school math competition problems, constructed by aggregating the AMC2022-2023\footnote{https://huggingface.co/datasets/AI-MO/aimo-validation-amc} and AMC2024\footnote{https://huggingface.co/datasets/rawsh/2024\_AMC12}. All problems are fill-in-the-blank, where the final answer is typically a real number or a real vector.

The difficulty level is characterized by the accuracy. In our implementation, we use the Math-Verify\footnote{https://github.com/huggingface/Math-Verify} package to verify the correctness of the extracted final answer. Table~\ref{tab:acc_app} shows the accuracy of the evaluated LLMs on the concerned datasets. It is observed that the accuracy decreases sequentially across GSM8K, MATH-500, and AMC2022-2024 for all LLM scales, indicating that the difficulty level increases sequentially across GSM8K, MATH-500, and AMC2022-2024.
\begin{table}[ht]
  \centering
  \footnotesize
    \begin{tabular}{cccccc}
    \toprule
     \multirow{2}{*}{Scale} & \multicolumn{3}{c}{\textbf{Mathematical Reasoning}} & \textbf{First-Order Logic Reasoning} & \textbf{Reasoning QA} \\
     \cmidrule(lr){2-4} \cmidrule(lr){5-5} \cmidrule(lr){6-6}
     & GSM8K & MATH-500 & AMC2022-2024 & FOLIO & MMLU-Pro\\
    \midrule
    8B             & 89.61          & 64.20             & 49.59                 & 72.41          & 69.58 \\
    4B             & 84.08          & 53.80             & 38.02                 & 68.47          & 60.83 \\
    1.7B           & 71.11          & 42.20             & 22.31                 & 43.84          & 43.33 \\
    0.6B           & 28.89          & 18.00             & \phantom{0}9.09       & 44.83          & 29.17 \\
    \bottomrule
  \end{tabular}
  \caption{Accuracy of the evaluated LLMs on the concerned datasets.}
  \label{tab:acc_app}
\end{table}

We note that the presented accuracy may differ from that of~\citep{an2025qwen}, since (1) we employ a different prompt template to extract the reasoning steps and final answers for UQ, and (2) we use a smaller token generation budget, as detailed in Appendix~\ref{app:subsec:UQ_hyperparameter}, due to the constraints of computational resources. However, this work mainly focuses on quantifying and optimizing the reasoning uncertainty, instead of improving the accuracy. Therefore, the accuracy differences between ours and that of~\citep{an2025qwen} are acceptable.

We also note that the low accuracy of Qwen3-0.6B on the AMC2022-2024 dataset often leads to single-class bootstrapping where all samples are incorrect, which results in the nan values of std in Tables~\ref{tab:UP_comparison_app} and~\ref{tab:UQ_0.6B_app} and missing std bars in Figure~\ref{fig:ablation_0.6B_app} since the AUROC is undefined for binary classification tasks containing only a single class.

Moreover, we conducted experiments on first-order logic reasoning tasks such as FOLIO~\citep{han2024folio} and challenging long-form QA tasks that may require reasoning like MMLU-Pro~\citep{wang2024mmlu}, to examine the generality of the proposed GUT. The instances of these two datasets are multiple-choice questions, where the final answer of FOLIO is within \{True, False, Uncertain\}, and that of MMLU-Pro is within \{A, B, C, D\}.

\subsection{Configurations of Hyperparameters}  \label{app:subsec:UQ_hyperparameter}
This subsection details configurations of hyperparameters in the GUT-Q module.

\paragraph{Configuration of the Step-level Uncertainty Calculation} Following~\citet{fu2025deep}, we set the hyperparameters for the ``step-level uncertainty calculation'' stage of GUT-Q as shown in Table~\ref{tab:hyperparams_of_steplevel}.

\begin{table}[htbp]
  \centering
  \begin{tabular}{lc}
    \toprule
    \textbf{Hyperparameter}                       & \textbf{Setting} \\
    \midrule
    Token-level UQ                                & Avg Log Prob      \\
    Selection Strategy                            & Top-3\%           \\
    Window Size $w$                               & 6                 \\
    Weight $\omega$ in Algorithm~\ref{alg:DAG_UP} & 0.2               \\
    \bottomrule
  \end{tabular}
    \caption{Configuration of hyperparameters in step-level uncertainty calculation.}
  \label{tab:hyperparams_of_steplevel}
\end{table}

\paragraph{Configuration of Sampling Parameters} Following prior work~\citep{manakul2023selfcheckgpt,farquhar2024detecting,qiu2024semantic}, we use different sampling setups for evaluating correctness and quantifying uncertainty.
\begin{itemize}
    \item We follow the official Qwen3 recommendation\footnote{https://huggingface.co/Qwen/Qwen3-4B} using $T=0.7$, Top-$k=20$, and Top-$p=0.8$ to obtain the reasoning chain which is compared to the reference answer. This sample determines whether the problem is correctly answered.
    \item We follow LM-Polygraph\footnote{https://github.com/IINemo/lm-polygraph}~\citep{vashurin2025benchmarking} and set $T=1.0$ and $K=10$ to obtain the candidate reasoning chains for ensuring a fair comparison with the UQ contenders. We also follow the official Qwen3 recommendation to set Top-$k=20$ and Top-$p=0.8$.
\end{itemize}

\begin{figure}[htb]
	\centering
	\includegraphics[width=0.8\linewidth]{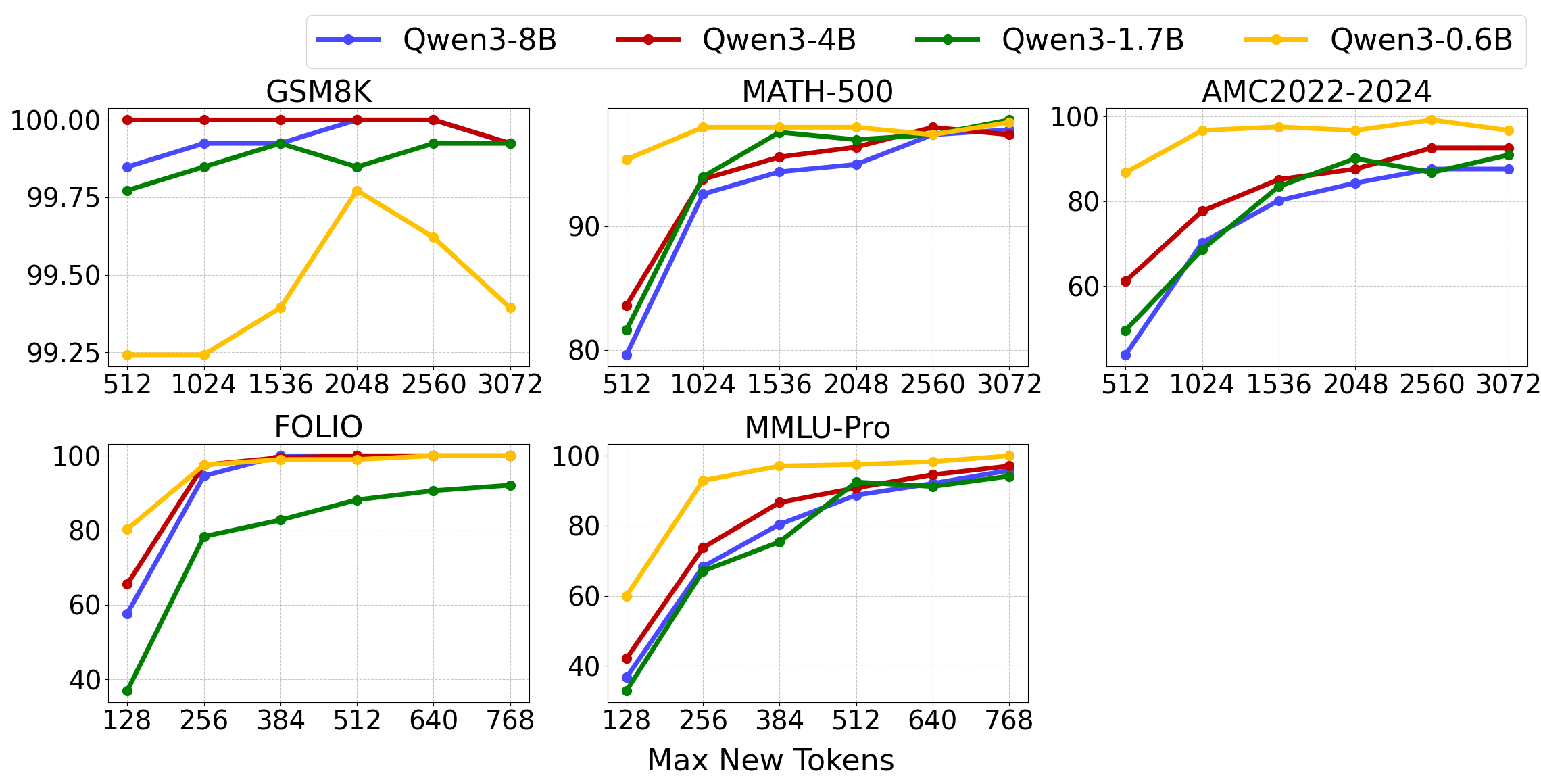}
	\caption{Complete generation rate.}
    \label{fig:complete_gen_rate_app}
\end{figure}

We conducted sensitivity analyses to the max\_new\_tokens hyperparameter for the concerned datasets. The max\_new\_tokens controls the maximum number of tokens that the LLM can generate. Hence, a larger max\_new\_tokens indicates higher computational cost. However, a lower max\_new\_tokens may lead to incomplete generations, such as missing the ``Final Answer'' that is required in our prompt template provided in Appendix~\ref{app:subsec:prompt_template}. Therefore, there is a trade-off between the max\_new\_tokens and the complete generation rate. Figure~\ref{fig:complete_gen_rate_app} shows the complete generation rate versus the max\_new\_tokens. We set max\_new\_tokens = 256 for FOLIO, max\_new\_tokens = 512 for GSM8K and MMLU-Pro, max\_new\_tokens = 1536 for MATH-500, and max\_new\_tokens = 2048 for AMC2022-2024 to achieve the concerned trade-off across all LLM scales.

\subsection{UQ Contenders}  \label{app:subsec:UQ_contenders}
Table~\ref{tab:abbr_of_ue_methods} lists the full names and corresponding abbreviations (Abbr.) of the 45 UQ contenders. We implemented CoT-UQ\footnote{https://github.com/ZBox1005/CoT-UQ}~\citep{zhang2025cot} and TopoUQ\footnote{https://github.com/LongchaoDa/LLM-Topology}~\citep{da2025understanding} using their official repositories, respectively, and the remaining contenders via LM-Polygraph\footnote{https://github.com/IINemo/lm-polygraph}~\citep{vashurin2025benchmarking}.

\begin{table}[htbp]
  \centering
  \footnotesize 
  \resizebox{\linewidth}{!}{$
  \begin{tabular}{lllll}
          \toprule
          \textbf{Abbr.} & \textbf{Full Name}                       & \textbf{Type}               & \textbf{Category}                   & \textbf{Paper}                  \\
          \midrule
          MSP            & Maximum Sequence Probability             & \multirow{27}{*}{White-box} & \multirow{14}{*}{Information-based} & \citet{fadeeva2023msp}          \\
          Ppl            & Perplexity                               &                             &                                     & \citet{fomicheva2020Ppl}        \\
          MTE            & Mean Token Entropy                       &                             &                                     & \citet{fomicheva2020Ppl}        \\
          MTLP           & Mean Token Log Probability               &                             &                                     & \citet{manakul2023selfcheckgpt} \\
          MPMI           & Mean Pointwise Mutual Information        &                             &                                     & \citet{takayama2019pmi}         \\
          CPMI           & Conditional Pointwise Mutual Information &                             &                                     & \citet{van2022cmpi}             \\
          RD             & R\'enyi Divergence                       &                             &                                     & \citet{darrin2023renyi}         \\
          FRD            & Fisher-Rao Distance                      &                             &                                     & \citet{darrin2023renyi}         \\
          AS             & Attention Score                          &                             &                                     & \citet{gaurang2024as}           \\
          CSL            & Contextualized Sequence Likelihood       &                             &                                     & \citet{lin2024csl}              \\
          MCSE           & Monte Carlo Sequence Entropy             &                             &                                     & \citet{malinin2021mcse}         \\
          MCNSE          & Monte Carlo Normalized Sequence Entropy  &                             &                                     & \citet{malinin2021mcse}         \\
          RAUQ           & Recurrent Attention-based UQ             &                             &                                     & \citet{vazhentsev2025rauq}      \\
          RAUQ-E         & Recurrent Attention-based UQ-Entropy     &                             &                                     & \citet{vazhentsev2025rauq}      \\[0.3em]
          SE             & Semantic Entropy                         &                             & \multirow{9}{*}{Diversity-based}    & \citet{farquhar2024detecting}   \\
          SAR            & Shifting Attention to more Relevant      &                             &                                     & \citet{duan2024sar}             \\
          TSAR           & TokenSAR                                 &                             &                                     & \citet{duan2024sar}             \\
          SSAR           & SequenceSAR                              &                             &                                     & \citet{duan2024sar}             \\
          SD             & Semantic Density                         &                             &                                     & \citet{qiu2024semantic}         \\
          ES             & Eigen Score                              &                             &                                     & \citet{chen2024es}              \\
          CoCoA-MSP      & CoCoA-Mean Sequence Probability          &                             &                                     & \citet{vashurin2025cocoa}       \\
          CoCoA-Ppl      & CoCoA-Perplexity                         &                             &                                     & \citet{vashurin2025cocoa}       \\
          CoCoA-MTE      & CoCoA-Mean Token Entropy                 &                             &                                     & \citet{vashurin2025cocoa}       \\[0.3em]
          P(True)        & P(True)                                  &                             & Reflexive-based                     & \citet{kadavath2022ptrue}       \\[0.3em]
          CoT-UQ-ME      & CoT enhanced UQ-probas MEan              &                             & \multirow{3}{*}{Reasoning-enhanced} & \citet{zhang2025cot}            \\
          CoT-UQ-MI      & CoT enhanced UQ-probas MIn               &                             &                                     & \citet{zhang2025cot}            \\
          CoT-UQ-SAR     & CoT enhanced UQ-TokenSAR                 &                             &                                     & \citet{zhang2025cot}            \\
          \midrule
          NS             & Number of Sets                           & \multirow{17}{*}{Black-box} & \multirow{16}{*}{Diversity-based}   & \citet{lin2024generating}       \\
          Eig-E          & Sum of Eigenvalues-NLI Score Entail.     &                             &                                     & \citet{lin2024generating}       \\
          Eig-C          & Sum of Eigenvalues-NLI Score Contra.     &                             &                                     & \citet{lin2024generating}       \\
          Eig-J          & Sum of Eigenvalues-Jaccard Score         &                             &                                     & \citet{lin2024generating}       \\
          Deg-E          & Degree Matrix-NLI Score Entail.          &                             &                                     & \citet{lin2024generating}       \\
          Deg-C          & Degree Matrix-NLI Score Contra.          &                             &                                     & \citet{lin2024generating}       \\
          Deg-J          & Degree Matrix-Jaccard Score              &                             &                                     & \citet{lin2024generating}       \\
          Ecc-E          & Eccentricity-NLI Score Entail.           &                             &                                     & \citet{lin2024generating}       \\
          Ecc-C          & Eccentricity-NLI Score Contra.           &                             &                                     & \citet{lin2024generating}       \\
          Ecc-J          & Eccentricity-Jaccard Score               &                             &                                     & \citet{lin2024generating}       \\
          LS-R1          & Lexical Similarity-Rouge 1               &                             &                                     & \citet{farquhar2024detecting}   \\
          LS-R2          & Lexical Similarity-Rouge 2               &                             &                                     & \citet{farquhar2024detecting}   \\
          LS-RL          & Lexical Similarity-Rouge L               &                             &                                     & \citet{farquhar2024detecting}   \\
          LS-B           & Lexical Similarity-BLEU                  &                             &                                     & \citet{farquhar2024detecting}   \\
          KLE            & Kernel Language Entropy                  &                             &                                     & \citet{nikitin2024KLE}          \\
          LUQ            & Long-text Uncertainty Quantification     &                             &                                     & \citet{zhang2024luq}            \\[0.3em]
          CoT-UQ-SP      & CoT enhanced UQ-Self Probing             &                             & Reasoning-enhanced                  & \citet{zhang2025cot}            \\[0.3em]
          Topo-UQ     & Topology-based UQ                 &                             &   Graph-based                                 & \citet{da2025understanding}            \\
          \bottomrule
        \end{tabular}
  $}
    \caption{Overview of the investigated UQ contenders.}
    \label{tab:abbr_of_ue_methods}
\end{table}

\clearpage
\subsection{Runtime Complexity Analysis of GUT-Q and Its Contenders} \label{app:subsec:runtime}
This subsection provides a runtime complexity analysis of GUT-Q and its contenders. Specifically, we select eight representative methods, namely Ppl, MCSE, SE, P(True), CoT-UQ-ME, Eig-E, CoT-UQ-SP, and Topo-UQ, to cover all investigated categories. The runtime of the considered UQ methods primarily consists of three parts arranged in decreasing order of computational cost, namely LLM inference, auxiliary model execution, and other specific computations such as calculating token-level statistics and deriving the eigenvalues of the similarity matrix. Modern LLMs like Qwen3 typically operate at the billion-parameter scale, whereas auxiliary models usually have millions of parameters. Thus, LLM inference is significantly longer than that of the auxiliary model, which makes the overall runtime of a UQ method primarily dominated by the number of LLM inferences.

\begin{table}[ht]
    \centering
    \footnotesize
    \setlength{\tabcolsep}{2.6pt}
    \resizebox{\linewidth}{!}{
        \begin{tabular}{@{} c c c c c c c @{}}
        \toprule
        \multirow{2}{*}{\textbf{Category}} & 
        \multirow{2}{*}{\textbf{UQ}} & 
        \multirow{2}{*}{\textbf{\makecell{Number of LLM \\ Inferences}}} & 
        \multicolumn{3}{c}{\textbf{Auxiliary Model Execution}} & 
        \multirow{2}{*}{\textbf{Other Computation}} \\
        \cmidrule(lr){4-6}
        & & & \textbf{Model} & \textbf{Scale} & \textbf{Number of Execution} & \\
        \midrule
        \multirow{2}{*}{\makecell{Information-\\based}} 
        & Ppl & 1 & - & - & - & Token-level statistics computation \\
        \cmidrule{2-7}
        & MCSE & $K$ & - & - & - & Token-level statistics computation \\
        \midrule
        \makecell{Reflexive-\\based} 
        & P(True) & 2 & - & - & - & - \\
        \midrule
        \multirow{2}{*}{\makecell{Reasoning-\\enhanced}} 
        & CoT-UQ-ME & 2 & - & - & - & Token-level statistics computation \\
        \cmidrule{2-7}
        & CoT-UQ-SP & 3 & - & - & - & - \\
        \midrule
        \multirow{3}{*}{\makecell{Diversity-\\based}} 
        & SE & $K$ & DeBERTa & 0.35B & \makecell{$K$ (Best case)\\ $K(K-1)/2$ (Worst case)} & Token-level statistics computation \\
        \cmidrule{2-7}
        & Eig-E & $K$ & DeBERTa & 0.35B & $K(K-1)/2$ & Eigen value computation \\
        \midrule
        \multirow{3}{*}{\makecell{Graph-\\based}} 
        & Topo-UQ & $L(n+2)$ & BERT-Base & 0.11B & $L(|V|+|E|)$ & Graph edit distance computation \\
        \cmidrule{2-7}
        & GUT-Q & $K$ & DeBERTa & 0.35B & \makecell{$KS$ (Best case)\\ $KS(K-1)(S-1)/4$ (Worst case)} & \makecell[l]{Token-level statistics computation\\ Graph complexity estimation} \\
        \bottomrule
    \end{tabular}
    }
    \caption{Runtime complexity comparison of GUT-Q and its contenders.}
    \label{tab:runtime_app}
\end{table}

Table~\ref{tab:runtime_app} shows the runtime complexity comparison of GUT-Q and its contenders, where $K \in \mathbb{N}^+$ is the number of CoT samples, $S \in \mathbb{N}^+$ denotes the average number of reasoning steps in a CoT chain, $L \in \mathbb{N}^+$ marks the per-specified number of constructed graphs, $n \in \mathbb{N}^+$ indicates the number of subproblems decomposed by prompting the LLM, $\vert{}V\vert{} \in \mathbb{N}^+$ is the number of vertices in the graph, and $\vert{}E\vert{} \in \mathbb{N}^+$ is the number of edges in the graph. Because SE and our proposed GUT-Q rely on chain-level or step-level merging guided by an auxiliary model, their best-case scenario occurs when all chains or steps belong to the same cluster, while the worst-case scenario arises when each forms a distinct cluster. There are two key observations. First, compared to representative diversity-based UQ methods such as SE and Eig-E, our proposed GUT-Q shares the same number of LLM prompts while requiring a slightly larger number of auxiliary model executions. Second, compared with classical graph-based methods like Topo-UQ, our proposed GUT-Q exhibits a comparable number of LLM prompts and auxiliary model executions. The above two observations suggest that our proposed GUT-Q runtime complexity is comparable to the investigated diversity-based and graph-based UQ methods, which typically achieve relatively better UQ performance in selective generation than the concerned information-based methods and reflexive-based methods, as shown in Section~\ref{sec:experiments}. 

\begin{figure}[h]
    \centering
    \includegraphics[width=\linewidth]{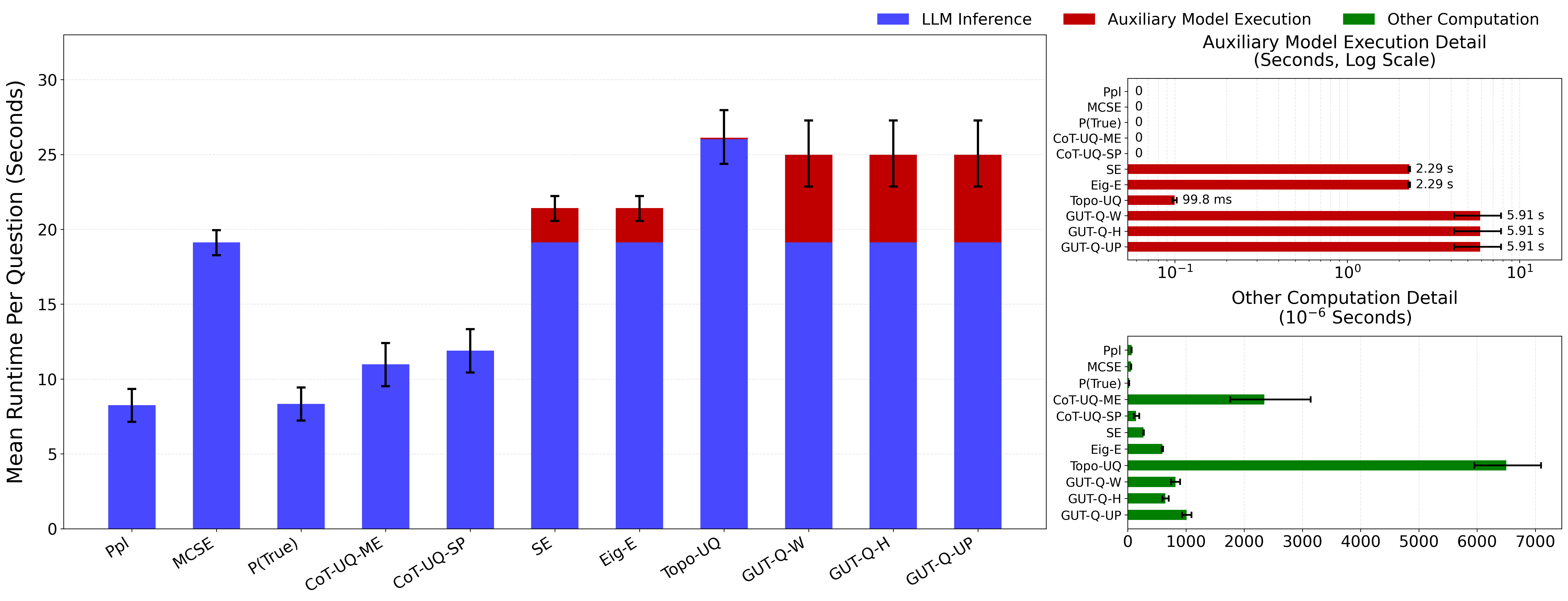}
    \caption{Average wall-clock time comparison of GUT-Q and its contenders on the MATH-500 dataset using Qwen3-4B.}
    \label{fig:runtime_qwen3-4B_math}
\end{figure}

Figure~\ref{fig:runtime_qwen3-4B_math} further visualizes the average wall-clock time comparison of GUT-Q and its contenders on the MATH-500 dataset using Qwen3-4B, where we set $K=10$ and $L=5$ to make LLM inference cost comparable. It is observed that GUT-Q exhibits a wall-clock time comparable to existing state-of-the-art methods such as SE, Eig-E, and Topo-UQ, which corroborates the aforementioned runtime analysis and corresponding conclusions. We can also observe that the sampling time of GUT-Q is only around 2.5 times that of the Ppl method, far less than $K=10$ times. This is because one can perform batch-parallel LLM prompting in implementation using packages like Transformers~\footnote{https://huggingface.co/docs/transformers/en/index} or vllm~\footnote{https://vllm.ai/}, which makes our proposed GUT-Q more efficient in practice.

\subsection{UQ Performance Evaluations}  \label{app:subsec:UQ_performance_evaluations}
Tables~\ref{tab:UQ_8B_app},~\ref{tab:UQ_4B_app},~\ref{tab:UQ_1.7B_app}, and~\ref{tab:UQ_0.6B_app} show the UQ performance comparisons across 5 datasets for Qwen3-8B, Qwen3-4B, Qwen3-1.7B, and Qwen3-0.6B, respectively. It is observed that the GUT-Q implemented with the UP algorithm generally outperforms all contenders across three evaluation metrics and datasets spanning three difficulty levels, thus validating the proposed GUT-Q.

\begin{sidewaystable}[ph] 
  \centering
  \footnotesize
  \setlength{\tabcolsep}{2.6pt}
  \resizebox{0.9\linewidth}{!}{
\begin{tabular}{lrrrrrrrrrrrrrrr}
      \toprule
      \multicolumn{1}{c}{\multirow{2}{*}[-0.8ex]{\textbf{UQ}}} & \multicolumn{3}{c}{\textbf{GSM8K}} & \multicolumn{3}{c}{\textbf{MATH500}} & \multicolumn{3}{c}{\textbf{AMC2022-2024}} & \multicolumn{3}{c}{\textbf{FOLIO}} & \multicolumn{3}{c}{\textbf{MMLU-Pro}} \\
      \cmidrule(lr){2-4} \cmidrule(lr){5-7} \cmidrule(lr){8-10} \cmidrule(lr){11-13} \cmidrule(lr){14-16}
                 & \multicolumn{1}{c}{\textbf{PRR}} & \multicolumn{1}{c}{\textbf{AUROC}} & \multicolumn{1}{c}{\textbf{AUPRC}} & \multicolumn{1}{c}{\textbf{PRR}} & \multicolumn{1}{c}{\textbf{AUROC}} & \multicolumn{1}{c}{\textbf{AUPRC}} & \multicolumn{1}{c}{\textbf{PRR}} & \multicolumn{1}{c}{\textbf{AUROC}} & \multicolumn{1}{c}{\textbf{AUPRC}} & \multicolumn{1}{c}{\textbf{PRR}} & \multicolumn{1}{c}{\textbf{AUROC}} & \multicolumn{1}{c}{\textbf{AUPRC}} & \multicolumn{1}{c}{\textbf{PRR}} & \multicolumn{1}{c}{\textbf{AUROC}} & \multicolumn{1}{c}{\textbf{AUPRC}} \\
      \midrule
      MSP        & -11.17$\pm$1.04                  & 56.18$\pm$1.43                     & 90.85$\pm$0.65                     & -28.55$\pm$2.57                  & 59.25$\pm$1.49                     & 70.09$\pm$1.76                     & -62.22$\pm$4.03                  & \underline{74.67$\pm$2.88}         & 67.62$\pm$4.05                     & 1.53$\pm$3.27                    & 51.63$\pm$3.05                     & 74.18$\pm$2.78                     & -18.28$\pm$3.51                  & 59.41$\pm$2.66                     & 69.52$\pm$2.79                     \\
\bottomrule
    \end{tabular}
    }
    \caption{UQ performance comparisons for Qwen3-8B, where bold and underline denote the group best and second-best results.}
  \label{tab:UQ_8B_app}
\end{sidewaystable}


\begin{sidewaystable*}[htbp]
  \centering
  \footnotesize
  \setlength{\tabcolsep}{2.6pt} 
  \resizebox{0.9\linewidth}{!}{
    \begin{tabular}{lrrrrrrrrrrrrrrr}
      \toprule
      \multicolumn{1}{c}{\multirow{2}{*}[-0.8ex]{\textbf{UQ}}} & \multicolumn{3}{c}{\textbf{GSM8K}} & \multicolumn{3}{c}{\textbf{MATH-500}} & \multicolumn{3}{c}{\textbf{AMC2022-2024}} & \multicolumn{3}{c}{\textbf{FOLIO}} & \multicolumn{3}{c}{\textbf{MMLU-Pro}} \\
      \cmidrule(lr){2-4} \cmidrule(lr){5-7} \cmidrule(lr){8-10} \cmidrule(lr){11-13} \cmidrule(lr){14-16}
                 & \multicolumn{1}{c}{\textbf{PRR}} & \multicolumn{1}{c}{\textbf{AUROC}} & \multicolumn{1}{c}{\textbf{AUPRC}} & \multicolumn{1}{c}{\textbf{PRR}} & \multicolumn{1}{c}{\textbf{AUROC}} & \multicolumn{1}{c}{\textbf{AUPRC}} & \multicolumn{1}{c}{\textbf{PRR}} & \multicolumn{1}{c}{\textbf{AUROC}} & \multicolumn{1}{c}{\textbf{AUPRC}} & \multicolumn{1}{c}{\textbf{PRR}} & \multicolumn{1}{c}{\textbf{AUROC}} & \multicolumn{1}{c}{\textbf{AUPRC}} & \multicolumn{1}{c}{\textbf{PRR}} & \multicolumn{1}{c}{\textbf{AUROC}} & \multicolumn{1}{c}{\textbf{AUPRC}} \\
      \midrule
      MSP        & -33.66$\pm$1.22                  & 58.89$\pm$0.97                     & 86.84$\pm$0.80                     & -8.09$\pm$2.46                   & 58.76$\pm$1.67                     & 60.85$\pm$2.04                     & -37.65$\pm$4.97                  & 67.35$\pm$3.70                     & 52.35$\pm$4.27                     & -11.70$\pm$3.74                  & 55.42$\pm$2.22                     & 74.25$\pm$2.73                     & -49.44$\pm$3.39                  & \underline{66.52$\pm$2.12}         & 71.30$\pm$2.55                     \\
\bottomrule
    \end{tabular}
  }
  \caption{UQ performance comparisons for Qwen3-4B, where bold and underline denote the group best and second-best results, respectively.}
  \label{tab:UQ_4B_app}
\end{sidewaystable*}

\clearpage

\begin{sidewaystable*}[htbp] 
    \centering
    \footnotesize
    \setlength{\tabcolsep}{2.6pt}
    \resizebox{0.9\linewidth}{!}{ 
        \begin{tabular}{lrrrrrrrrrrrrrrr}
          \toprule
          \multicolumn{1}{c}{\multirow{2}{*}[-0.8ex]{\textbf{UQ}}} & \multicolumn{3}{c}{\textbf{GSM8K}} & \multicolumn{3}{c}{\textbf{MATH-500}} & \multicolumn{3}{c}{\textbf{AMC2022-2024}} & \multicolumn{3}{c}{\textbf{FOLIO}} & \multicolumn{3}{c}{\textbf{MMLU-Pro}} \\
          \cmidrule(lr){2-4} \cmidrule(lr){5-7} \cmidrule(lr){8-10} \cmidrule(lr){11-13} \cmidrule(lr){14-16}
                     & \multicolumn{1}{c}{\textbf{PRR}} & \multicolumn{1}{c}{\textbf{AUROC}} & \multicolumn{1}{c}{\textbf{AUPRC}} & \multicolumn{1}{c}{\textbf{PRR}} & \multicolumn{1}{c}{\textbf{AUROC}} & \multicolumn{1}{c}{\textbf{AUPRC}} & \multicolumn{1}{c}{\textbf{PRR}} & \multicolumn{1}{c}{\textbf{AUROC}} & \multicolumn{1}{c}{\textbf{AUPRC}} & \multicolumn{1}{c}{\textbf{PRR}} & \multicolumn{1}{c}{\textbf{AUROC}} & \multicolumn{1}{c}{\textbf{AUPRC}} & \multicolumn{1}{c}{\textbf{PRR}} & \multicolumn{1}{c}{\textbf{AUROC}} & \multicolumn{1}{c}{\textbf{AUPRC}} \\
          \midrule
          MSP        & -8.75$\pm$1.44                   & 52.25$\pm$0.52                     & 73.41$\pm$1.01                     & -30.34$\pm$2.30                  & 59.01$\pm$1.03                     & 46.33$\pm$1.97                     & -31.58$\pm$3.15                  & 62.78$\pm$3.01                     & 29.86$\pm$3.80                     & -58.17$\pm$2.48                  & 73.98$\pm$1.98                     & 61.46$\pm$3.11                     & -41.57$\pm$2.80                  & 62.75$\pm$1.70                     & 51.66$\pm$2.77                     \\
\bottomrule
        \end{tabular}
    }
    \caption{UQ performance comparisons for Qwen3-1.7B, where bold and underline denote the group best and second-best results.}
    \label{tab:UQ_1.7B_app}
\end{sidewaystable*}

\clearpage

\begin{sidewaystable*}[htbp]
	\centering
	\footnotesize
    \setlength{\tabcolsep}{2.6pt}
	\resizebox{0.9\linewidth}{!}{
        \begin{tabular}{lrrrrrrrrrrrrrrr}
          \toprule
          \multicolumn{1}{c}{\multirow{2}{*}[-0.8ex]{\textbf{UQ}}} & \multicolumn{3}{c}{\textbf{GSM8K}} & \multicolumn{3}{c}{\textbf{MATH-500}} & \multicolumn{3}{c}{\textbf{AMC2022-2024}} & \multicolumn{3}{c}{\textbf{FOLIO}} & \multicolumn{3}{c}{\textbf{MMLU-Pro}} \\
          \cmidrule(lr){2-4} \cmidrule(lr){5-7} \cmidrule(lr){8-10} \cmidrule(lr){11-13} \cmidrule(lr){14-16}
                     & \multicolumn{1}{c}{\textbf{PRR}} & \multicolumn{1}{c}{\textbf{AUROC}} & \multicolumn{1}{c}{\textbf{AUPRC}} & \multicolumn{1}{c}{\textbf{PRR}} & \multicolumn{1}{c}{\textbf{AUROC}} & \multicolumn{1}{c}{\textbf{AUPRC}} & \multicolumn{1}{c}{\textbf{PRR}} & \multicolumn{1}{c}{\textbf{AUROC}} & \multicolumn{1}{c}{\textbf{AUPRC}} & \multicolumn{1}{c}{\textbf{PRR}} & \multicolumn{1}{c}{\textbf{AUROC}} & \multicolumn{1}{c}{\textbf{AUPRC}} & \multicolumn{1}{c}{\textbf{PRR}} & \multicolumn{1}{c}{\textbf{AUROC}} & \multicolumn{1}{c}{\textbf{AUPRC}} \\
          \midrule
          MSP        & 0.05$\pm$1.42                    & 49.90$\pm$0.68                     & 29.94$\pm$1.07                     & -15.13$\pm$1.43                  & 51.67$\pm$1.65                     & 15.07$\pm$1.61                     & -20.88$\pm$1.81                  & 56.64$\pm$1.30                     & 7.55$\pm$2.31                      & 5.71$\pm$3.92                    & 50.57$\pm$0.91                     & 43.15$\pm$2.86                     & 11.28$\pm$3.36                   & 46.69$\pm$1.68                     & 26.58$\pm$2.30                     \\
\bottomrule
        \end{tabular}
	}
    \caption{UQ performance comparisons for Qwen3-0.6B, where bold and underline denote the group best and second-best results.}
	\label{tab:UQ_0.6B_app}
\end{sidewaystable*}

\clearpage

\section{Additional Experimental Details on GUT-O}  \label{app:UO_additional_exp}
This appendix provides additional implementation details and experimental results on the proposed GUT-O module.

\subsection{Implementation Details}  \label{app:subsec:UO_hyperparameter}
The GUT-O was implemented with the open-r1\footnote{https://github.com/huggingface/open-r1} repository. Table~\ref{tab:UO_hyperparameter_app} lists the configuration of hyperparameters employed in the GUT-O. We fine-tuned the Qwen3-8B, Qwen3-4B, Qwen3-1.7B, and Qwen3-0.6B on the training set of the GSM8K~\citep{cobbe2021training}, MATH~\citep{hendrycks2021measuring}, FOLIO~\citep{han2024folio}, MMLU-Pro~\citep{wang2024mmlu}, and AMC2022-2024, respectively, where we randomly select 20\% instances of the synthetic AMC2022-2024 as training instances. See Appendix~\ref{app:subsec:dataset} for details of AMC2022-2024.
 
\begin{table}[htbp]
  \centering
  \footnotesize
  \begin{tabular}{cc|cc}
    \toprule
    \textbf{Hyperparameter} & \textbf{Value} & \textbf{Hyperparameter} & \textbf{Value}     \\
    \midrule
    Batch Size              & 128            & Max New Tokens          & 3072               \\
    Group Size $M$          & 7              & Learning Rate           & $3 \times 10^{-6}$ \\
    \text{KL} Penalty $\beta$      & 0.0005         & Learning Rate Scheduler & Cosine             \\
    Temperature $T$         & 1.0            & Warmup Ratio            & 0.1                \\
    Clip Ratio $\epsilon$   & 0.2            & Optimizer               & AdamW              \\
    Training Steps          & 60                 \\
    \bottomrule
  \end{tabular}
  \caption{Configuration of hyperparameters for GUT-O.}
  \label{tab:UO_hyperparameter_app}
\end{table}

\subsection{UO Performance Evaluation}  \label{app:subsec:uncertainty_reduction}
Table~\ref{tab:UO_performance_app} shows the percentage changes in mean uncertainty and absolute changes in accuracy for Qwen3-1.7B and Qwen3-0.6B. It is observed that the proposed GUT-O effectively reduces the uncertainty and improves accuracy across LLM scales, datasets, and UQ evaluation metrics. This observation validates the proposed GUT-O module.

\begingroup
\footnotesize
\setlength{\tabcolsep}{2pt}
\begin{longtable}{cccclll}
\caption{Comparisons of UO performance for Qwen3 families across 5 reasoning datasets, where `+' indicates the applied fine-tuning method, and values in parentheses indicate the percentage change in mean uncertainty.}
\label{tab:UO_performance_app} \\
\toprule
\multirow{2}{*}{\textbf{Dataset}} & \multirow{2}{*}{\textbf{Models}} & \multirow{2}{*}{\textbf{Accuracy}} & \multirow{2}{*}{\textbf{MTLP}} & \multicolumn{3}{c}{\textbf{Mean Uncertainty}} \\
\cmidrule(lr){5-7}
                                  &                                  &                                    &                                   & \multicolumn{1}{c}{\textbf{GUT-Q-W}} & \multicolumn{1}{c}{\textbf{GUT-Q-H}} & \multicolumn{1}{c}{\textbf{GUT-Q-UP}} \\
\midrule
\endfirsthead

\caption[]{Comparisons of UO performance for Qwen3 families (Continued).} \\
\toprule
\multirow{2}{*}{\textbf{Dataset}} & \multirow{2}{*}{\textbf{Models}} & \multirow{2}{*}{\textbf{Accuracy}} & \multirow{2}{*}{\textbf{MTLP}} & \multicolumn{3}{c}{\textbf{Mean Uncertainty}} \\
\cmidrule(lr){5-7}
                                  &                                  &                                    &                                   & \multicolumn{1}{c}{\textbf{GUT-Q-W}} & \multicolumn{1}{c}{\textbf{GUT-Q-H}} & \multicolumn{1}{c}{\textbf{GUT-Q-UP}} \\
\midrule
\endhead

\midrule
\multicolumn{7}{r}{\textit{Continued on next page}} \\
\endfoot

\bottomrule
\endlastfoot

\multirow{12}{*}{GSM8K}        & Qwen3-0.6B       & $28.89 \pm 0.97$ & $-7.11 \pm 0.02$  & $3.10 \pm 0.02$          & $5.93 \pm 0.04$          & $4.14 \pm 0.04$ \\
                               & \quad+GRPO  & $38.97 \pm 1.05$ & $-7.38 \pm 0.02$  & $3.03 \pm 0.02$ (-2.3\%) & $5.77 \pm 0.04$ (-2.7\%) & $3.99 \pm 0.04$ (-3.6\%) \\
                               & \quad+GUT-O & $38.06 \pm 1.07$ & $-8.54 \pm 0.02$  & $2.78 \pm 0.02$ (-10.3\%)& $5.27 \pm 0.04$ (-11.1\%)& $3.32 \pm 0.03$ (-19.8\%)\\
\cmidrule{2-7}
                               & Qwen3-1.7B       & $71.11 \pm 1.00$ & $-14.54 \pm 0.03$ & $1.68 \pm 0.02$          & $5.51 \pm 0.06$          & $1.58 \pm 0.03$ \\
                               & \quad+GRPO  & $72.93 \pm 0.95$ & $-14.28 \pm 0.04$ & $1.69 \pm 0.02$ (+0.6\%) & $5.41 \pm 0.06$ (-1.8\%) & $1.62 \pm 0.03$ (+2.5\%) \\
                               & \quad+GUT-O & $72.78 \pm 0.96$ & $-15.37 \pm 0.04$ & $1.58 \pm 0.02$ (-6.0\%) & $5.22 \pm 0.06$ (-5.3\%) & $1.51 \pm 0.02$ (-4.4\%) \\
\cmidrule{2-7}
                               & Qwen3-4B         & $84.08 \pm 0.79$ & $-13.48 \pm 0.04$ & $1.45 \pm 0.02$          & $4.47 \pm 0.04$          & $1.18 \pm 0.02$ \\
                               & \quad+GRPO   & $86.81 \pm 0.75$ & $-13.22 \pm 0.04$ & $1.41 \pm 0.02$ (-2.8\%) & $4.42 \pm 0.04$ (-1.1\%) & $1.19 \pm 0.02$ (+0.8\%) \\
                               & \quad+GUT-O  & $84.91 \pm 0.80$ & $-13.49 \pm 0.03$ & $1.39 \pm 0.02$ (-4.1\%) & $4.37 \pm 0.04$ (-2.2\%) & $1.15 \pm 0.02$ (-2.5\%) \\
\cmidrule{2-7}
                               & Qwen3-8B         & $89.61 \pm 0.67$ & $-16.56 \pm 0.05$ & $1.28 \pm 0.01$          & $4.15 \pm 0.03$          & $0.95 \pm 0.01$ \\
                               & \quad+GRPO   & $90.75 \pm 0.64$ & $-15.96 \pm 0.05$ & $1.28 \pm 0.01$ (+0.0\%) & $4.14 \pm 0.03$ (-0.2\%) & $0.94 \pm 0.01$ (-1.1\%) \\
                               & \quad+GUT-O  & $89.76 \pm 0.64$ & $-17.20 \pm 0.05$ & $1.25 \pm 0.01$ (-2.3\%) & $4.14 \pm 0.04$ (-0.2\%) & $0.92 \pm 0.01$ (-3.2\%) \\
\midrule

\multirow{12}{*}{MATH-500}     & Qwen3-0.6B       & $18.00 \pm 1.31$ & $-7.55 \pm 0.04$  & $0.77 \pm 0.01$          & $1.86 \pm 0.03$          & $1.14 \pm 0.01$ \\
                               & \quad+GRPO  & $23.40 \pm 1.51$ & $-7.67 \pm 0.04$  & $0.74 \pm 0.01$ (-3.9\%) & $1.88 \pm 0.03$ (+1.1\%) & $1.17 \pm 0.02$ (+2.6\%) \\
                               & \quad+GUT-O & $21.40 \pm 1.43$ & $-8.94 \pm 0.04$  & $0.64 \pm 0.01$ (-16.9\%)& $1.51 \pm 0.03$ (-18.8\%)& $0.88 \pm 0.01$ (-22.8\%)\\
\cmidrule{2-7}
                               & Qwen3-1.7B       & $42.20 \pm 1.79$ & $-13.45 \pm 0.06$ & $0.43 \pm 0.01$          & $2.03 \pm 0.05$          & $0.47 \pm 0.01$ \\
                               & \quad+GRPO  & $46.40 \pm 1.81$ & $-13.56 \pm 0.07$ & $0.42 \pm 0.01$ (-2.3\%) & $1.89 \pm 0.05$ (-6.9\%) & $0.46 \pm 0.01$ (-2.1\%) \\
                               & \quad+GUT-O & $43.00 \pm 1.67$ & $-14.03 \pm 0.07$ & $0.39 \pm 0.01$ (-9.3\%) & $1.82 \pm 0.05$ (-10.3\%)& $0.43 \pm 0.01$ (-8.5\%) \\
\cmidrule{2-7}
                               & Qwen3-4B         & $53.80 \pm 1.71$ & $-14.14 \pm 0.08$ & $0.37 \pm 0.01$          & $1.76 \pm 0.05$          & $0.40 \pm 0.01$ \\
                               & \quad+GRPO   & $61.00 \pm 1.68$ & $-14.11 \pm 0.09$ & $0.38 \pm 0.01$ (+2.7\%) & $1.78 \pm 0.04$ (+1.1\%) & $0.40 \pm 0.01$ (+0.0\%) \\
                               & \quad+GUT-O  & $57.00 \pm 1.71$ & $-14.25 \pm 0.07$ & $0.34 \pm 0.00$ (-8.1\%) & $1.70 \pm 0.04$ (-3.4\%) & $0.32 \pm 0.01$ (-20.0\%)\\
\cmidrule{2-7}
                               & Qwen3-8B         & $64.20 \pm 1.68$ & $-15.14 \pm 0.09$ & $0.33 \pm 0.01$          & $1.46 \pm 0.04$          & $0.24 \pm 0.01$ \\
                               & \quad+GRPO   & $66.60 \pm 1.64$ & $-14.97 \pm 0.09$ & $0.33 \pm 0.01$ (+0.0\%) & $1.43 \pm 0.04$ (-2.1\%) & $0.25 \pm 0.01$ (+4.2\%) \\
                               & \quad+GUT-O  & $64.80 \pm 1.72$ & $-15.78 \pm 0.08$ & $0.31 \pm 0.01$ (-6.1\%) & $1.38 \pm 0.04$ (-5.5\%) & $0.23 \pm 0.01$ (-4.2\%) \\
\midrule

\multirow{12}{*}{AMC2022-2024} & Qwen3-0.6B       & $9.09 \pm 2.09$  & $-7.12 \pm 0.07$  & $5.10 \pm 0.10$          & $14.14 \pm 0.41$         & $5.72 \pm 0.14$ \\
                               & \quad+GRPO  & $11.57 \pm 2.32$ & $-7.22 \pm 0.08$  & $4.99 \pm 0.10$ (-2.2\%) & $15.07 \pm 0.38$ (+6.6\%)& $6.37 \pm 0.18$ (+11.4\%)\\
                               & \quad+GUT-O & $9.09 \pm 2.11$  & $-8.51 \pm 0.09$  & $3.97 \pm 0.09$ (-22.2\%)& $10.34 \pm 0.30$ (-26.9\%)& $4.16 \pm 0.12$ (-27.3\%)\\
\cmidrule{2-7}
                               & Qwen3-1.7B       & $22.31 \pm 3.03$ & $-12.96 \pm 0.15$ & $2.92 \pm 0.08$          & $20.35 \pm 0.94$         & $3.54 \pm 0.14$ \\
                               & \quad+GRPO  & $30.58 \pm 3.31$ & $-12.98 \pm 0.14$ & $3.10 \pm 0.08$ (+6.2\%) & $17.69 \pm 0.88$ (-13.1\%)& $3.60 \pm 0.16$ (+1.7\%) \\
                               & \quad+GUT-O & $24.79 \pm 3.25$ & $-14.08 \pm 0.15$ & $2.46 \pm 0.07$ (-15.8\%)& $16.16 \pm 0.74$ (-20.6\%)& $2.80 \pm 0.13$ (-20.9\%)\\
\cmidrule{2-7}
                               & Qwen3-4B         & $39.67 \pm 3.52$ & $-13.53 \pm 0.15$ & $2.69 \pm 0.08$          & $16.88 \pm 0.86$         & $3.27 \pm 0.17$ \\
                               & \quad+GRPO   & $42.98 \pm 3.73$ & $-13.92 \pm 0.16$ & $2.74 \pm 0.08$ (+1.9\%) & $15.19 \pm 0.67$ (-10.0\%)& $3.07 \pm 0.15$ (-6.1\%) \\
                               & \quad+GUT-O  & $44.63 \pm 3.62$ & $-14.45 \pm 0.14$ & $2.16 \pm 0.07$ (-19.7\%)& $14.66 \pm 0.75$ (-13.2\%)& $1.67 \pm 0.09$ (-48.9\%)\\
\cmidrule{2-7}
                               & Qwen3-8B         & $48.76 \pm 3.84$ & $-15.32 \pm 0.19$ & $2.30 \pm 0.07$          & $16.99 \pm 1.00$         & $1.58 \pm 0.10$ \\
                               & \quad+GRPO   & $51.24 \pm 3.71$ & $-14.90 \pm 0.17$ & $2.37 \pm 0.08$ (+3.0\%) & $15.94 \pm 0.77$ (-6.2\%)& $1.75 \pm 0.10$ (+10.8\%)\\
                               & \quad+GUT-O  & $49.59 \pm 3.64$ & $-16.11 \pm 0.20$ & $2.24 \pm 0.07$ (-2.6\%) & $15.87 \pm 1.00$ (-6.6\%)& $1.43 \pm 0.09$ (-9.5\%) \\
\midrule

\multirow{12}{*}{FOLIO}        & Qwen3-0.6B       & $44.83 \pm 2.69$ & $-7.04 \pm 0.04$  & $0.47 \pm 0.01$          & $1.49 \pm 0.03$          & $0.46 \pm 0.01$ \\
                               & \quad+GRPO  & $51.23 \pm 2.73$ & $-7.23 \pm 0.04$  & $0.46 \pm 0.01$ (-2.1\%) & $1.34 \pm 0.03$ (-10.1\%)& $0.44 \pm 0.01$ (-4.3\%) \\
                               & \quad+GUT-O & $46.31 \pm 2.78$ & $-7.19 \pm 0.04$  & $0.46 \pm 0.01$ (-2.1\%) & $1.38 \pm 0.02$ (-7.4\%) & $0.45 \pm 0.01$ (-2.2\%) \\
\cmidrule{2-7}
                               & Qwen3-1.7B       & $43.84 \pm 2.98$ & $-13.46 \pm 0.09$ & $0.24 \pm 0.01$          & $1.31 \pm 0.04$          & $0.25 \pm 0.01$ \\
                               & \quad+GRPO  & $48.77 \pm 2.97$ & $-13.15 \pm 0.08$ & $0.24 \pm 0.01$ (+0.0\%) & $1.33 \pm 0.04$ (+1.5\%) & $0.25 \pm 0.01$ (+0.0\%) \\
                               & \quad+GUT-O & $45.81 \pm 3.04$ & $-14.43 \pm 0.08$ & $0.21 \pm 0.00$ (-12.5\%)& $1.08 \pm 0.03$ (-17.6\%)& $0.23 \pm 0.01$ (-8.0\%) \\
\cmidrule{2-7}
                               & Qwen3-4B         & $68.47 \pm 2.56$ & $-10.49 \pm 0.07$ & $0.31 \pm 0.01$          & $1.10 \pm 0.03$          & $0.31 \pm 0.01$ \\
                               & \quad+GRPO   & $72.91 \pm 2.54$ & $-10.52 \pm 0.06$ & $0.29 \pm 0.01$ (-6.5\%) & $1.08 \pm 0.03$ (-1.8\%) & $0.30 \pm 0.01$ (-3.2\%) \\
                               & \quad+GUT-O  & $68.97 \pm 2.53$ & $-11.01 \pm 0.06$ & $0.27 \pm 0.01$ (-12.9\%)& $1.05 \pm 0.03$ (-4.5\%) & $0.28 \pm 0.01$ (-9.7\%) \\
\cmidrule{2-7}
                               & Qwen3-8B         & $72.41 \pm 2.45$ & $-12.75 \pm 0.15$ & $0.27 \pm 0.01$          & $1.03 \pm 0.03$          & $0.25 \pm 0.01$ \\
                               & \quad+GRPO   & $74.88 \pm 2.44$ & $-13.18 \pm 0.16$ & $0.27 \pm 0.01$ (+0.0\%) & $0.98 \pm 0.03$ (-4.9\%) & $0.25 \pm 0.01$ (+0.0\%) \\
                               & \quad+GUT-O  & $74.88 \pm 2.48$ & $-13.31 \pm 0.16$ & $0.25 \pm 0.01$ (-7.4\%) & $0.97 \pm 0.03$ (-5.8\%) & $0.24 \pm 0.01$ (-4.0\%) \\
\midrule

\multirow{12}{*}{MMLU-Pro}     & Qwen3-0.6B       & $29.17 \pm 2.39$ & $-6.19 \pm 0.04$  & $0.87 \pm 0.02$          & $1.68 \pm 0.03$          & $0.78 \pm 0.01$ \\
                               & \quad+GRPO  & $33.33 \pm 2.39$ & $-6.21 \pm 0.05$  & $0.85 \pm 0.02$ (-2.3\%) & $1.73 \pm 0.03$ (+3.0\%) & $0.78 \pm 0.01$ (+0.0\%) \\
                               & \quad+GUT-O & $29.58 \pm 2.30$ & $-7.31 \pm 0.05$  & $0.75 \pm 0.02$ (-13.8\%)& $1.31 \pm 0.03$ (-22.0\%)& $0.65 \pm 0.01$ (-16.7\%)\\
\cmidrule{2-7}
                               & Qwen3-1.7B       & $43.33 \pm 2.65$ & $-9.86 \pm 0.09$  & $0.58 \pm 0.01$          & $2.31 \pm 0.06$          & $0.54 \pm 0.01$ \\
                               & \quad+GRPO  & $50.42 \pm 2.55$ & $-9.90 \pm 0.09$  & $0.58 \pm 0.01$ (+0.0\%) & $2.32 \pm 0.06$ (+0.4\%) & $0.54 \pm 0.01$ (+0.0\%) \\
                               & \quad+GUT-O & $46.25 \pm 2.60$ & $-10.61 \pm 0.09$ & $0.53 \pm 0.01$ (-8.6\%) & $1.88 \pm 0.05$ (-18.6\%)& $0.48 \pm 0.01$ (-11.1\%)\\
\cmidrule{2-7}
                               & Qwen3-4B         & $60.83 \pm 2.60$ & $-11.05 \pm 0.09$ & $0.46 \pm 0.01$          & $1.50 \pm 0.04$          & $0.36 \pm 0.01$ \\
                               & \quad+GRPO   & $67.08 \pm 2.47$ & $-11.04 \pm 0.08$ & $0.45 \pm 0.01$ (-2.2\%) & $1.50 \pm 0.04$ (+0.0\%) & $0.36 \pm 0.01$ (+0.0\%) \\
                               & \quad+GUT-O  & $62.08 \pm 2.49$ & $-11.57 \pm 0.08$ & $0.45 \pm 0.01$ (-2.2\%) & $1.27 \pm 0.04$ (-15.3\%)& $0.31 \pm 0.01$ (-13.9\%)\\
\cmidrule{2-7}
                               & Qwen3-8B         & $67.92 \pm 2.49$ & $-12.79 \pm 0.14$ & $0.43 \pm 0.01$          & $1.33 \pm 0.04$          & $0.30 \pm 0.01$ \\
                               & \quad+GRPO   & $70.42 \pm 2.42$ & $-12.61 \pm 0.15$ & $0.44 \pm 0.01$ (+2.3\%) & $1.33 \pm 0.04$ (+0.0\%) & $0.30 \pm 0.01$ (+0.0\%) \\
                               & \quad+GUT-O  & $69.58 \pm 2.39$ & $-13.53 \pm 0.16$ & $0.41 \pm 0.01$ (-4.7\%) & $1.23 \pm 0.04$ (-7.5\%) & $0.29 \pm 0.01$ (-3.3\%) \\
\end{longtable}
\endgroup

\subsection{Validation of the MTLP optimization target}  \label{app:subsec:proxy_validation}
Table~\ref{tab:PCC_app} shows the PCCs between the GUT-Q-derived uncertainty $U(x)$ and the sum of the MTLP optimization target $\sum_{i=1}^N{U( x, c^i )}/N$ for the Qwen3 family, at scales of 0.6B, 1.7B, 4B, and 8B, across five investigated datasets. It is observed that the $U(x)$ generally has significant positive PCCs with the MTLP optimization target spanning datasets for all 4 LLMs. Therefore, we can consider that the MTLP serves as a valid optimization target for optimizing $U(x)$.

\begin{table*}[ht]
  \centering
  \footnotesize
  \setlength{\tabcolsep}{3pt}
  \begin{tabular}{llccccc}
    \toprule
    \textbf{Model} & \textbf{UQ} & \textbf{GSM8K} & \textbf{MATH-500} & \textbf{AMC2022-2024} & \textbf{FOLIO} & \textbf{MMLU-Pro} \\
    \midrule
    \multirow{3}{*}{Qwen3-0.6B} 
    & GUT-Q-W  & 0.2755 $\pm$ 0.0203 & 0.1611 $\pm$ 0.0314 & 0.3310 $\pm$ 0.0646 & 0.2397 $\pm$ 0.0548 & 0.2566 $\pm$ 0.0432 \\
    & GUT-Q-H & 0.2033 $\pm$ 0.0220 & 0.1718 $\pm$ 0.0345 & 0.0724 $\pm$ 0.0690 & 0.1917 $\pm$ 0.0484 & 0.2542 $\pm$ 0.0449 \\
    & GUT-Q-UP     & 0.2980 $\pm$ 0.0207 & 0.0734 $\pm$ 0.0337 & 0.1034 $\pm$ 0.0695 & 0.2450 $\pm$ 0.0495 & 0.2243 $\pm$ 0.0470 \\
    \midrule
    \multirow{3}{*}{Qwen3-1.7B} 
    & GUT-Q-W  & 0.2961 $\pm$ 0.0206 & 0.3874 $\pm$ 0.0308 & 0.3462 $\pm$ 0.0601 & 0.3145 $\pm$ 0.0539 & 0.3521 $\pm$ 0.0406 \\
    & GUT-Q-H & 0.2199 $\pm$ 0.0227 & 0.3213 $\pm$ 0.0291 & 0.3297 $\pm$ 0.0604 & 0.2773 $\pm$ 0.0594 & 0.3191 $\pm$ 0.0382 \\
    & GUT-Q-UP     & 0.3302 $\pm$ 0.0238 & 0.3260 $\pm$ 0.0366 & 0.1531 $\pm$ 0.0637 & 0.3952 $\pm$ 0.0512 & 0.2042 $\pm$ 0.0544 \\
    \midrule
    \multirow{3}{*}{Qwen3-4B} 
    & GUT-Q-W  & 0.0764 $\pm$ 0.0211 & 0.1857 $\pm$ 0.0307 & 0.3365 $\pm$ 0.0564 & 0.1531 $\pm$ 0.0637 & 0.2829 $\pm$ 0.0464 \\
    & GUT-Q-H & 0.0342 $\pm$ 0.0220 & 0.1581 $\pm$ 0.0309 & 0.3332 $\pm$ 0.0623 & 0.1228 $\pm$ 0.0570 & 0.1259 $\pm$ 0.0424 \\
    & GUT-Q-UP     & 0.0706 $\pm$ 0.0223 & 0.2121 $\pm$ 0.0304 & 0.2780 $\pm$ 0.0558 & 0.2339 $\pm$ 0.0523 & 0.2767 $\pm$ 0.0432 \\
    \midrule
    \multirow{3}{*}{Qwen3-8B} 
    & GUT-Q-W  & 0.1156 $\pm$ 0.0208 & 0.2114 $\pm$ 0.0331 & 0.4276 $\pm$ 0.0558 & 0.1259 $\pm$ 0.0557 & 0.2659 $\pm$ 0.0450 \\
    & GUT-Q-H & 0.0887 $\pm$ 0.0247 & 0.1266 $\pm$ 0.0327 & 0.5433 $\pm$ 0.0491 & 0.1128 $\pm$ 0.0493 & 0.2077 $\pm$ 0.0361 \\
    & GUT-Q-UP     & 0.1452 $\pm$ 0.0183 & 0.2159 $\pm$ 0.0327 & 0.4998 $\pm$ 0.0511 & 0.1684 $\pm$ 0.0473 & 0.3073 $\pm$ 0.0366 \\
    \bottomrule
  \end{tabular}
  \caption{PCCs between the MTLP optimization target and GUT-Q-derived uncertainties across 4 LLMs and 5 datasets.}
  \label{tab:PCC_app}
\end{table*}

\section{Additional Ablation and Sensitivity Analyses}  \label{app:ablation_sensitivity}
This appendix provides additional ablation and sensitivity analyses for Qwen3-8B, Qwen3-1.7B, and Qwen3-0.6B, as well as the first-order logic dataset FOLIO~\citep{han2024folio} and the reasoning QA dataset MMLU-Pro~\citep{wang2024mmlu}.

\subsection{Additional Ablation Analyses} \label{app:subsec:ablation}
\begin{figure}[htb]
    \centering
    \begin{subfigure}[b]{0.49\linewidth}
    	\centering
    	\includegraphics[width=\linewidth]{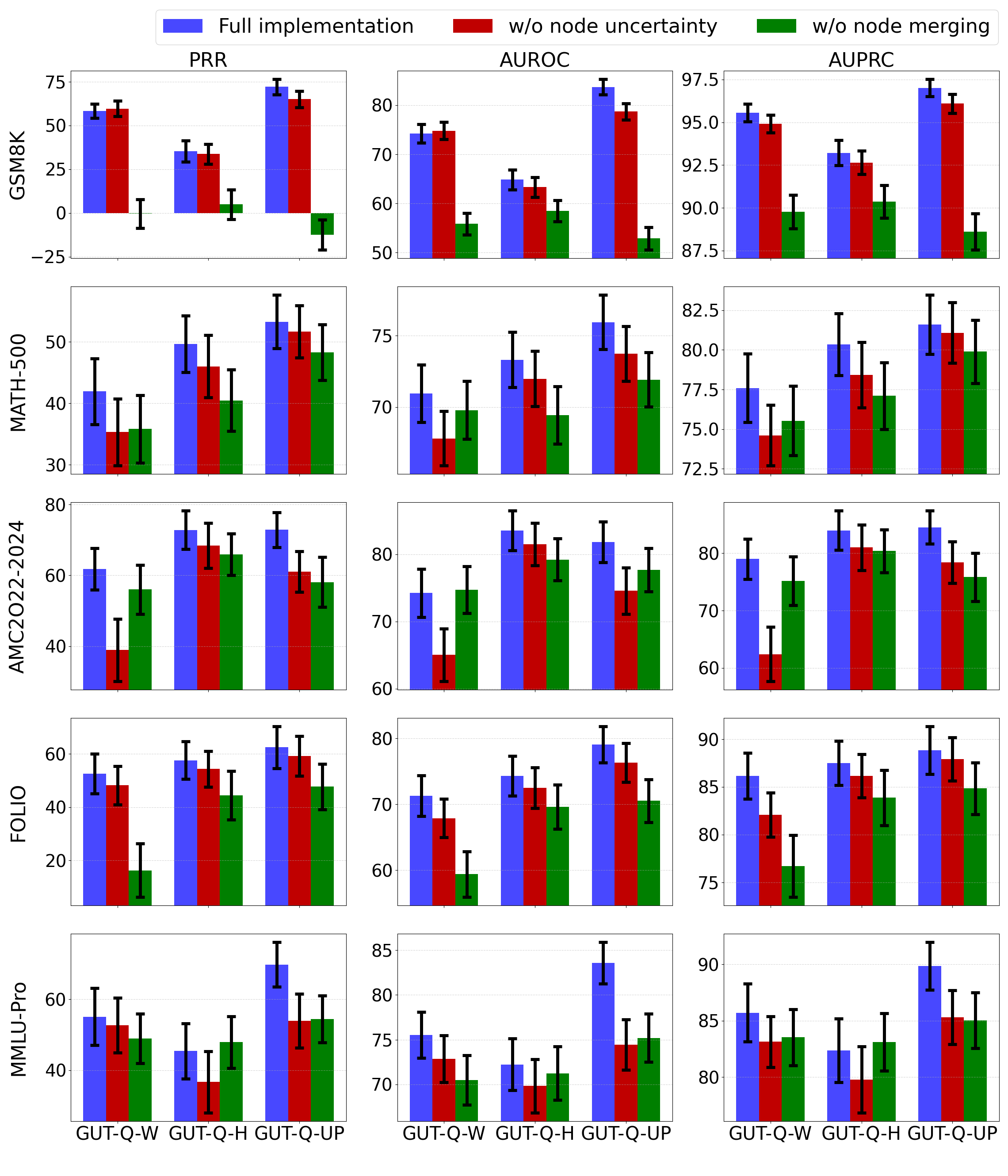}
    	\caption{Qwen3-8B}
        \label{fig:ablation_8B_app}
    \end{subfigure}
    \hfill
    \begin{subfigure}[b]{0.49\linewidth}
    	\centering
    	\includegraphics[width=\linewidth]{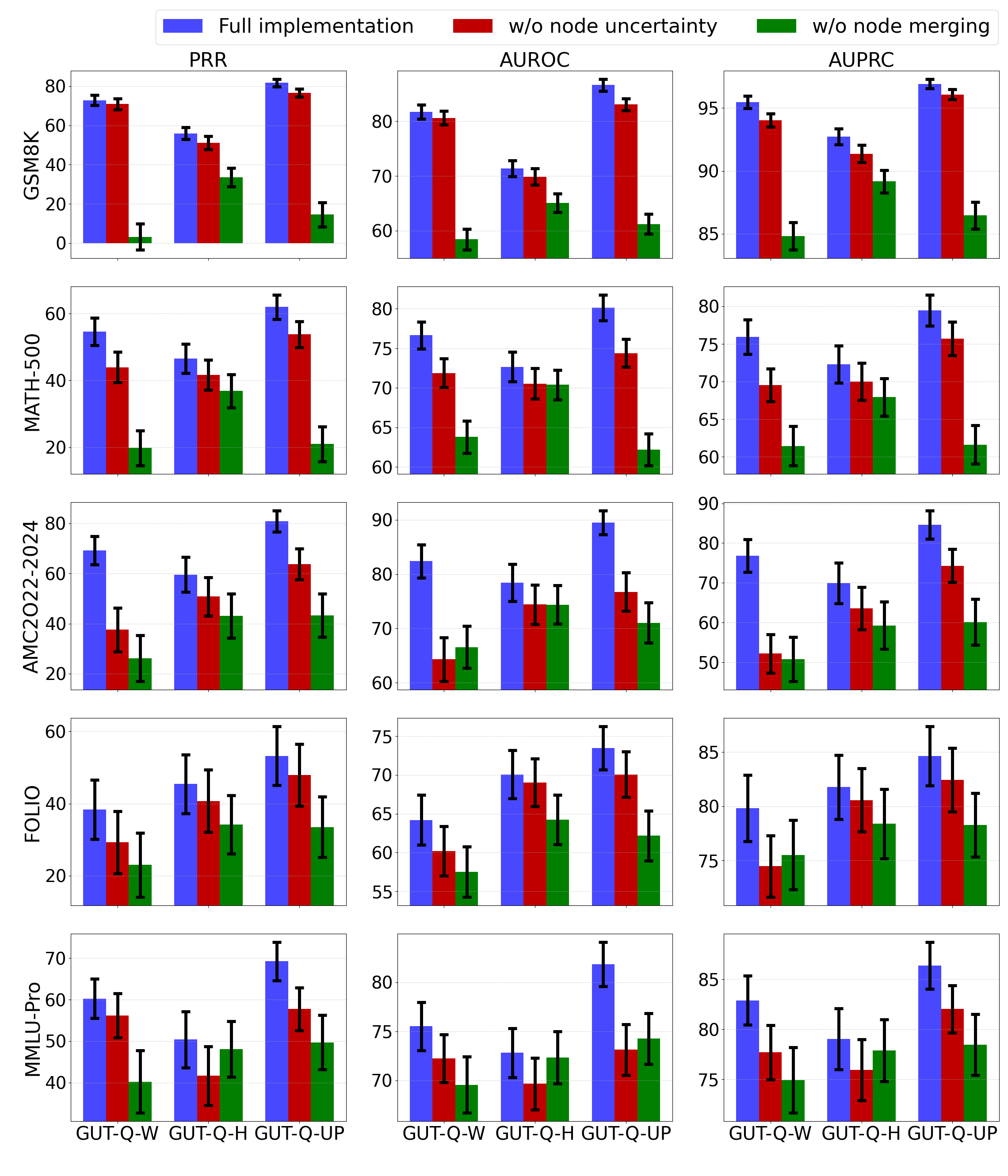}
    	\caption{Qwen3-4B}
        \label{fig:ablation_4B_app}
    \end{subfigure}\\
    \begin{subfigure}[b]{0.49\linewidth}
    	\centering
    	\includegraphics[width=\linewidth]{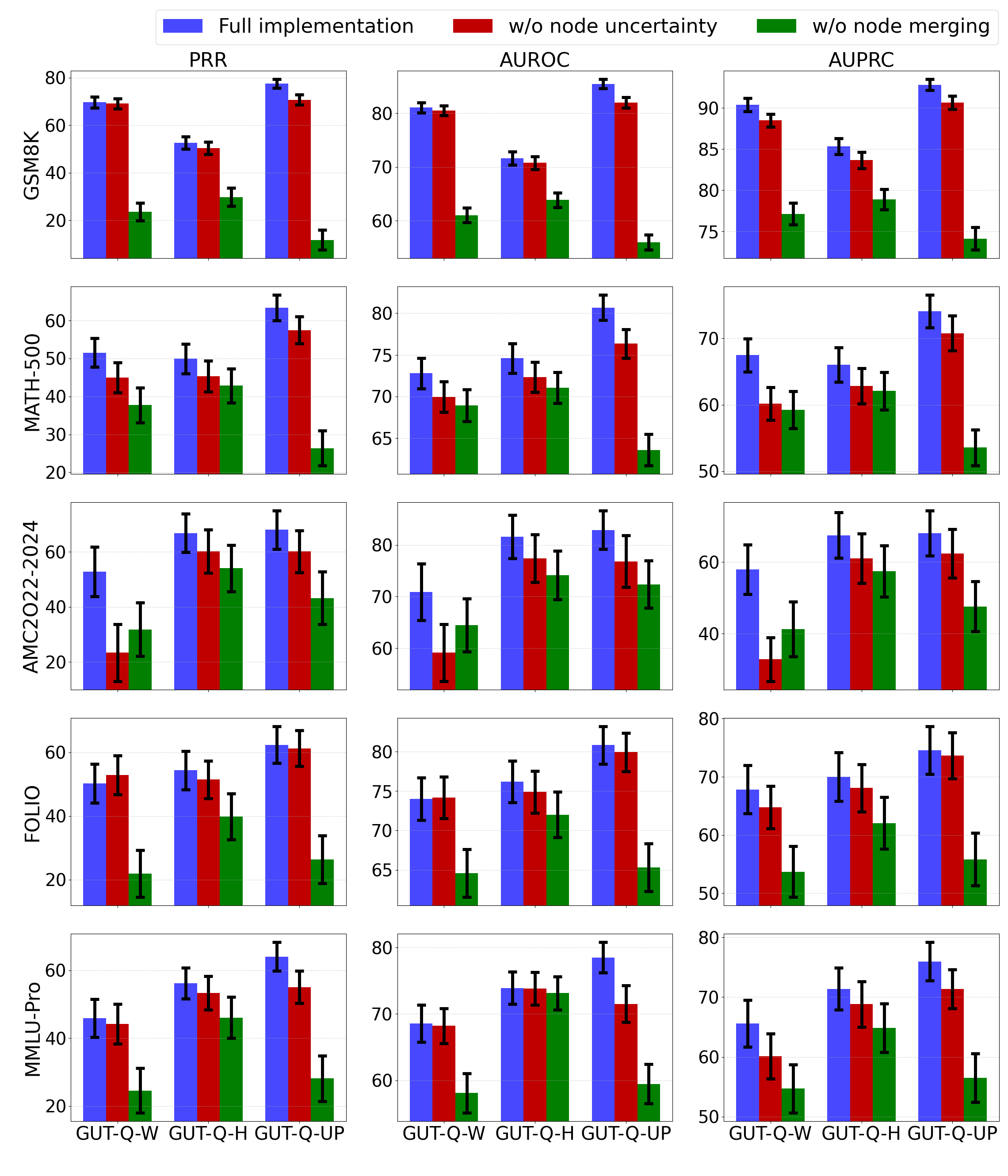}
    	\caption{Qwen3-1.7B}
        \label{fig:ablation_1.7B_app}
    \end{subfigure}
    \hfill
    \begin{subfigure}[b]{0.49\linewidth}
    	\centering
    	\includegraphics[width=\linewidth]{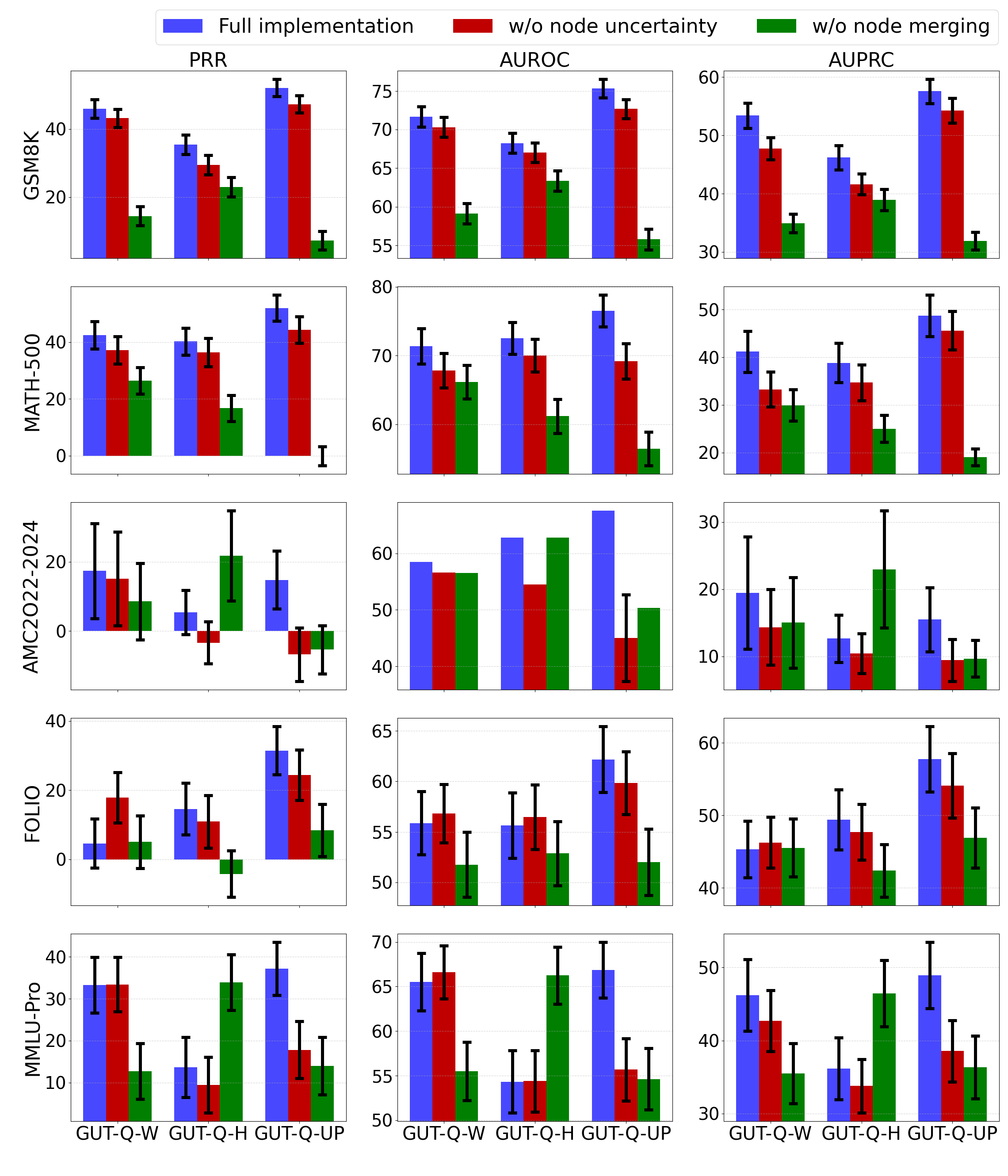}
    	\caption{Qwen3-0.6B}
        \label{fig:ablation_0.6B_app}
    \end{subfigure}
    \caption{Ablation comparisons of UQ performance.}
    \label{fig:ablation_8B4B1.7B0.6B_app}
\end{figure}

Figures~\ref{fig:ablation_8B_app},~\ref{fig:ablation_4B_app} ,~\ref{fig:ablation_1.7B_app}, and~\ref{fig:ablation_0.6B_app} show the ablation comparisons of UQ performance for Qwen3-8B, Qwen3-4B, Qwen3-1.7B, and Qwen3-0.6B, respectively. It is observed that, compared to the blue bars, the red bars are closer while the green bars are notably shorter for the UP methods. This observation demonstrates that node merging is more important than node uncertainty in the UP algorithm. We also observe that the red and green bars are far shorter than the blue bars for both the GUT-Q-W and GUT-Q-H, indicating that both node uncertainty and node merging are essential for width and height calculations.

\clearpage
\subsection{Additional Sensitivity Analyses} \label{app:subsec:sensitivity}
This subsection analyzes how the UQ performance is affected by the number of samples $K$, temperature $T$, number of shots $F$, and node merging criterion for Qwen3-8B, Qwen3-4B, Qwen3-1.7B, and Qwen3-0.6B. Based on these analyses, we recommend proper configurations of hyperparameters and node merging criteria.

\begin{figure}[htb]
    \centering
    \begin{subfigure}[b]{0.49\linewidth}
        \centering
        \includegraphics[width=\linewidth]{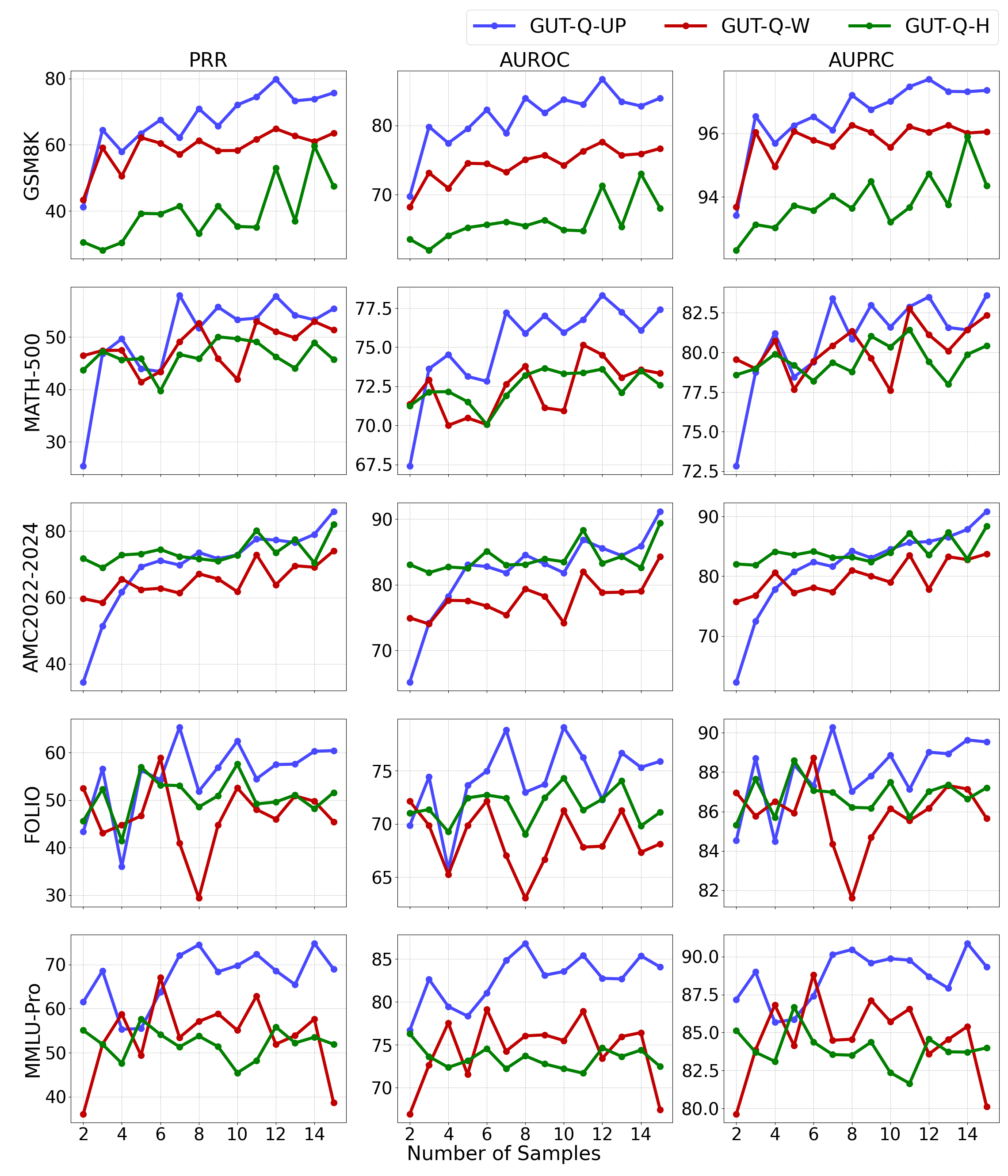}
        \caption{Qwen3-8B}
        \label{fig:num_of_samples_8B_app}
    \end{subfigure}
    \hfill
    \begin{subfigure}[b]{0.49\linewidth}
        \centering
        \includegraphics[width=\linewidth]{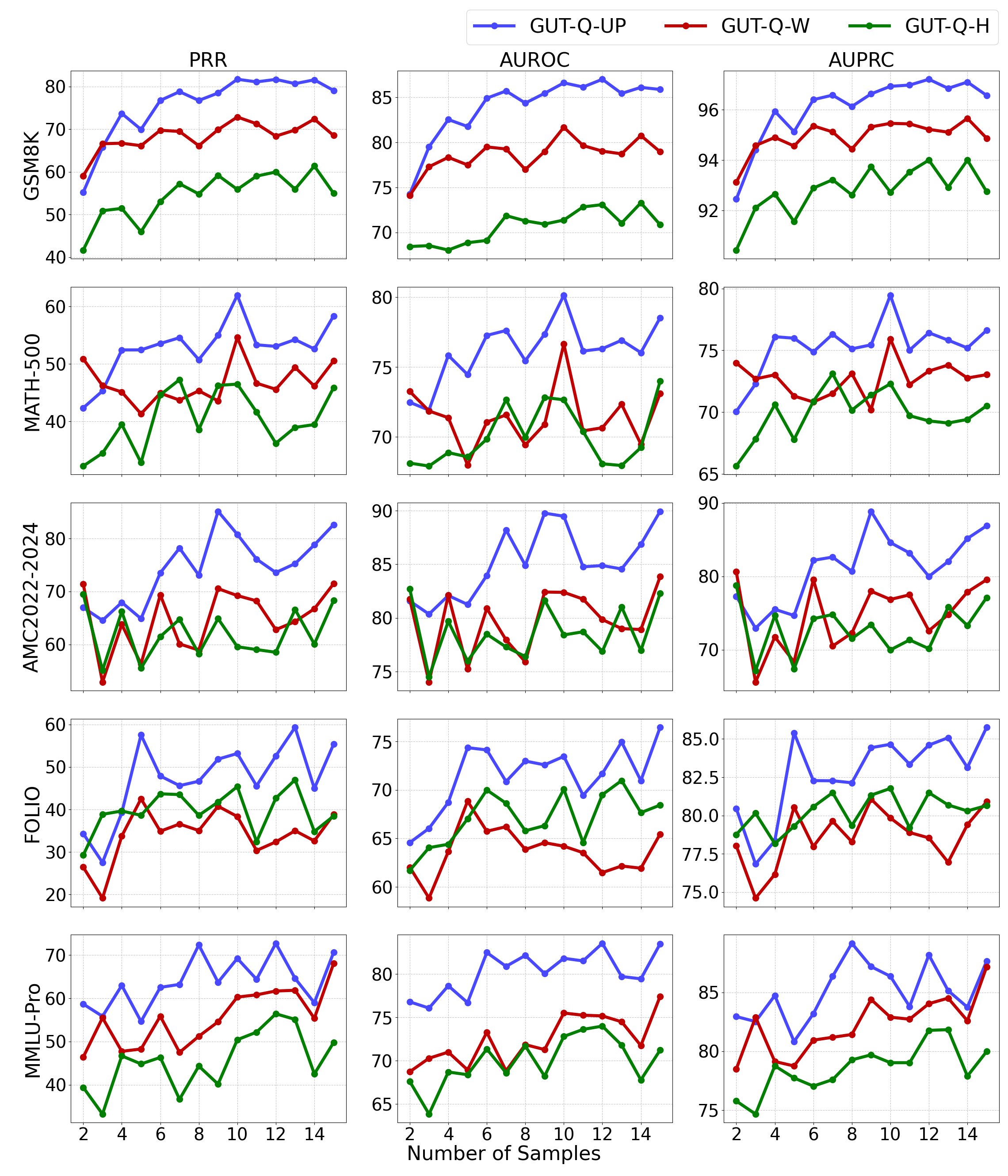}
        \caption{Qwen3-4B}
        \label{fig:num_of_samples_4B_app}
    \end{subfigure}\\
    \begin{subfigure}[b]{0.49\linewidth}
        \centering
        \includegraphics[width=\linewidth]{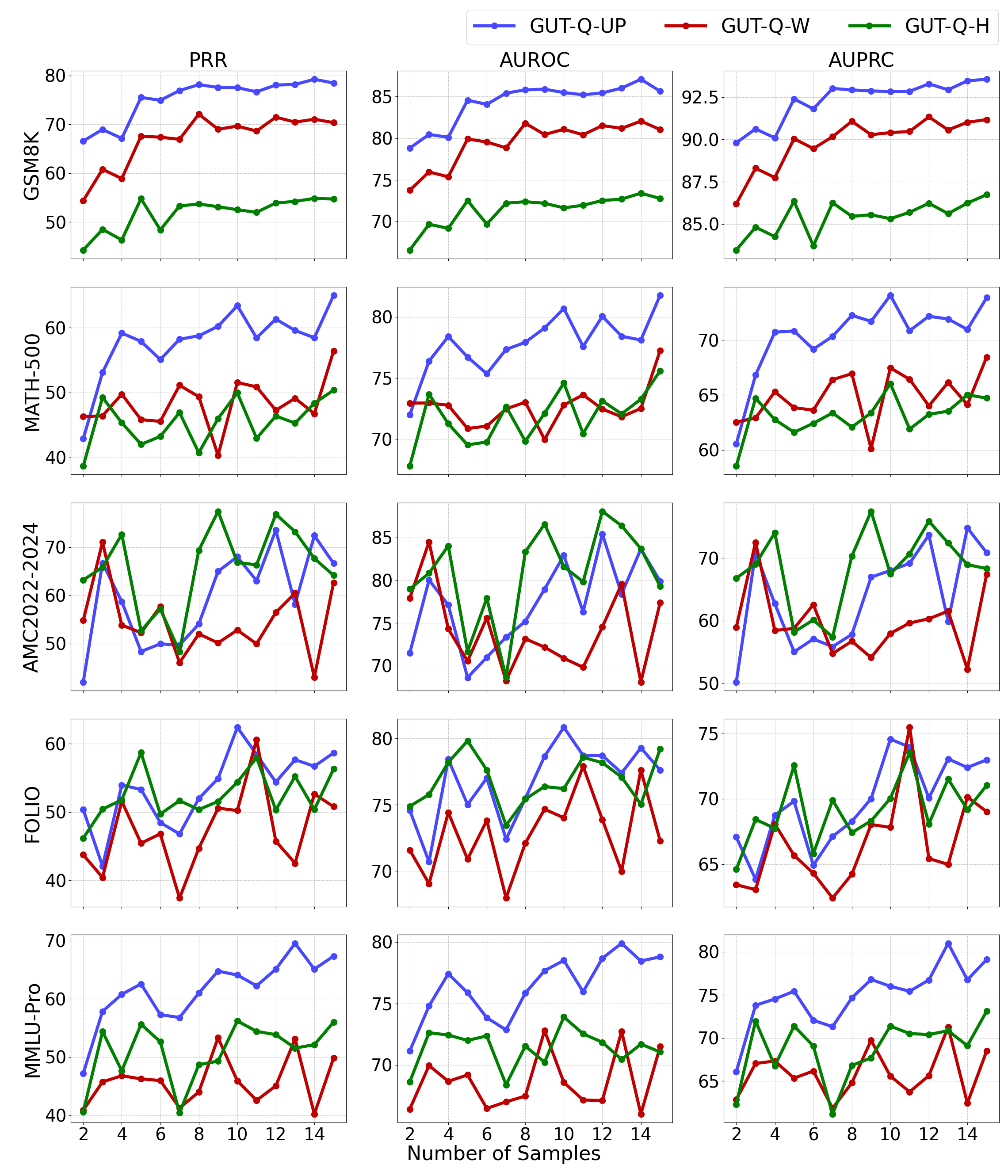}
        \caption{Qwen3-1.7B}
        \label{fig:num_of_samples_1.7B_app}
    \end{subfigure}
    \hfill
    \begin{subfigure}[b]{0.49\linewidth}
        \centering
        \includegraphics[width=\linewidth]{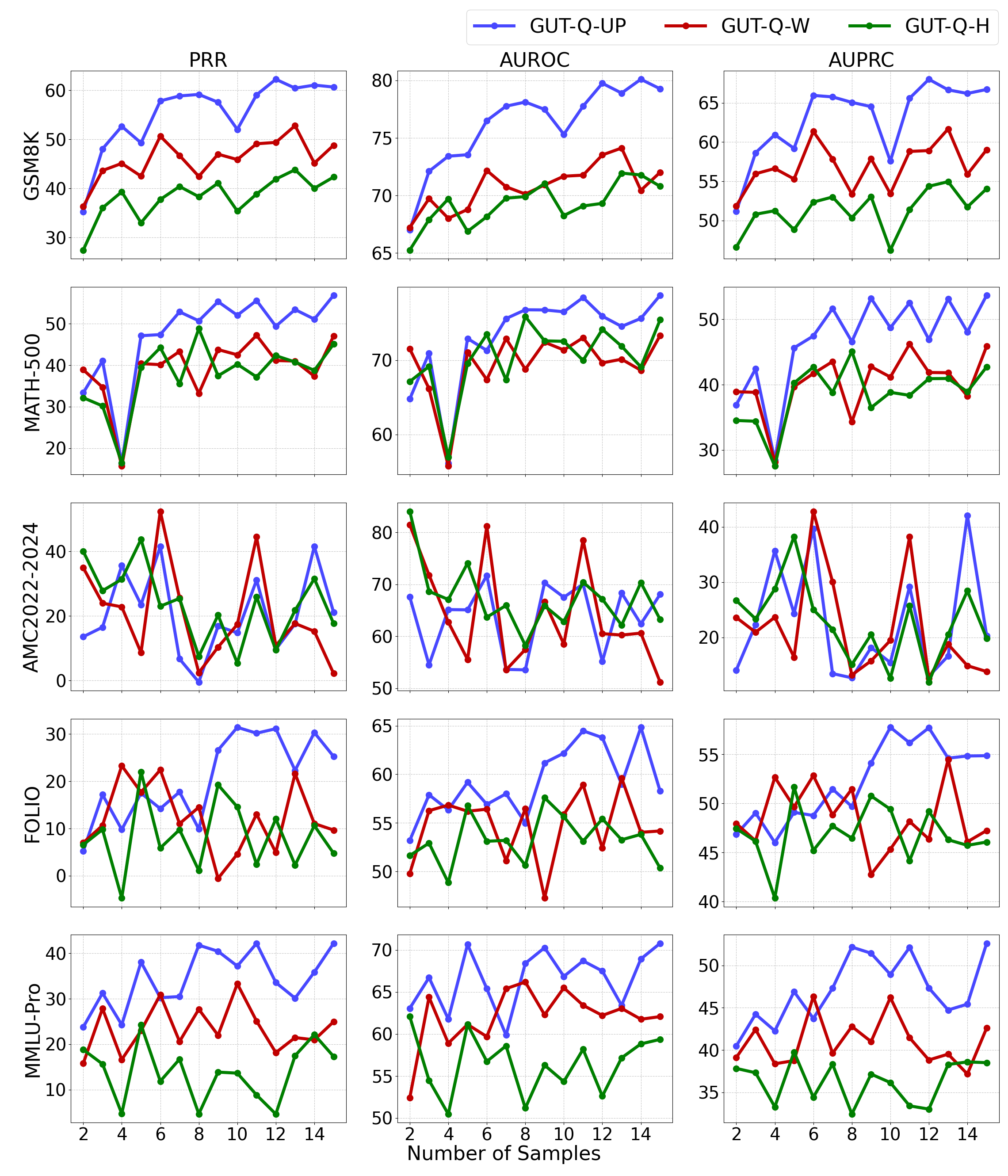}
        \caption{Qwen3-0.6B}
        \label{fig:num_of_samples_0.6B_app}
    \end{subfigure}
    \caption{Impact of the number of samples $K$ on UQ performance.}
\end{figure}

\paragraph{Number of Samples $K$} Figures~\ref{fig:num_of_samples_8B_app},~\ref{fig:num_of_samples_4B_app},~\ref{fig:num_of_samples_1.7B_app}, and~\ref{fig:num_of_samples_0.6B_app} illustrate the impact of the number of samples $K$ on UQ performance for Qwen3-8B, Qwen3-4B, Qwen3-1.7B, and Qwen3-0.6B, respectively. We recommend $K=9$ for Qwen3-4B, $K=12$ for Qwen3-1.7B, and $K=11$ for Qwen3-8B and Qwen3-0.6B to balance efficiency and performance since a larger $K$ leads to higher computational costs.

\begin{figure}[htb]
    \centering
    \begin{subfigure}[b]{0.41\linewidth}
        \centering
        \includegraphics[width=\linewidth]{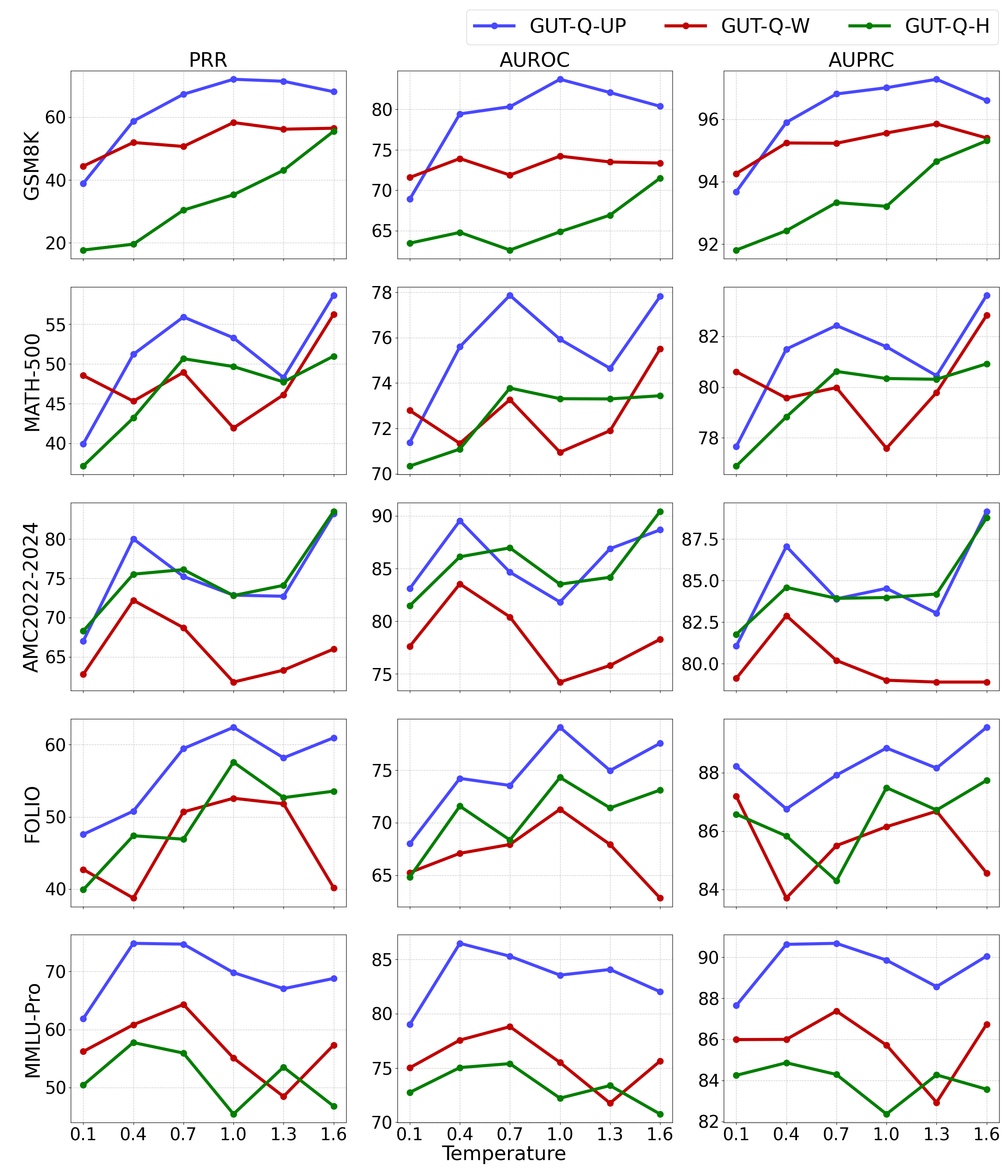}
        \caption{Qwen3-8B}
        \label{fig:temperature_8B_app}
    \end{subfigure}
    \begin{subfigure}[b]{0.41\linewidth}
        \centering
        \includegraphics[width=\linewidth]{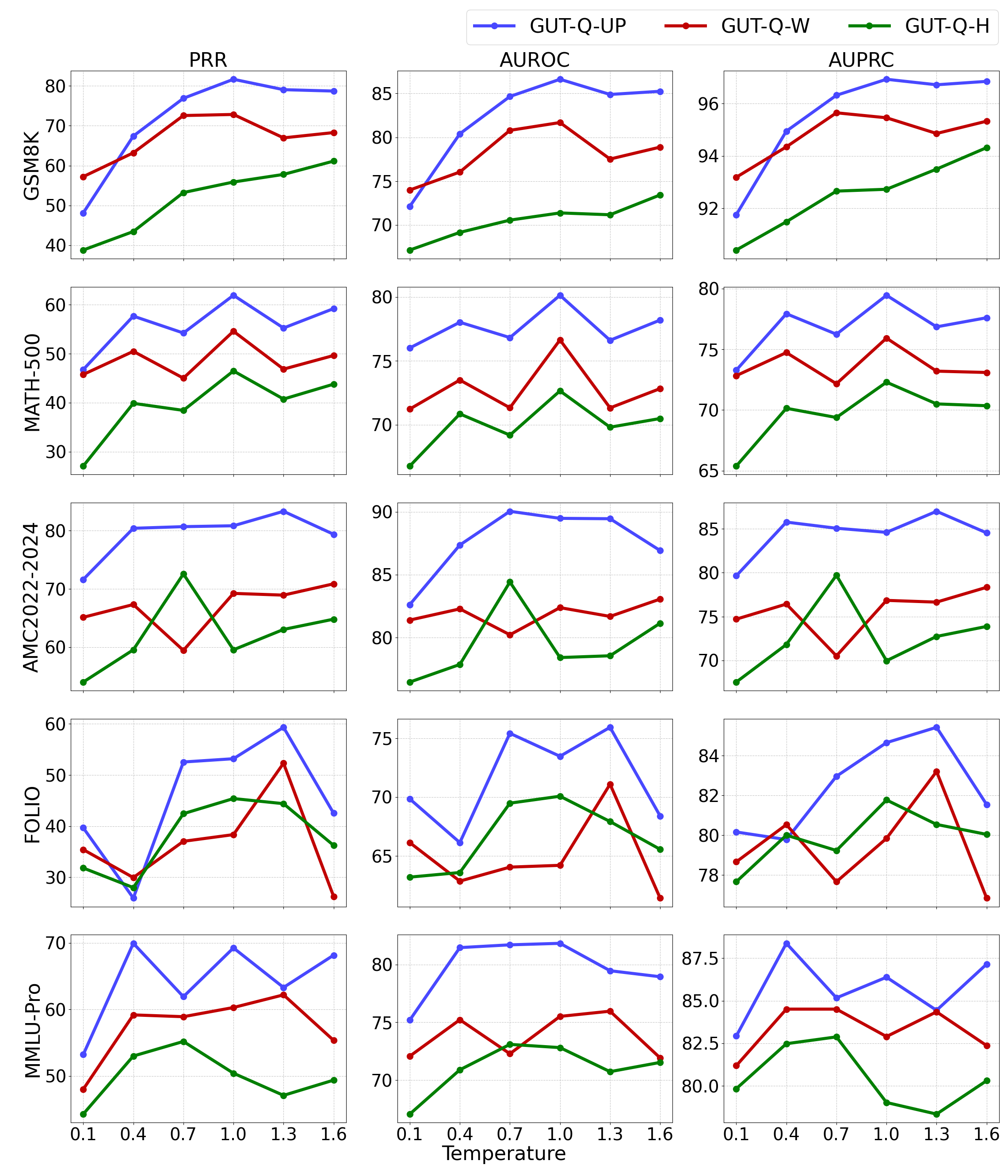}
        \caption{Qwen3-4B}
        \label{fig:temperature_4B_app}
    \end{subfigure}\\
        \begin{subfigure}[b]{0.41\linewidth}
        \centering
        \includegraphics[width=\linewidth]{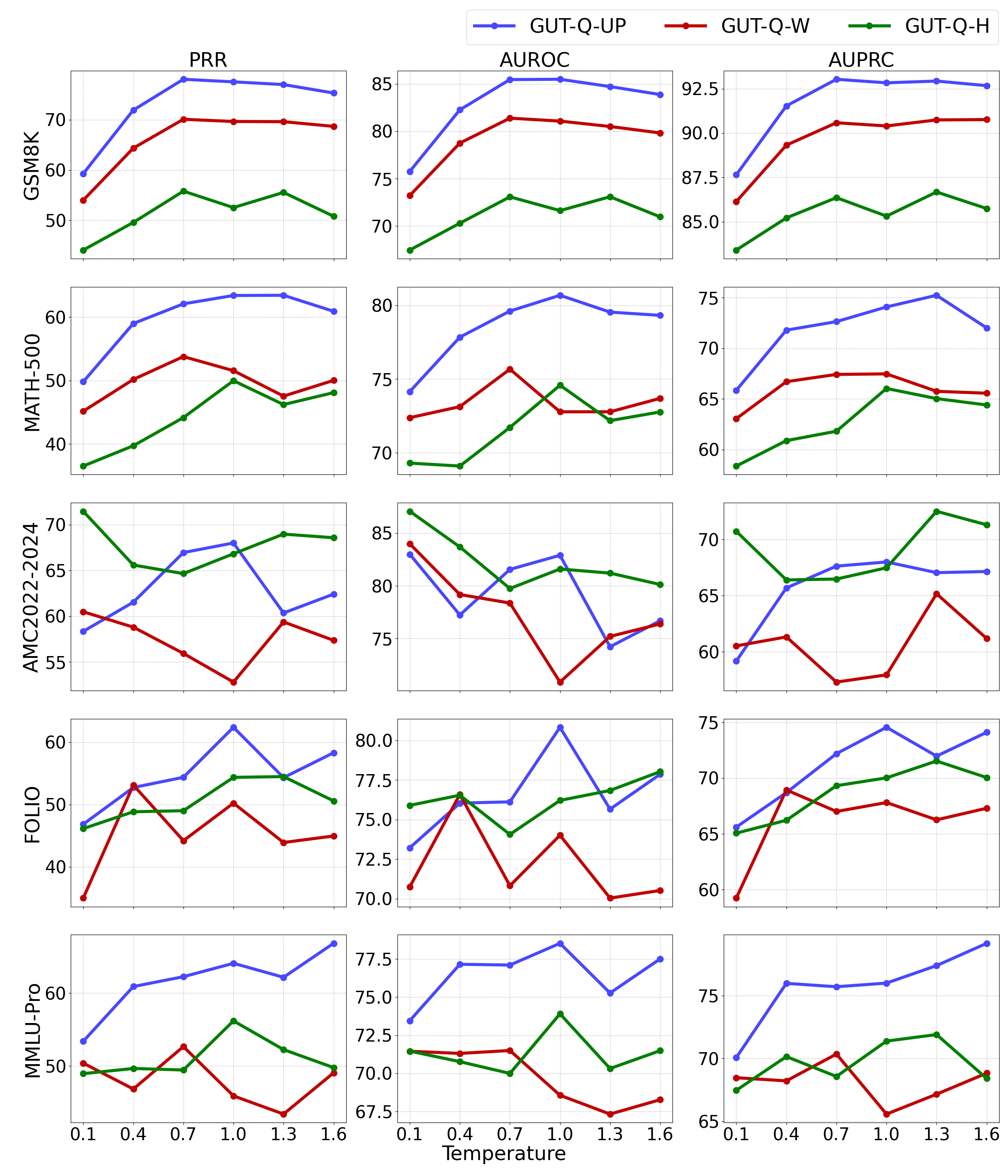}
        \caption{Qwen3-1.7B}
        \label{fig:temperature_1.7B_app}
    \end{subfigure}
    \begin{subfigure}[b]{0.41\linewidth}
        \centering
        \includegraphics[width=\linewidth]{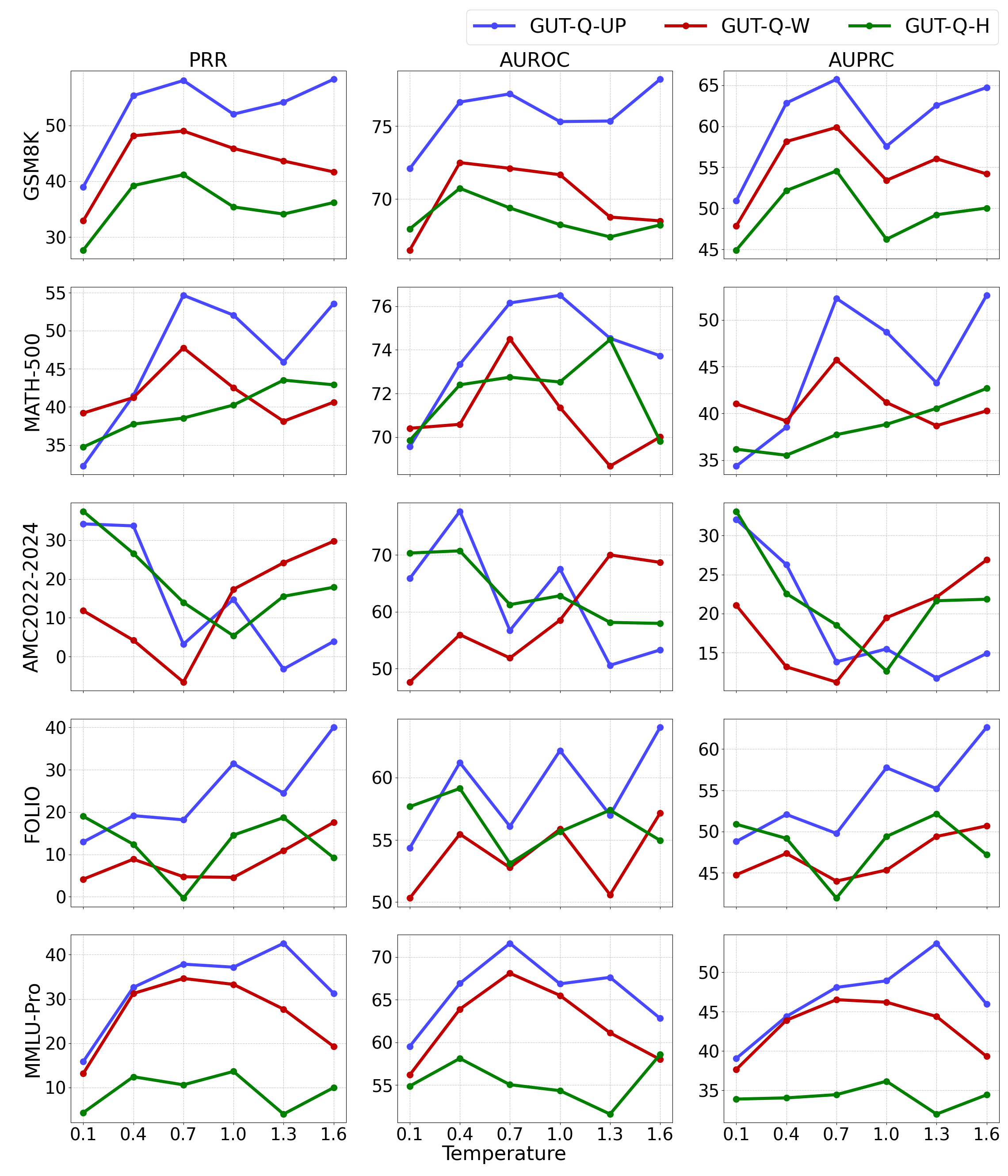}
        \caption{Qwen3-0.6B}
        \label{fig:temperature_0.6B_app}
    \end{subfigure}
    \caption{Impact of the temperature $T$ on UQ performance.}
\end{figure}

\clearpage
\paragraph{Temperature $T$} Figures~\ref{fig:temperature_8B_app},~\ref{fig:temperature_4B_app},~\ref{fig:temperature_1.7B_app}, and~\ref{fig:temperature_0.6B_app} illustrate the impact of the temperature $T$ on UQ performance for Qwen3-8B, Qwen3-1.7B, and Qwen3-0.6B, respectively. We recommend $T=1.6$ for Qwen3-8B, $T=1.0$ for Qwen3-4B, and $T=0.7$ for both Qwen3-1.7B and Qwen3-0.6B based on the UQ performance.

\begin{figure}[htb]
    \centering
    \begin{subfigure}[b]{0.41\linewidth}
        \centering
        \includegraphics[width=\linewidth]{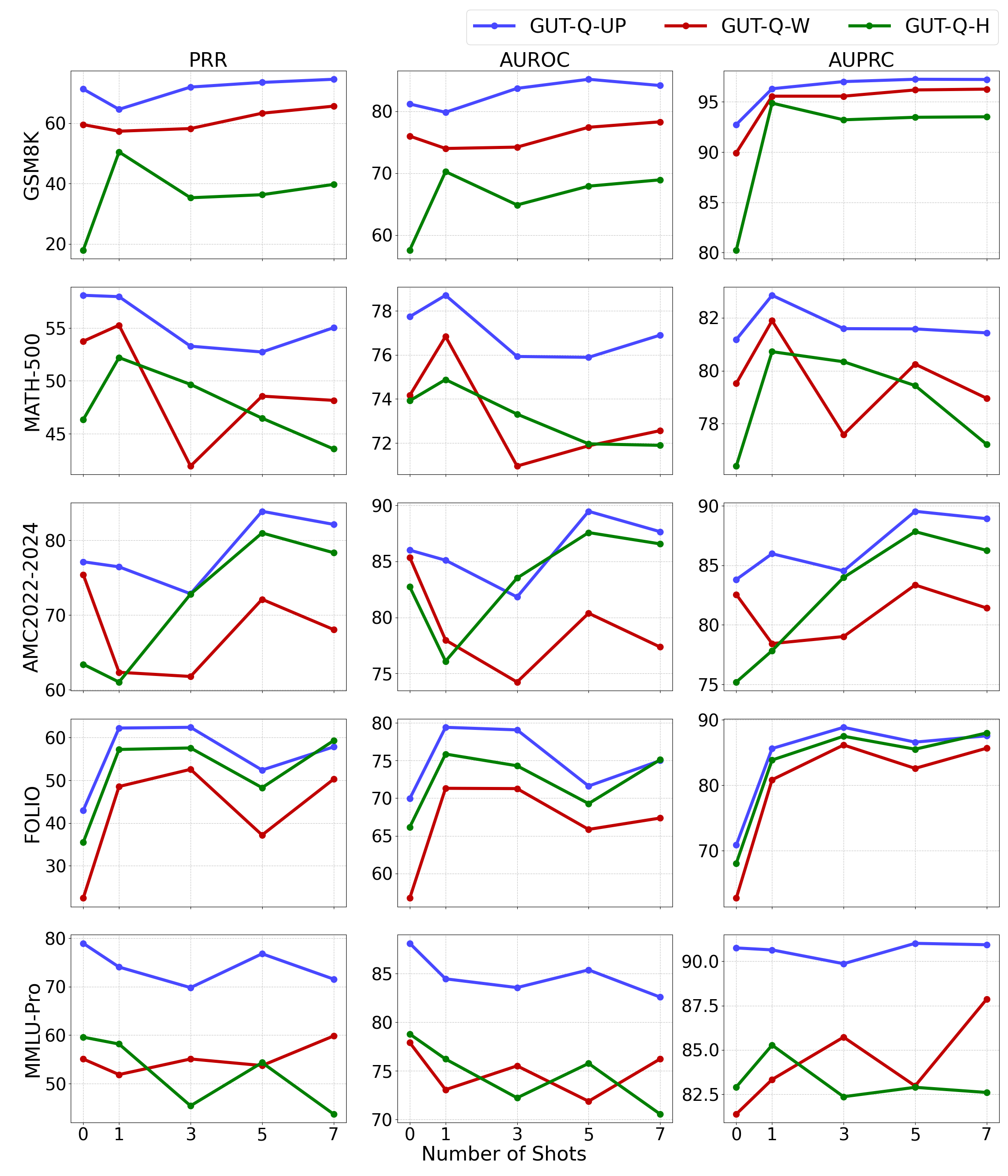}
        \caption{Qwen3-8B}
        \label{fig:num_of_shots_8B_app}
    \end{subfigure}
    \begin{subfigure}[b]{0.41\linewidth}
        \centering
        \includegraphics[width=\linewidth]{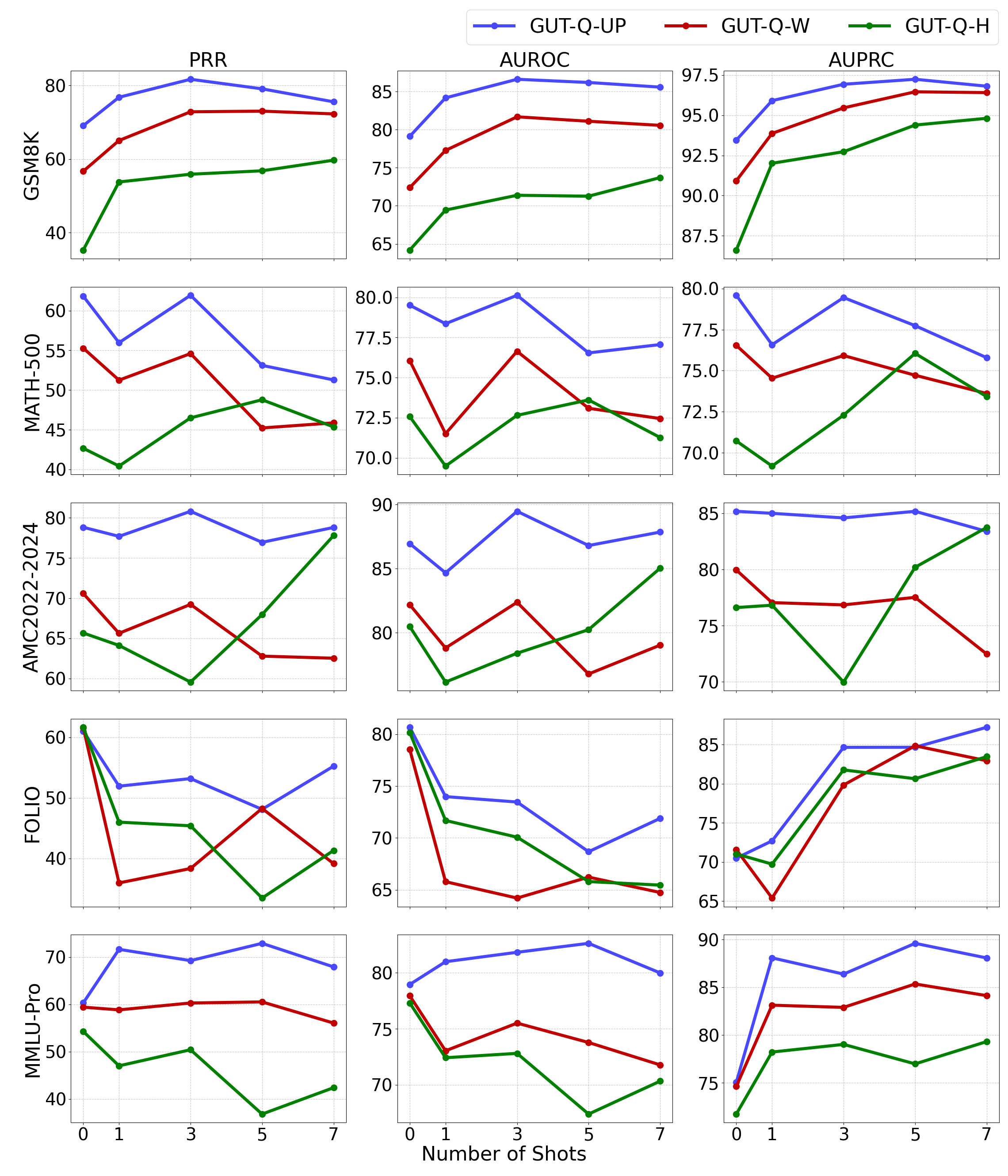}
        \caption{Qwen3-4B}
        \label{fig:num_of_shots_4B_app}
    \end{subfigure}\\
    \begin{subfigure}[b]{0.41\linewidth}
        \centering
        \includegraphics[width=\linewidth]{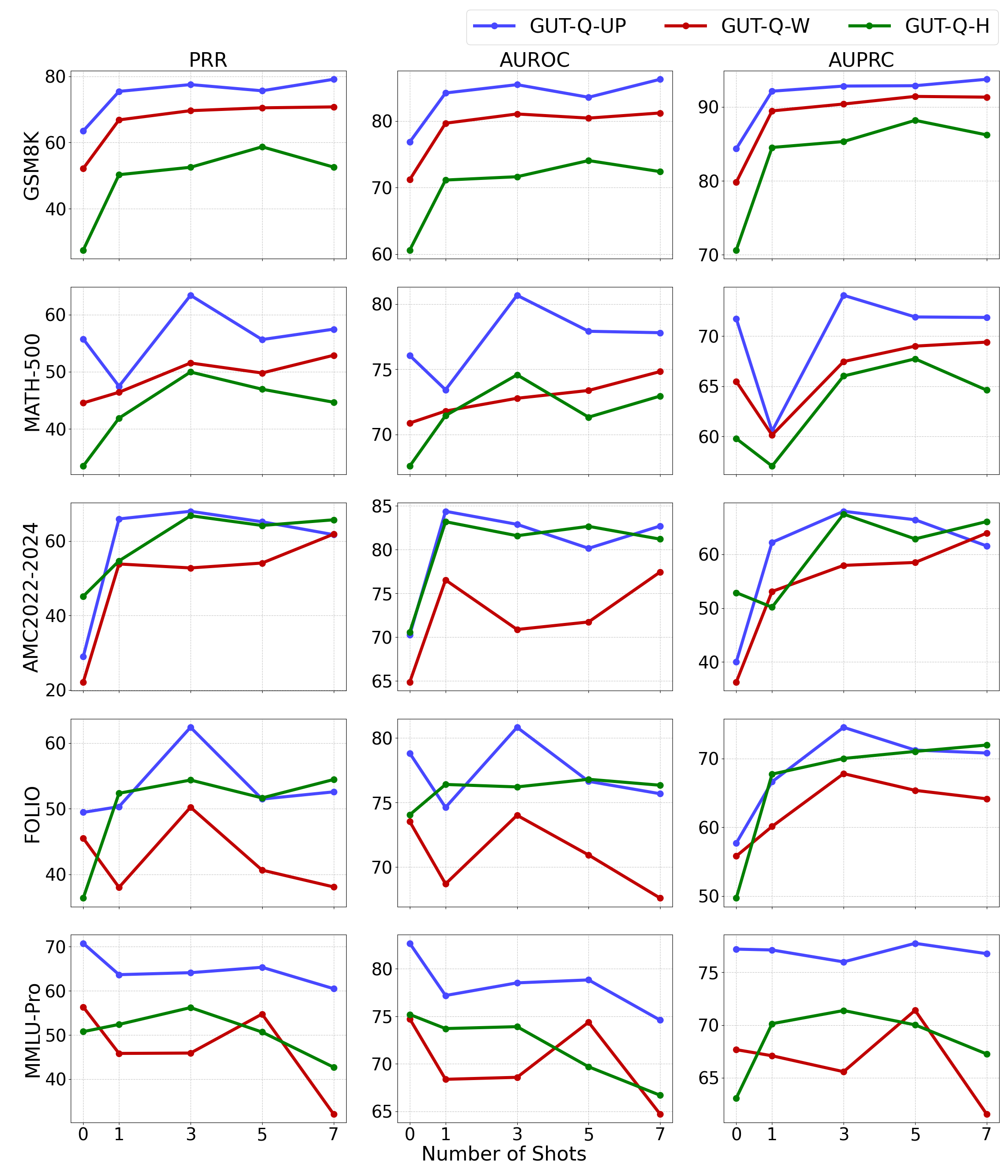}
        \caption{Qwen3-1.7B}
        \label{fig:num_of_shots_1.7B_app}
    \end{subfigure}
    \begin{subfigure}[b]{0.41\linewidth}
        \centering
        \includegraphics[width=\linewidth]{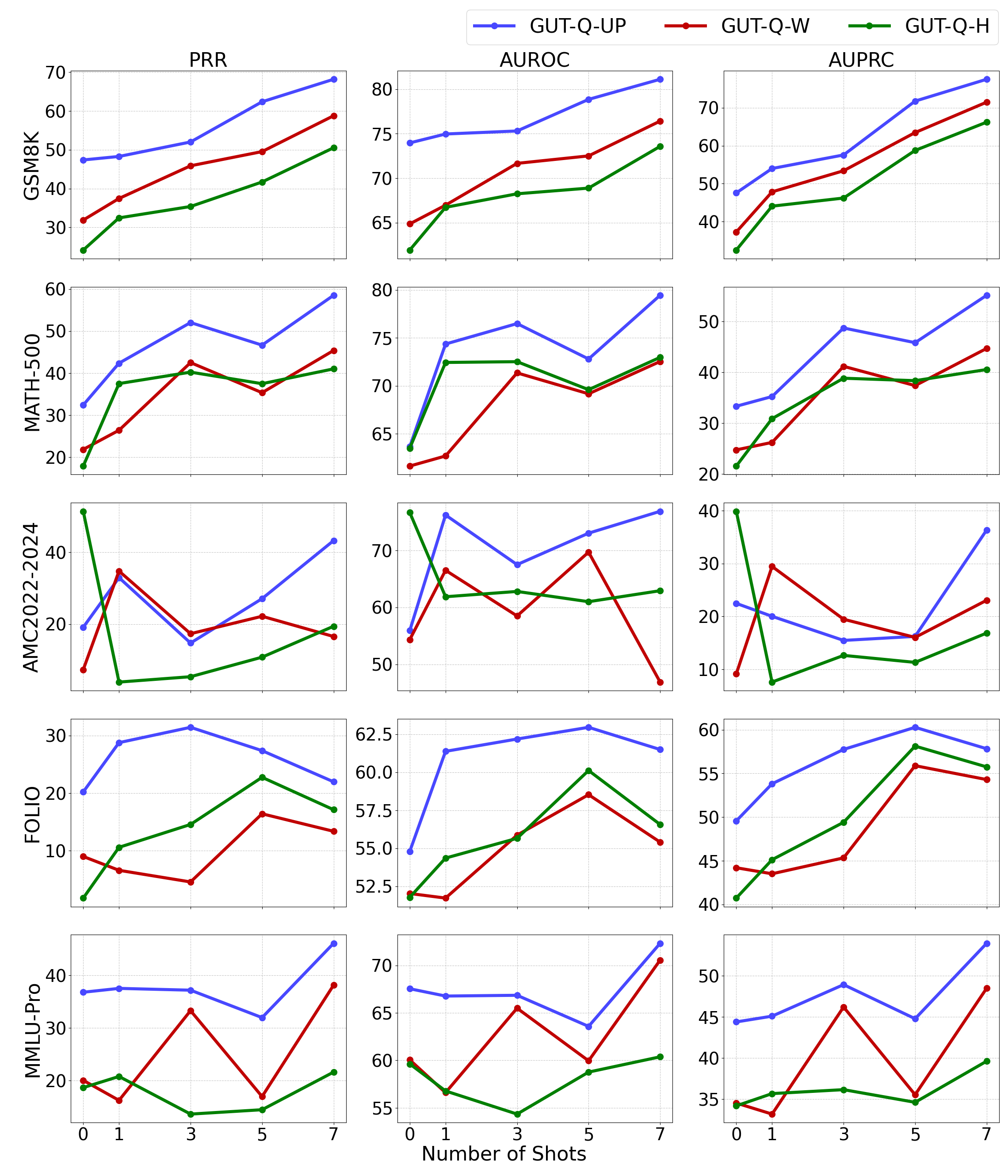}
        \caption{Qwen3-0.6B}
        \label{fig:num_of_shots_0.6B_app}
    \end{subfigure}
    \caption{Impact of the number of shots $F$ on UQ performance.}
\end{figure}

\clearpage
\paragraph{Number of Shots $F$} Figures~\ref{fig:num_of_shots_8B_app},~\ref{fig:num_of_shots_4B_app},~\ref{fig:num_of_shots_1.7B_app}, and~\ref{fig:num_of_shots_0.6B_app} illustrate the impact of the number of shots $F$ on UQ performance for Qwen3-8B, Qwen3-1.7B, and Qwen3-0.6B, respectively. We recommend $F=5$ for Qwen3-8B and Qwen3-4B, $F=3$ for Qwen3-1.7B, and $F=7$ for Qwen3-0.6B based on the UQ performance.

\clearpage
\paragraph{Node Merging Criterion} Following~\citep{lin2024generating}, we conducted sensitivity analyses on the criterion for determining equivalence with the NLI model. Specifically, we investigate the following three criteria in Step 15 of Algorithm~\ref{alg:AOV} to determine whether two nodes, denoted as $\mathscr{N}_1$ and $\mathscr{N}_2$, are semantically equivalent.
\begin{itemize}
    \item \textbf{Bi-entailment.} $\mathscr{N}_1$ entails $\mathscr{N}_2$ and $\mathscr{N}_2$ entails $\mathscr{N}_1$, and they do not contradict each other.
    \item \textbf{Uni-entailment.} $\mathscr{N}_1$ entails $\mathscr{N}_2$ or $\mathscr{N}_2$ entails $\mathscr{N}_1$, and they do not contradict each other.
    \item \textbf{Non-contradiction.} $\mathscr{N}_1$ and $\mathscr{N}_2$ do not contradict each other.
\end{itemize}

\begin{figure}[htb]
    \centering
    \begin{subfigure}[b]{0.4\textwidth}
        \centering
        \includegraphics[width=\linewidth]{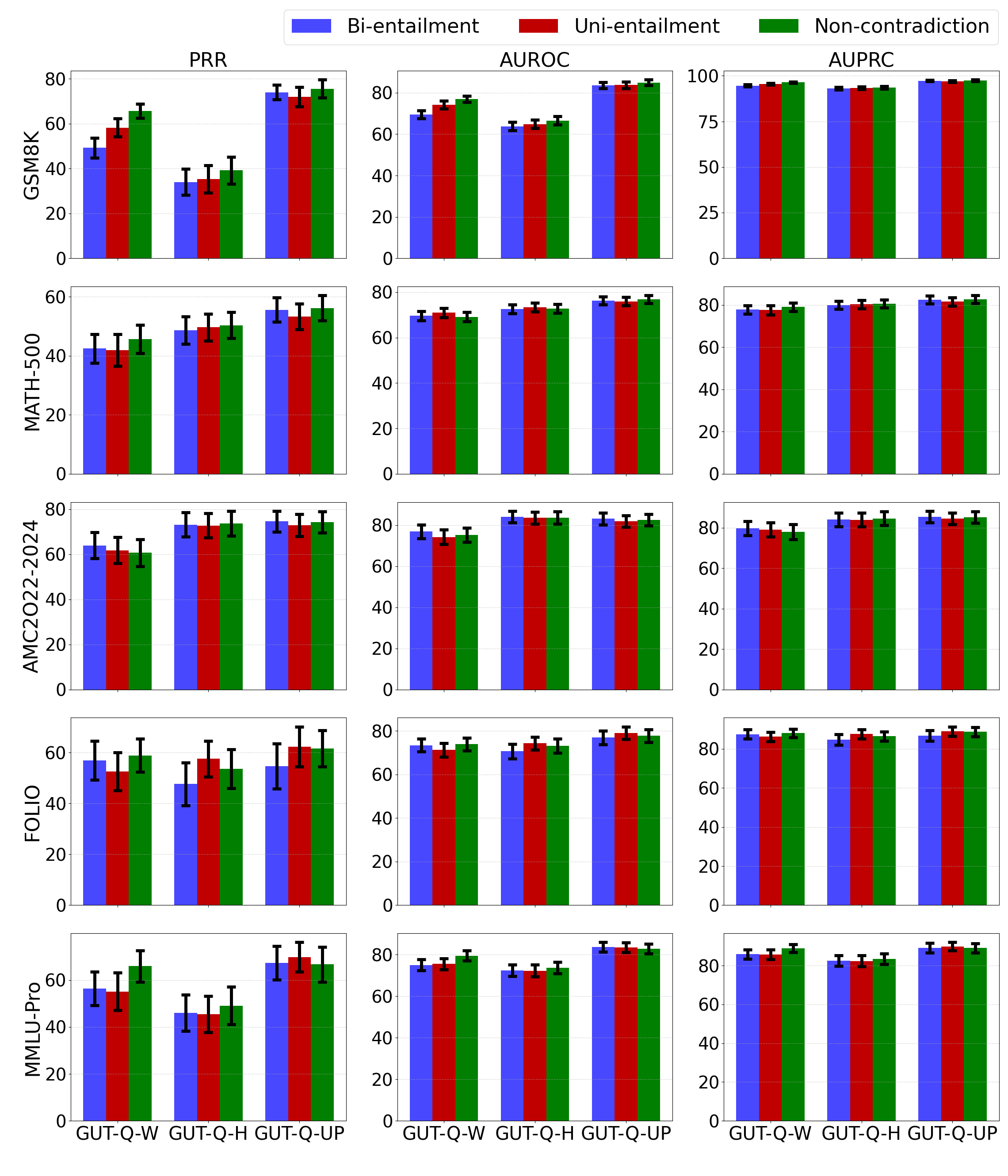}
        \caption{Qwen3-8B}
        \label{fig:nli_8B_app}
    \end{subfigure}
    \begin{subfigure}[b]{0.4\textwidth}
        \centering
        \includegraphics[width=\linewidth]{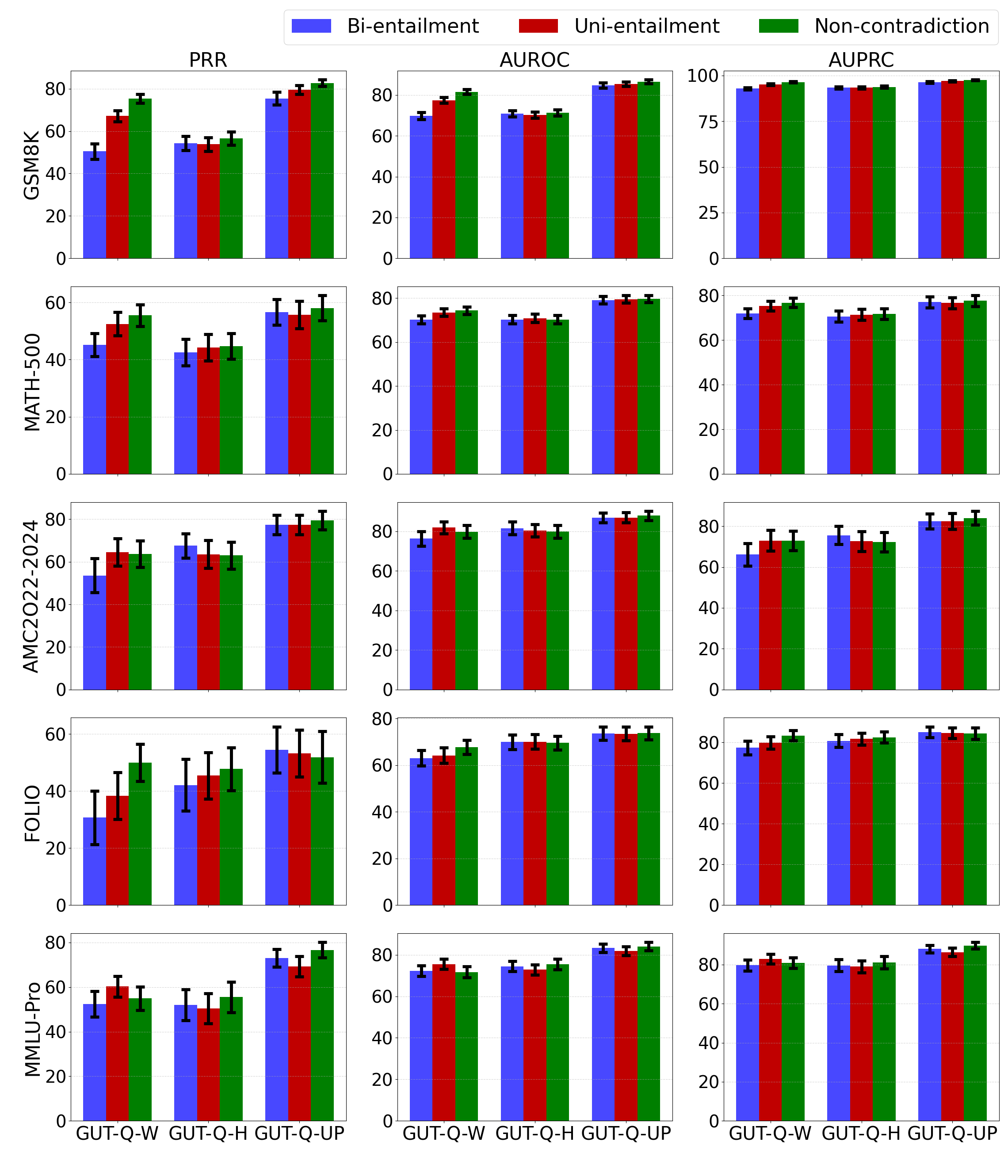}
        \caption{Qwen3-4B}
        \label{fig:nli_4B_app}
    \end{subfigure}\\
    \begin{subfigure}[b]{0.4\textwidth}
        \centering
        \includegraphics[width=\linewidth]{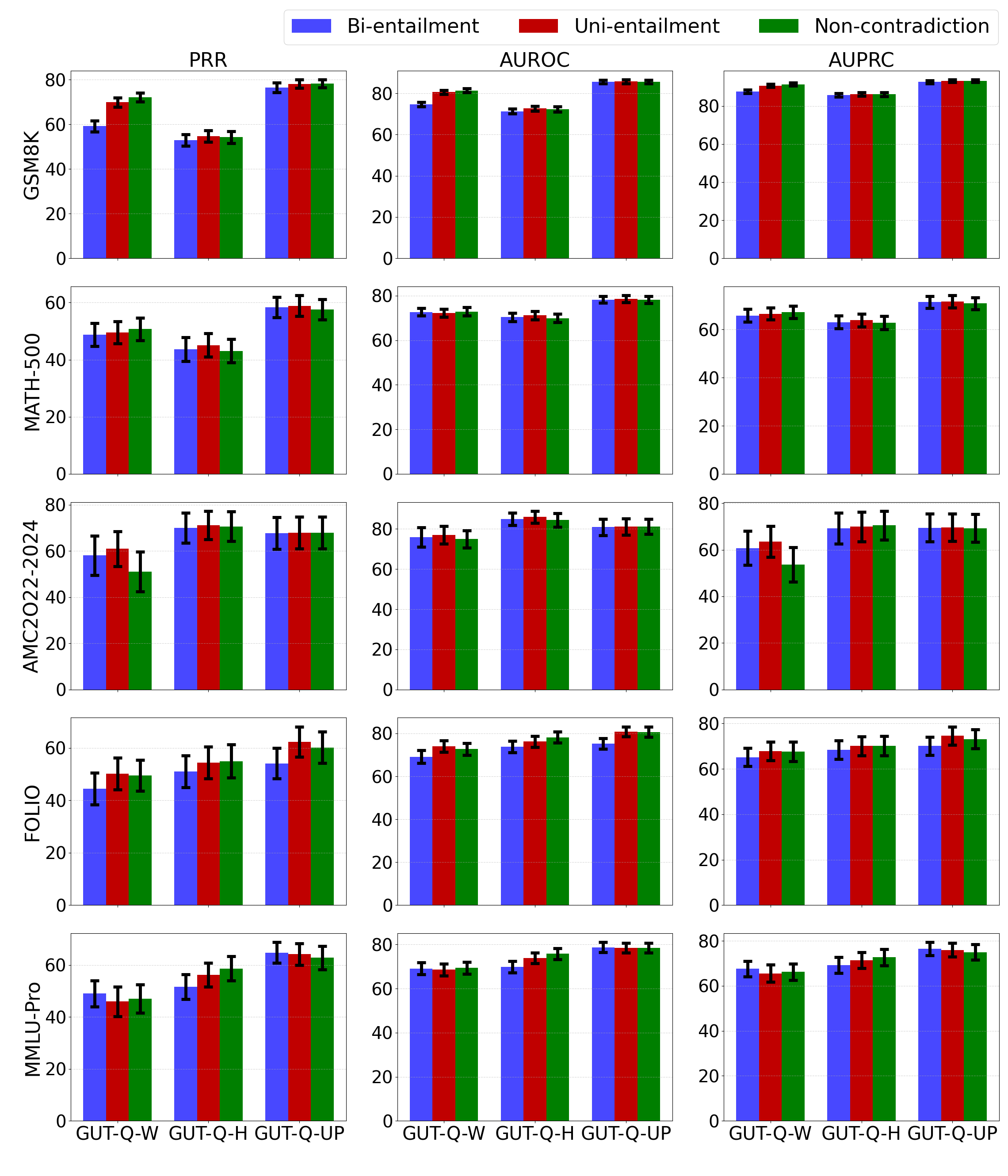}
        \caption{Qwen3-1.7B}
        \label{fig:nli_1.7B_app}
    \end{subfigure}
    \begin{subfigure}[b]{0.4\textwidth}
        \centering
        \includegraphics[width=\linewidth]{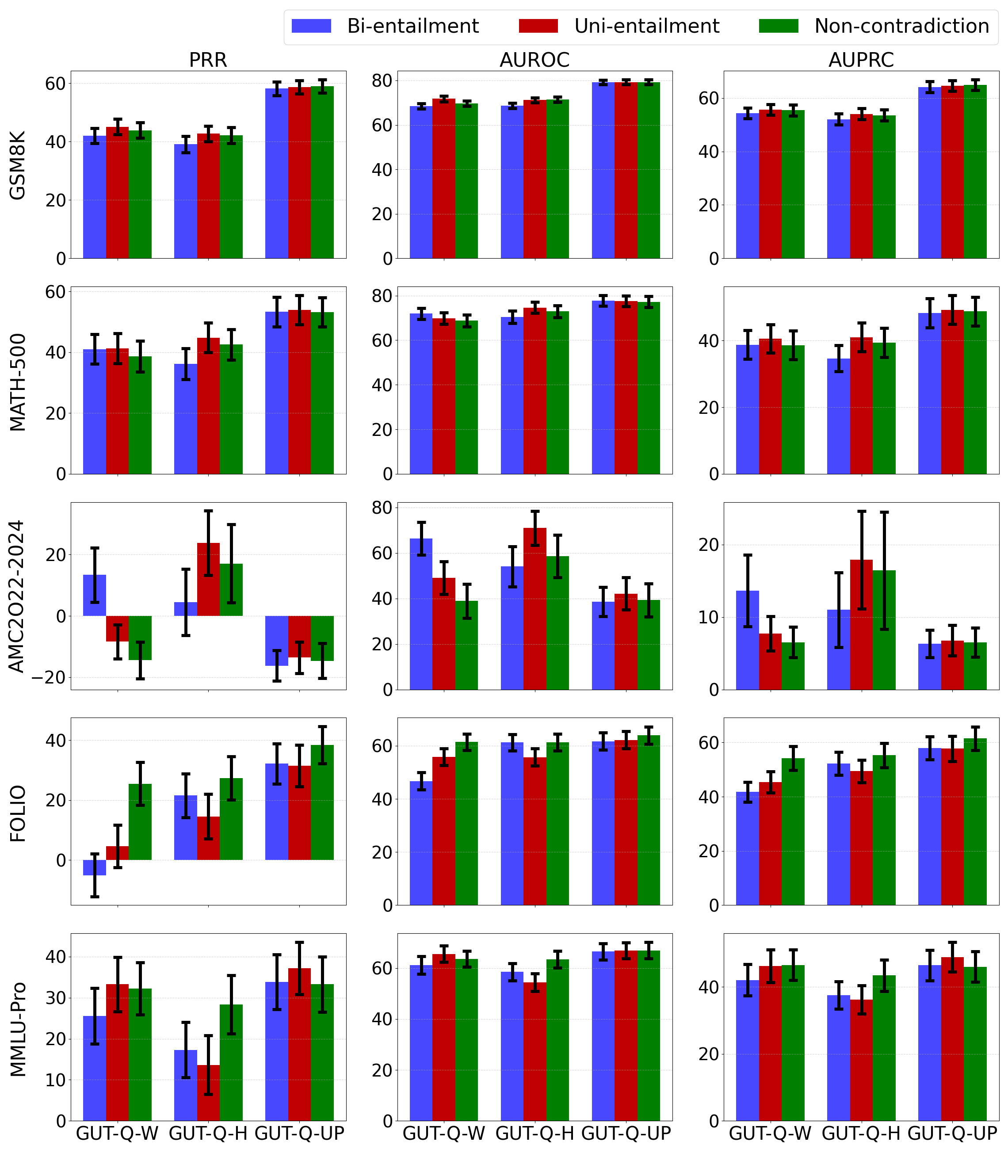}
        \caption{Qwen3-0.6B}
        \label{fig:nli_0.6B_app}
    \end{subfigure}
    \caption{Impact of the criterion for determining equivalence with the NLI model.}
    \label{fig:nli_comparison}
\end{figure}

Figures~\ref{fig:nli_8B_app},~\ref{fig:nli_4B_app},~\ref{fig:nli_1.7B_app}, and~\ref{fig:nli_0.6B_app} show the impact of the criterion for determining equivalence in the NLI model for Qwen3-8B, Qwen3-4B, Qwen3-1.7B, and Qwen3-0.6B, respectively, where the ``bi-entailment'' represents the original criterion in Algorithm~\ref{alg:AOV}. It is observed that the three criteria yield comparable UQ performance across LLM scales, datasets, and UQ evaluation metrics. Therefore, we can conclude that GUT-Q is robust against the choice of specific criteria of node merging.


\section{Case study of the GUT}  \label{app:case_study}
\begin{figure}[h]
    \centering
    \includegraphics[width=\linewidth]{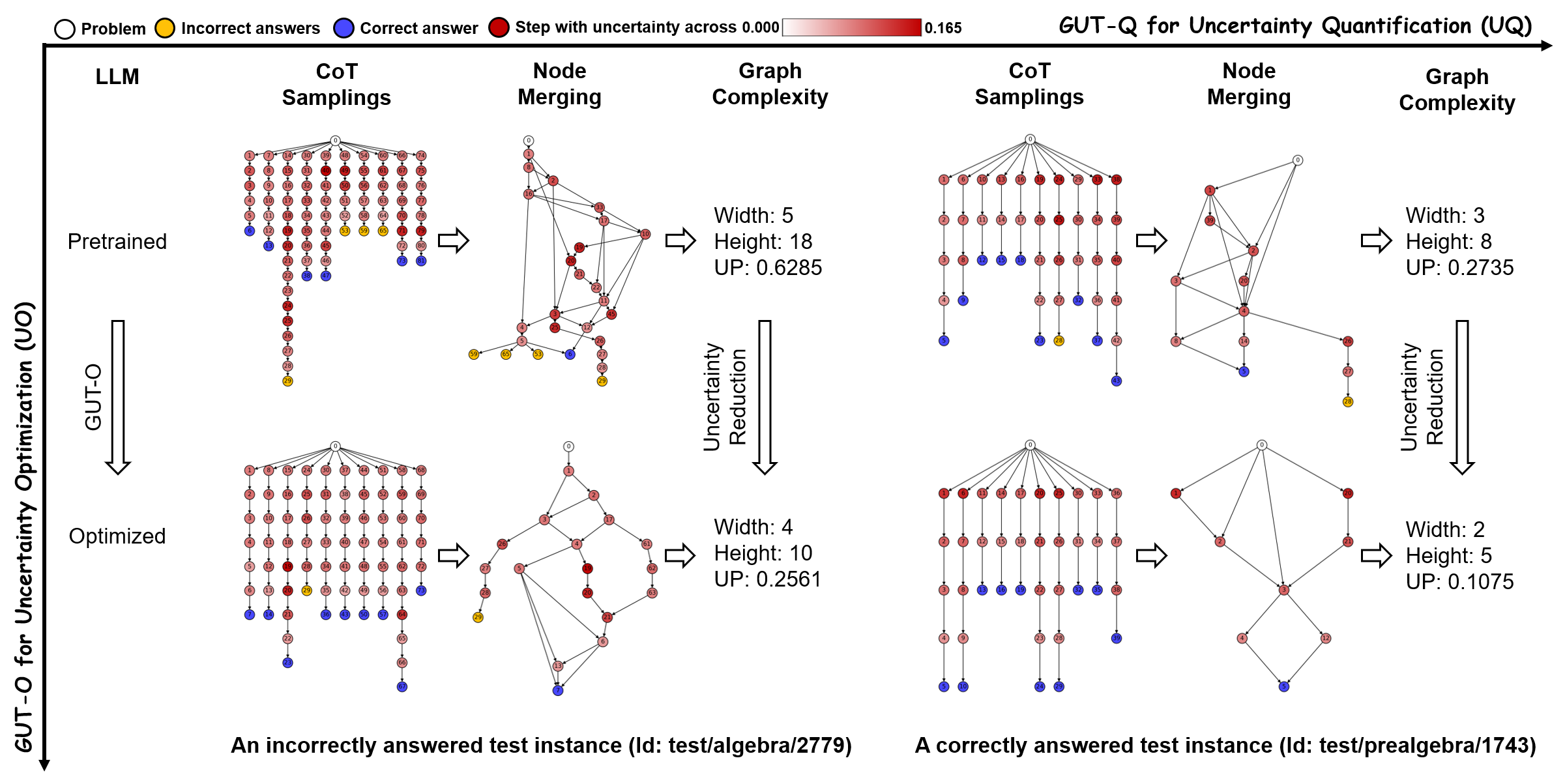}
    \caption{Application of our proposed GUT to two test instances from the MATH-500 dataset using Qwen3-4B. For illustrative purposes, the width and height are obtained by removing the node uncertainty in GUT-Q-W and GUT-Q-H, respectively.}
    \label{fig:case_study_app}
\end{figure}

This appendix presents a case study of instances from MATH-500 on Qwen3-4B to demonstrate our proposed GUT. Figure~\ref{fig:case_study_app} presents the application of our proposed GUT to two test instances in the MATH-500 dataset with Qwen3-4B, where the left part illustrates an incorrectly answered one and the right part illustrates a correctly answered one. We observe that the graph complexity metrics of the incorrectly answered instance are typically larger than those of the correctly answered one. This observation demonstrates the power of our proposed GUT-Q in discriminating between correct instances and incorrect ones. Moreover, it is obvious that the LLM optimized by GUT-O tends to exhibit fewer reasoning steps for the same question.

Now we detail the reasoning space represented by the constructed DAG for these four test instances as follows, where the nodes with boxes denote the final answers, such as $\boxed{1}$.
\begin{promptbox}[title=The test instance incorrectly answered by the pretrained Qwen3-4B.]
\textbf{Problem:} The number $(\sqrt{2}+\sqrt{3})^3$ can be written in the form $a\sqrt{2} + b\sqrt{3} + c\sqrt{6}$, where $a$, $b$, and $c$ are integers.  What is $a+b+c$? \\
\textbf{Nodes within the constructed DAG:} \\
\textbf{Topological Level=1:} \\
Node 1: Expand $(\sqrt{2}+\sqrt{3})^3$ using the binomial theorem. \\
\textbf{Topological Level=2:} \\
Node 8: APply the binomial theorem: $(\sqrt{2}+\sqrt{3})^3 = \binom{3}{0}(\sqrt{2})^3(\sqrt{3})^0 + \binom{3}{1}(\sqrt{2})^2(\sqrt{3})^1 + \binom{3}{2}(\sqrt{2})^1(\sqrt{3})^2 + \binom{3}{3}(\sqrt{2})^0(\sqrt{3})^3$. \\
\textbf{Topological Level=3:} \\
Node 2: Compute each term of the expansion. \\
\textbf{Topological Level=4:} \\
Node 16 :  Simplify each term: $2\sqrt{2} + 3 \cdot 2 \cdot \sqrt{3} + 3 \cdot \sqrt{2} \cdot 3 + 3\sqrt{3}$. \\
\textbf{Topological Level=5:} \\
Node 33 :  Calculate the coefficients: $1(2\sqrt{2}) + 3(2\sqrt{3}) + 3(3\sqrt{2}) + 1(3\sqrt{3})$. \\
\textbf{Topological Level=6:} \\
Node 17 :  Combine like terms: $2\sqrt{2} + 6\sqrt{3} + 9\sqrt{2} + 3\sqrt{3}$.  \\
\textbf{Topological Level=7:} \\
Node 10 :  Combine like terms: $2\sqrt{2} + 9\sqrt{2} = 11\sqrt{2}$, $6\sqrt{3} + 3\sqrt{3} = 9\sqrt{3}$. \\
\textbf{Topological Level=8:} \\
Node 19 :  However, this result is missing the $\sqrt{6}$ term, which suggests that I made a mistake in the expansion. \\
\textbf{Topological Level=9:} \\
Node 20 :  Re-evaluating the binomial expansion, the correct expansion should be $(\sqrt{2})^3 + 3(\sqrt{2})^2(\sqrt{3}) + 3(\sqrt{2})(\sqrt{3})^2 + (\sqrt{3})^3$. \\
\textbf{Topological Level=10:} \\
Node 21 :  Correctly simplifying each term: $2\sqrt{2} + 3 \cdot 2 \cdot \sqrt{3} + 3 \cdot \sqrt{2} \cdot 3 + 3\sqrt{3}$. \\
\textbf{Topological Level=11:} \\
Node 22 :  This simplifies to $2\sqrt{2} + 6\sqrt{3} + 9\sqrt{2} + 3\sqrt{3}$. \\
\textbf{Topological Level=12:} \\
Node 11 :  Therefore, $(\sqrt{2}+\sqrt{3})^3 = 11\sqrt{2} + 9\sqrt{3} + 0\sqrt{6}$. \\
\textbf{Topological Level=13:} \\
Node 3 :  Combine like terms to express the result in the form $a\sqrt{2} + b\sqrt{3} + c\sqrt{6}$. \\
Node 45 :  Recognize that the expression is missing the $\sqrt{6}$ term, so $c=0$. \\
\end{promptbox}

\begin{promptbox}[title=The test instance incorrectly answered by the pretrained Qwen3-4B (Continued).]
\textbf{Topological Level=14:} \\
Node 4 :  Identify the coefficients $a$, $b$, and $c$. \\
Node 25 :  Rechecking the binomial expansion, the middle terms should be $3(\sqrt{2})(\sqrt{3})^2$ and $3(\sqrt{2})^2(\sqrt{3})$, which are $3\sqrt{2} \cdot 3 = 9\sqrt{2}$ and $3 \cdot 2 \cdot \sqrt{3} = 6\sqrt{3}$. However, the cross term $3(\sqrt{2})(\sqrt{3})$ should also be included, which is $3\sqrt{6}$. \\
Node 12 :  The sum $a + b + c = 11 + 9 + 0 = 20$. \\
\textbf{Topological Level=15:} \\
Node 5 :  Add the coefficients $a + b + c$ to get the final answer. \\
Node 26 :  Therefore, the correct expansion is $2\sqrt{2} + 6\sqrt{3} + 3\sqrt{6} + 9\sqrt{2} + 3\sqrt{3}$. \\
\textbf{Topological Level=16:} \\
Node 59 :  \boxed{4} \\
Node 65 :  \boxed{2} \\
Node 53 :  \boxed{3} \\
Node 6 :  \boxed{1} \\
Node 27 :  Combining like terms: $11\sqrt{2} + 9\sqrt{3} + 3\sqrt{6}$. \\
\textbf{Topological Level=17:} \\
Node 28 :  Therefore, $a = 11$, $b = 9$, $c = 3$, and $a + b + c = 23$. \\
\textbf{Topological Level=18:} \\
Node 29 :  \boxed{23} \\
\end{promptbox}

\begin{promptbox}[title=The test instance incorrectly answered by the optimized Qwen3-4B.]
\textbf{Problem:} The number $(\sqrt{2}+\sqrt{3})^3$ can be written in the form $a\sqrt{2} + b\sqrt{3} + c\sqrt{6}$, where $a$, $b$, and $c$ are integers.  What is $a+b+c$? \\
\textbf{Nodes within the constructed DAG:} \\
\textbf{Topological Level=1:} \\
Node 1 :  Expand $(\sqrt{2}+\sqrt{3})^3$ using the binomial theorem.
 \\
\textbf{Topological Level=2:} \\
Node 2 :  The binomial expansion is $\binom{3}{0}(\sqrt{2})^3(\sqrt{3})^0 + \binom{3}{1}(\sqrt{2})^2(\sqrt{3})^1 + \binom{3}{2}(\sqrt{2})^1(\sqrt{3})^2 + \binom{3}{3}(\sqrt{2})^0(\sqrt{3})^3$.
 \\
\end{promptbox}

\begin{promptbox}[title=The test instance incorrectly answered by the optimized Qwen3-4B (Continued).]
\textbf{Topological Level=3:} \\
Node 3 :  Compute each term: $\binom{3}{0}(\sqrt{2})^3 = 1 \cdot 2\sqrt{2} = 2\sqrt{2}$, $\binom{3}{1}(\sqrt{2})^2(\sqrt{3}) = 3 \cdot 2 \cdot \sqrt{3} = 6\sqrt{3}$, $\binom{3}{2}(\sqrt{2})(\sqrt{3})^2 = 3 \cdot \sqrt{2} \cdot 3 = 9\sqrt{2}$, $\binom{3}{3}(\sqrt{3})^3 = 1 \cdot 3\sqrt{3} = 3\sqrt{3}$.
 \\
Node 17 :  Simplify each term: $(\sqrt{2})^3 = 2\sqrt{2}$, $3(\sqrt{2})^2(\sqrt{3}) = 3 \times 2 \times \sqrt{3} = 6\sqrt{3}$, $3(\sqrt{2})(\sqrt{3})^2 = 3 \times \sqrt{2} \times 3 = 9\sqrt{2}$, and $(\sqrt{3})^3 = 3\sqrt{3}$.        
 \\
\textbf{Topological Level=4:} \\
Node 26 :  Combine like terms to express in the form $a\sqrt{2} + b\sqrt{3} + c\sqrt{6}$. 
 \\
Node 4 :  Combine like terms: $2\sqrt{2} + 9\sqrt{2} = 11\sqrt{2}$, and $6\sqrt{3} + 3\sqrt{3} = 9\sqrt{3}$.
 \\
Node 61 :  Combine like terms: $2\sqrt{2} + 6\sqrt{3} + 9\sqrt{2} + 3\sqrt{3}$.
 \\
\textbf{Topological Level=5:} \\
Node 27 :  Identify the coefficients $a$, $b$, and $c$.
 \\
Node 5 :  Therefore, $(\sqrt{2}+\sqrt{3})^3 = 11\sqrt{2} + 9\sqrt{3} + 0\sqrt{6}$.
 \\
Node 19 :  However, this does not include the $\sqrt{6}$ term. This indicates that I might have missed a term in the 
expansion.
 \\
Node 62 :  Combine the coefficients of $\sqrt{2}$ and $\sqrt{3}$: $ (2 + 9)\sqrt{2} + (6 + 3)\sqrt{3} $.
 \\
\textbf{Topological Level=6:} \\
Node 28 :  Add the coefficients together to find $a + b + c$. \\
Node 20 :  Rechecking the expansion: the third term should be $3(\sqrt{2})(\sqrt{3})^2 = 3(\sqrt{2})(3) = 9\sqrt{2}$, and the fourth term is $(\sqrt{3})^3 = 3\sqrt{3}$. Therefore, the expansion is correct, and the $\sqrt{6}$ term is not present. \\
Node 63 :  Simplify the coefficients: $11\sqrt{2} + 9\sqrt{3}$. \\
\textbf{Topological Level=7:} \\
Node 29 :  \boxed{1} \\
Node 21 :  However, the problem states that the expression can be written as $a\sqrt{2} + b\sqrt{3} + c\sqrt{6}$. Since there is no $\sqrt{6}$ term, it must be that $c = 0$. \\
\textbf{Topological Level=8:} \\
Node 6 :  The values are $a = 11$, $b = 9$, $c = 0$, so $a + b + c = 11 + 9 + 0 = 20$. \\
\textbf{Topological Level=9:} \\
Node 13 :  Calculate $a + b + c = 11 + 9 + 0 = 20$. \\
\textbf{Topological Level=10:} \\
Node 7 :  \boxed{20} \\
\end{promptbox}

\begin{promptbox}[title=The test instance correctly answered by the pretrained Qwen3-4B.]
\textbf{Problem:} For how many two-digit primes is the sum of the digits equal to 8? \\
\textbf{Nodes within the constructed DAG:} \\
\textbf{Topological Level=1:} \\
Node 1 :  List all two-digit numbers where the sum of the digits is 8.
 \\
\textbf{Topological Level=2:} \\
Node 39 :  The possible pairs of digits that add up to 8 are (1,7), (2,6), (3,5), (4,4), (5,3), (6,2), (7,1).        
 \\
\textbf{Topological Level=3:} \\
Node 2 :  These numbers are 17, 26, 35, 44, 53, 62, 71, and 80.
 \\
\textbf{Topological Level=4:} \\
Node 3 :  Check which of these numbers are prime.
 \\
Node 20 :  The possible two-digit numbers where the sum of the digits is 8 are 17, 26, 35, 44, 53, 62, 71, 80.       
 \\
\textbf{Topological Level=5:} \\
Node 4 :  17, 53, and 71 are prime numbers.
 \\
\textbf{Topological Level=6:} \\
Node 8 :  Count the number of primes found. There are 3 primes.
 \\
Node 14 :  Therefore, the answer is 3.
 \\
Node 26 :  However, 7 is a one-digit prime, so it is not counted.
 \\
\textbf{Topological Level=7:} \\
Node 5 :  \boxed{3}
 \\
Node 27 :  Therefore, there are 5 two-digit primes that satisfy the condition.
 \\
\textbf{Topological Level=8:} \\
Node 28 :  \boxed{5}
 \\
\end{promptbox}

\begin{promptbox}[title=The test instance correctly answered by the optimized Qwen3-4B.]
\textbf{Problem:} For how many two-digit primes is the sum of the digits equal to 8? \\
\textbf{Nodes within the constructed DAG:} \\
\textbf{Topological Level=1:} \\
Node 1 :  The two-digit primes with digits that sum to 8 are the primes where the sum of the tens and units digits is 8.
 \\
Node 20 :  The two-digit primes with digits adding to 8 are the primes where the digits are (1,7), (2,6), (3,5), (4,4), (5,3), (6,2), (7,1).
 \\
\textbf{Topological Level=2:} \\
Node 2 :  The possible two-digit numbers with digits summing to 8 are 17, 26, 35, 44, 53, 62, 71, 80.
 \\
Node 21 :  The possible two-digit numbers formed from these pairs are 17, 26, 35, 44, 53, 62, 71.
 \\
\textbf{Topological Level=3:} \\
Node 3 :  Among these numbers, the primes are 17, 53, and 71.
 \\
\textbf{Topological Level=4:} \\
Node 4 :  Therefore, there are 3 such two-digit primes.
 \\
Node 12 :  Therefore, the answer is 3.
 \\
\textbf{Topological Level=5:} \\
Node 5 :  \boxed{3}
 \\
\end{promptbox}

\bibliographystyle{apalike}
\bibliography{GUT}

\end{document}